\documentclass{article}

     \usepackage[preprint]{neurips_2019}

\usepackage[utf8]{inputenc} % allow utf-8 input
\usepackage[T1]{fontenc}    % use 8-bit T1 fonts
\usepackage{hyperref}       % hyperlinks
\usepackage{url}            % simple URL typesetting
\usepackage{booktabs}       % professional-quality tables
\usepackage{amsfonts}       % blackboard math symbols
\usepackage{nicefrac}       % compact symbols for 1/2, etc.
\usepackage{microtype}      % microtypography
\usepackage{multirow}
\usepackage{booktabs}       % professional-quality tables
\usepackage{amsmath,amsthm,amssymb,amsfonts}
\usepackage{mathtools}% For figures
\usepackage{graphicx} % more modern
\usepackage[export]{adjustbox}

\newcommand{\hpobench}{HPO-Bench}
\newcommand{\ie}{i.\,e.}

\title{Tabular Benchmarks for Joint Architecture and Hyperparameter Optimization}

\author{%
  Aaron Klein \\
  Department of Computer Science\\
  University of Freiburg\\
  \texttt{kleinaa@cs.uni-freiburg.de} \\
  \And
  Frank Hutter\\
  Department of Computer Science\\
  University of Freiburg\\
  \texttt{fh@cs.uni-freiburg.de} \\ 
}

\begin{document}

\maketitle

\begin{abstract}

Due to the high computational demands executing a rigorous comparison between hyperparameter optimization (HPO) methods is often cumbersome.
The goal of this paper is to facilitate a better empirical evaluation of HPO methods by providing benchmarks that are cheap to evaluate, but still represent realistic use cases.
We believe these benchmarks provide an easy and efficient way to conduct reproducible experiments for neural hyperparameter search.
Our benchmarks consist of a large grid of configurations of a feed forward neural network on four different regression datasets including architectural hyperparameters and hyperparameters concerning the training pipeline.
Based on this data, we performed an in-depth analysis to gain a better understanding of the properties of the optimization problem, as well as of the importance of different types of hyperparameters.
Second, we exhaustively compared various different state-of-the-art methods from the hyperparameter optimization literature on these benchmarks in terms of performance and robustness.

\end{abstract}

\section{Introduction}\label{sec:intro}

Despite the tremendous success achieved by deep neural networks in the last few years~\citep{krizhevsky-nips12,sutskever-nips14}, using them in practice remains challenging due to their sensitivity to many hyperparameters and architectural choices.
Even experts often only find the right setting to train the network successfully by trial-and-error.
There has been a recent line of work in hyperparameter optimization (HPO)~\citep{snoek-nips12a,hutter-lion11a,bergstra-nips11a,li-iclr17,klein-aistats17,falkner-icml18}\footnote{for a review see \citet{feurer-automlbook18a}} and neural architecture search (NAS)~\citep{baker-iclr17,zoph-iclr17a,real-icml17a,elsken-iclr19,liu-iclr19} that tries to automate this process by casting it as an optimization problem. 
However, since each function evaluation consists of training and evaluating a deep neural network, running these methods can take several hours or even days. 

We believe that this hinders advancing the field since thorough evaluation is key to develop new methods and, due to their internal randomness, requires many independent runs of every method to get robust statistical results.
Recent work~\citep{eggensperger-aaai15} proposed to use surrogate benchmarks, which replace the original benchmark by a regression model trained on data generated offline.
During optimization, instead of training and validating the actual hyperparameter configuration, the regression model is queried and its prediction is returned to the optimizer.
Orthogonally to this work, we performed an exhaustive search for a large neural architecture search problem and compiled all architecture and performance pairs into a neural architecture search benchmark~\citep{ying-arxiv19}.

For the current work, we collected a large grid of hyperparameter configurations of feed forward neural networks (see Section \ref{sec:setup_hpobench}) for regression.
Based on the gathered data, we give an in-depth analysis of the properties of the optimization problem (see Section \ref{sec:statistics_hpobench}), as well as of the importance of hyperparameters and architectural choices (see Section \ref{sec:hpo_importance_hpobench}).
Finally, we benchmark a variety of well-known HPO methods from the literature, such as Bayesian optimization, evolutionary algorithms, reinforcement learning, a bandit based method and random-search (Section \ref{sec:comparison_hpobench}) leading to new insights on how the different methods compare.
The dataset, as well as the code to carry out these experiments is publicly available at \url{https://github.com/automl/nas_benchmarks}.

\section{Setup}\label{sec:setup_hpobench}

We use 4 popular UCI~\citep{lichman-13} datasets for regression: protein structure~\citep{rana-protein}, slice localization~\citep{slice-dataset}, naval propulsion~\citep{coraddu-naval} and parkinsons telemonitoring~\citep{tsanas-parkinson}. We call them \hpobench-Protein, \hpobench-Slice, \hpobench-Naval and \hpobench-Parkinson, respectively.
For each dataset we used $60\%$ for training, $20\%$ for validation and $20\%$ for testing (see Table~\ref{tab:uci_dataset_statistics} for an overview) and removed features that were constant over the entire dataset.
Afterwards, all features and targets values were normalized by subtracting the mean and dividing by the variance of the training data.
These datasets do not require deeper neural network architectures which means we can train them on CPUs rather than GPUs and hence we can afford to run many configurations.

\begin{table}[h!]
  \center
  \scriptsize
  \caption[Dataset split for HPOBench]{Dataset splits}
  \begin{tabular}{c c c c c}
	\toprule
	Dataset & $\#$ training datapoints & $\#$ validation datapoints & $\#$ test datapoints & $\#$ features \\
	\midrule
	\hpobench-Protein & 27\,438 & 9\,146 & 9\,146 & 9 \\
	\hpobench-Slice & 32\,100 & 10\,700 & 10\,700 & 385 \\
	\hpobench-Naval & 7\,160 & 2\,388 & 2\,388 & 15  \\
	\hpobench-Parkinson & 3\,525 & 1\,175 & 1\,175 &  20 \\
	\bottomrule
  \end{tabular}
  \label{tab:uci_dataset_statistics}
\end{table}

As the base architecture, we used a two layer feed forward neural network followed by a linear output layer on top.
The configuration space (denoted in Table~\ref{tab:cs_fcnet}) only includes a modest number of 4 architectural choice (number of units and activation functions for both layers) and 5 hyperparameters (dropout rates per layer, batch size, initial learning rate and learning rate schedule) in order to allow for an exhaustive evaluation of all the 62\,208 configurations resulting from discretizing the hyperparameters as in Table~\ref{tab:cs_fcnet}. 
We encode numerical hyperparameters as ordinals and all other hyperparameters as categoricals.
Each network was trained with Adam~\citep{kingma-iclr15} for 100 epochs, optimizing the mean squared error.
We repeated the training of each configuration 4 independent times with a different seed for the random number generator and recorded for each run the training / validation / test accuracy, training time and the number of trainable parameters.
We provide full learning curves (\ie{} validation and training error for each epoch) as an additional fidelity that can be used to benchmark multi-fidelity algorithms with the number of epochs as the budget.

\begin{table}[h!]
  \center
  \scriptsize
  \caption[Configuration space FC-Net HPOBench]{Configuration space of the fully connected neural network}
  \begin{tabular}{c c }
	\toprule
	Hyperparameters & Choices \\
	\midrule
	Initial LR & $\{.0005, .001, .005, .01, .05, .1\}$ \\
	Batch Size & $\{8, 16, 32, 64\}$ \\
	LR Schedule & $\{\text{cosine}, \text{fix}\}$\\
	Activation/Layer 1 & $\{\text{relu}, \text{tanh}\}$\\
	Activation/Layer 2 & $\{\text{relu}, \text{tanh}\}$\\
	Layer 1 Size & $\{16, 32, 64, 128, 256, 512\}$\\
	Layer 2 Size & $\{16, 32, 64, 128, 256, 512\}$\\
	Dropout/Layer 1 & $\{0.0, 0.3, 0.6\}$\\
	Dropout/Layer 2 & $\{0.0, 0.3, 0.6\}$\\
	\bottomrule
  \end{tabular}
  \label{tab:cs_fcnet}
\end{table}

\section{Dataset Statistics}\label{sec:statistics_hpobench}

We now analyze the properties of these datasets. First, for each dataset we computed the empirical cumulative distribution function (ECDF) of the test, validation and training error after 100 epochs and the total training time.
For each metric, we averaged over the 4 repetitions.
Additionally we computed the ECDF for the number of trainable parameters of each neural network architecture. 
To avoid clutter, we show here only the results for the \hpobench-Protein which we found to be consistent with the other datasets and present all results in Section~\ref{sec:supp_hpobench_dataset} in the supplemental material.

One can see in Figure~\ref{fig:ecdf} that the mean-squared-error (MSE) for training, validation and test is spread over an order of magnitude or more.
On one side only a small subset of configurations achieve a final MSE lower than 0.3 and on the other hand many outliers exist that achieve errors orders of magnitude above the average. Furthermore, due to the changing number of parameters, also the training time varies dramatically across configurations.

\begin{figure}[h!]
\begin{center}
 \includegraphics[width=0.48\linewidth]{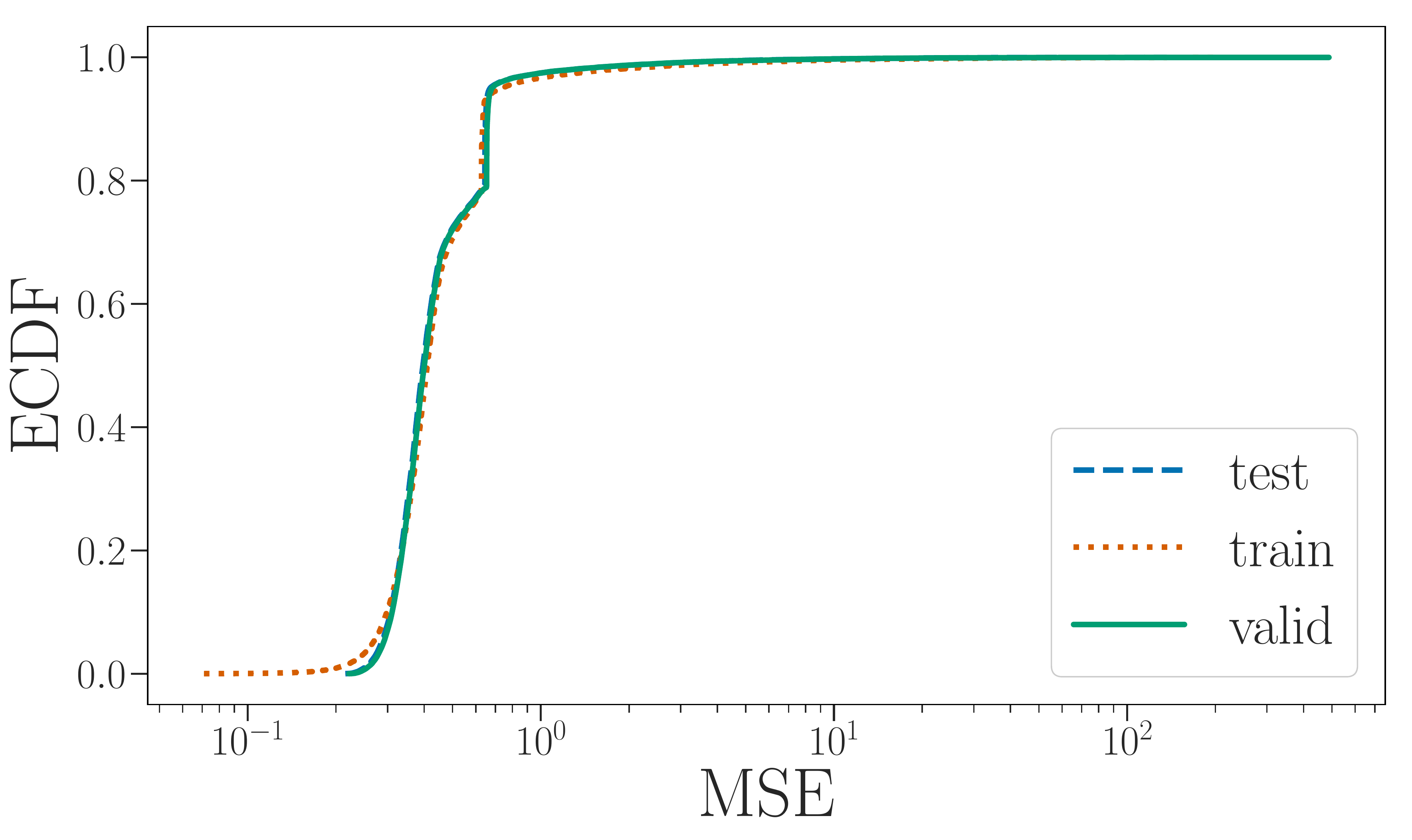}
 \includegraphics[width=0.48\linewidth]{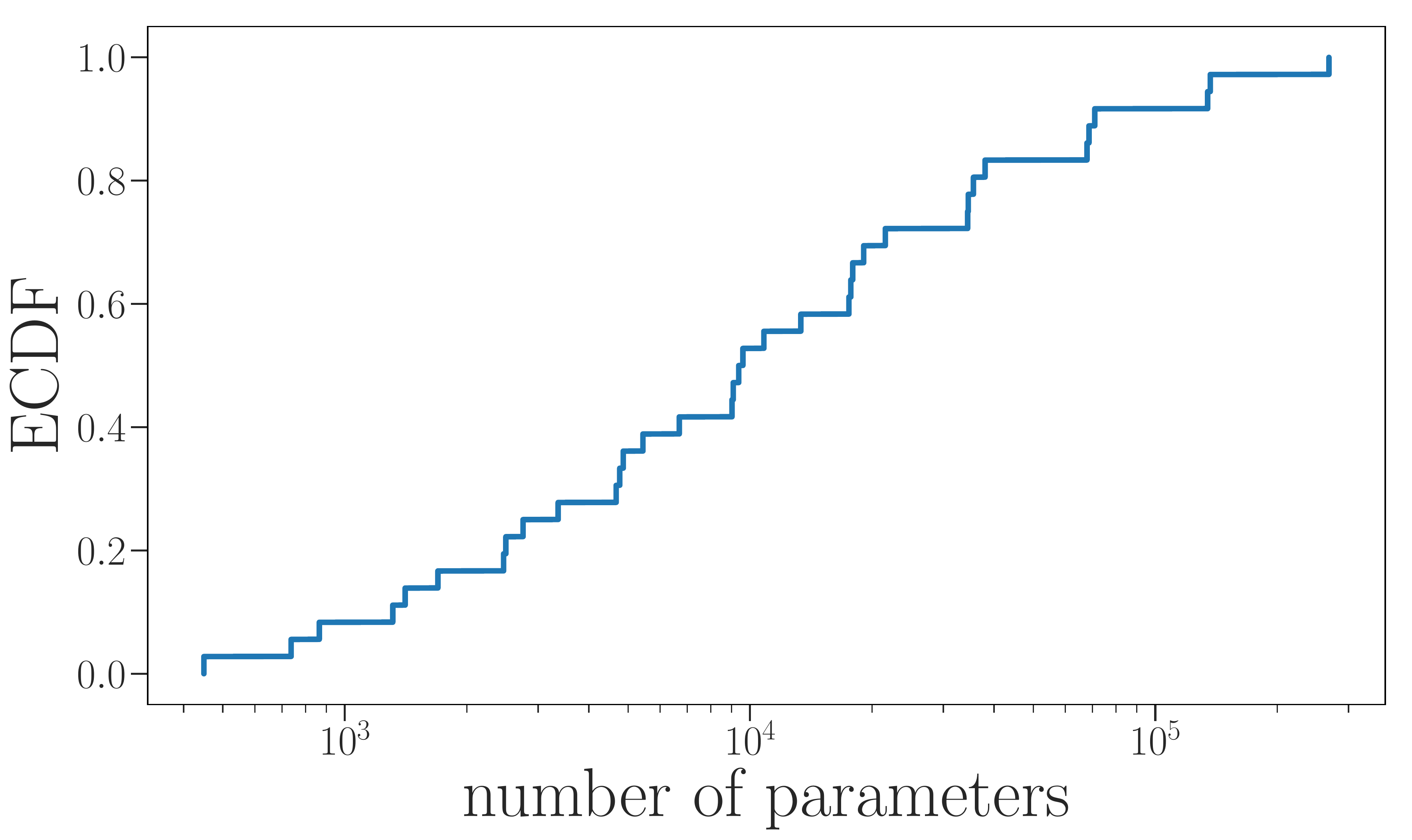}\\
 \includegraphics[width=0.48\linewidth]{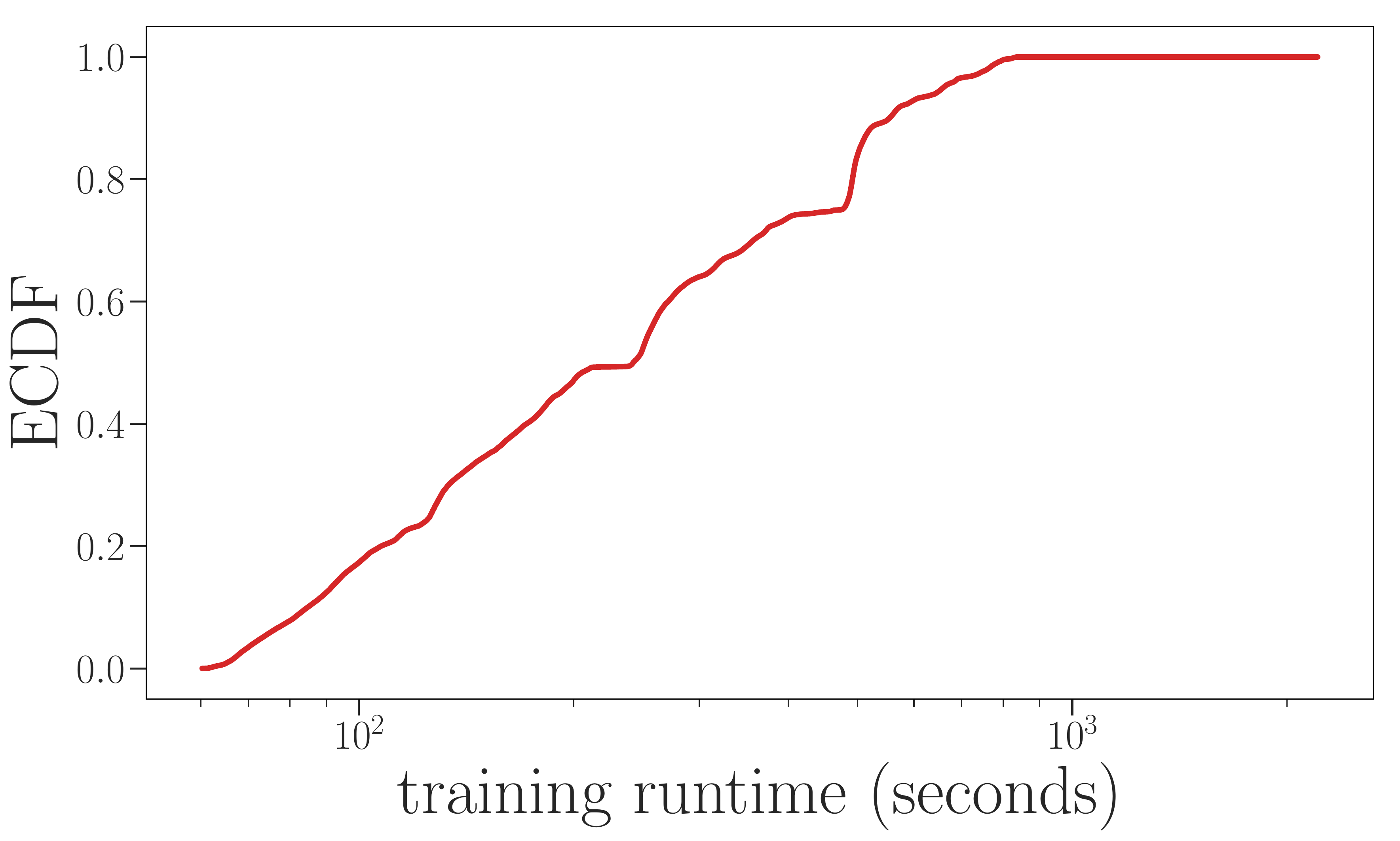}
 \includegraphics[width=0.48\linewidth]{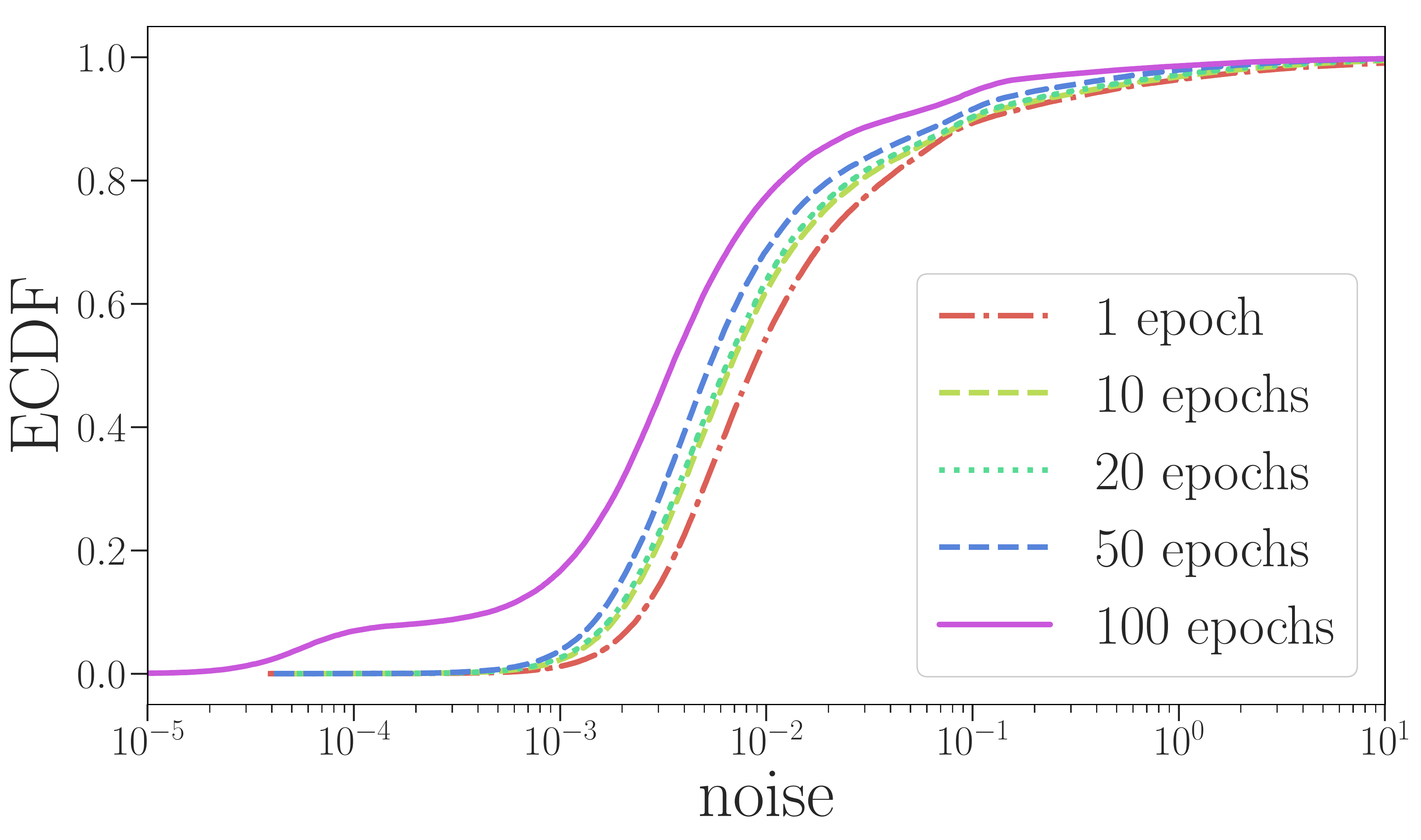}
 \caption[Empirical cumulative distributions \hpobench]{The empirical cumulative distribution (ECDF) of the average train/valid/test error after 100 epochs of training (upper left), the number of parameters (upper right), the training runtime (lower left) and the noise for different number of epochs (lower right) computed on HPO-NAS-bench-Protein. See Appendix~\ref{sec:supp_hpobench_dataset} for the ECDF plots of all datasets.}
 \label{fig:ecdf}
\end{center}
\end{figure}

Figure~\ref{fig:ecdf} bottom right shows the empirical cumulative distribution of the noise, defined as the standard deviation between the 4 repetitions for different number of epochs. We can see that the noise is heteroscedastic. That is, different configurations come with a different noise level. As expected, the noise decreases with an increasing number of epochs. 

\begin{figure}[h!]
\begin{center}
  \includegraphics[width=\linewidth]{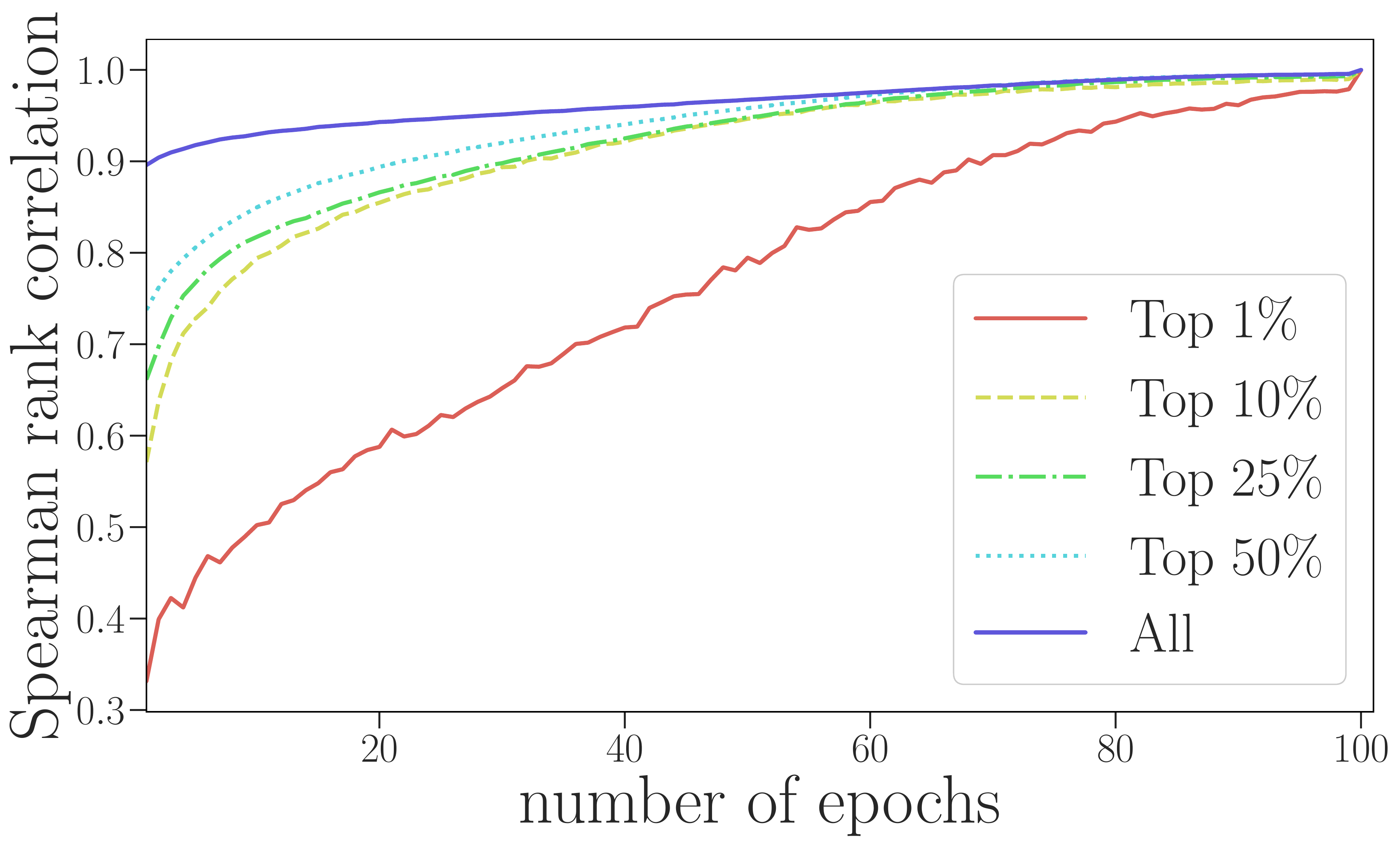}
  \caption[Rank correlation across budgets]{The Spearman rank correlation between different number of epochs to the highest budget of 100 epochs for the \hpobench-Protein when we consider all configurations or only the top $1\%$, $10\%$, $20\%$, and $50\%$ of all configurations based on their average test error. Results for other datasets are presented in Appendix~\ref{sec:supp_hpobench_dataset}.}
 \label{fig:rank_correlation_protein}
\end{center}
\end{figure}

For many multi-fidelity hyperparameter optimization methods, such as Hyperband~\citep{li-iclr17} or BOHB~\citep{falkner-icml18}, it is essential that the ranking of configurations on smaller budgets to higher budgets is preserved.
In Figures~\ref{fig:rank_correlation_protein}, we visualize the Spearman rank correlation between the performance of all hyperparameter configurations across different number of epochs and the highest budget of 100 epochs.
Since every hyperparameter optimization method needs to mainly focus on the top performing configurations, we also show the correlation for only the top $1\%$, $10\%$, $20\%$, and $50\%$ of all configurations.
As expected the correlation to the highest budget increases with increasing budgets.
If only top-performing configurations are considered, the correlation decreases, since their final performances are closer to each other.

\section{Hyperparameter Importance}\label{sec:hpo_importance_hpobench}

We now analyze how the different hyperparameters affect the final performance, first globally with help of the functional ANOVA~\citep{sobol-93,hutter-icml14a} and then from a more local point of view. Finally, we show how the top performing hyperparameter configurations correlate across the different datasets.
As in the previous section, we show here only the results for \hpobench-Protein and for all other dataset in Appendix~\ref{sec:supp_hpobench_importance}.

\subsection{Functional ANOVA}

To analyze the importance of hyperparameters, assessing the change of the final error with respect to changing a single hyperparameter at a time, we used the fANOVA tool by~\citet{hutter-icml14a}.
It quantifies the importance of a hyperparameter by marginalizing the error obtained by setting it to a specific value over all possible values of all other hyperparameters.
The importance of a hyperparameter is then the variation in error that is explained by this hyperparameter.
In default setting this tool fits a random forest model on the observed function values in order to compute the marginal predictions.
However, since we already evaluated the full configuration space, we do not even need to use a model and can compute the required integrals directly.

\begin{figure}[t]
\centering
\includegraphics[width=0.32\linewidth]{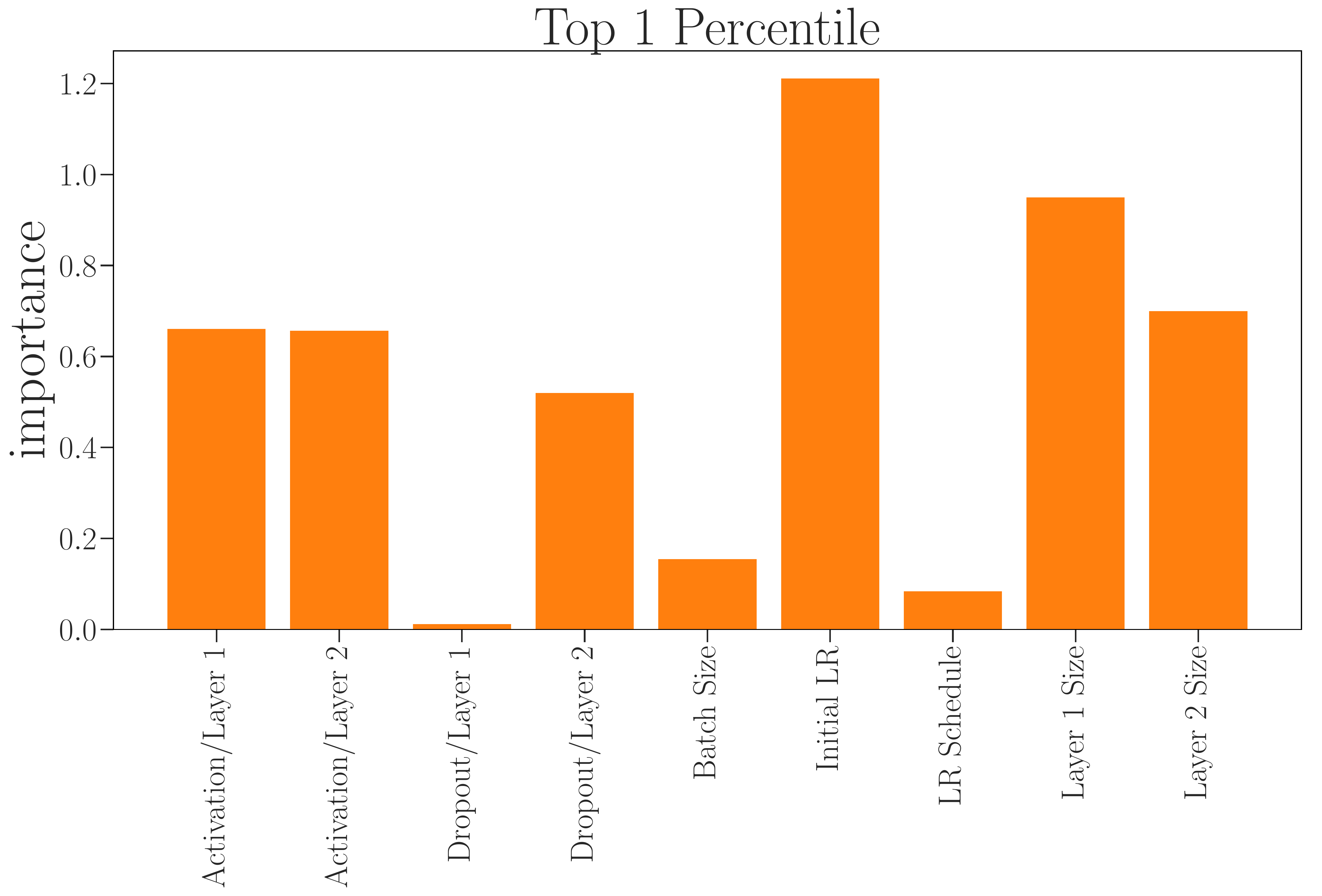}
\includegraphics[width=0.32\linewidth]{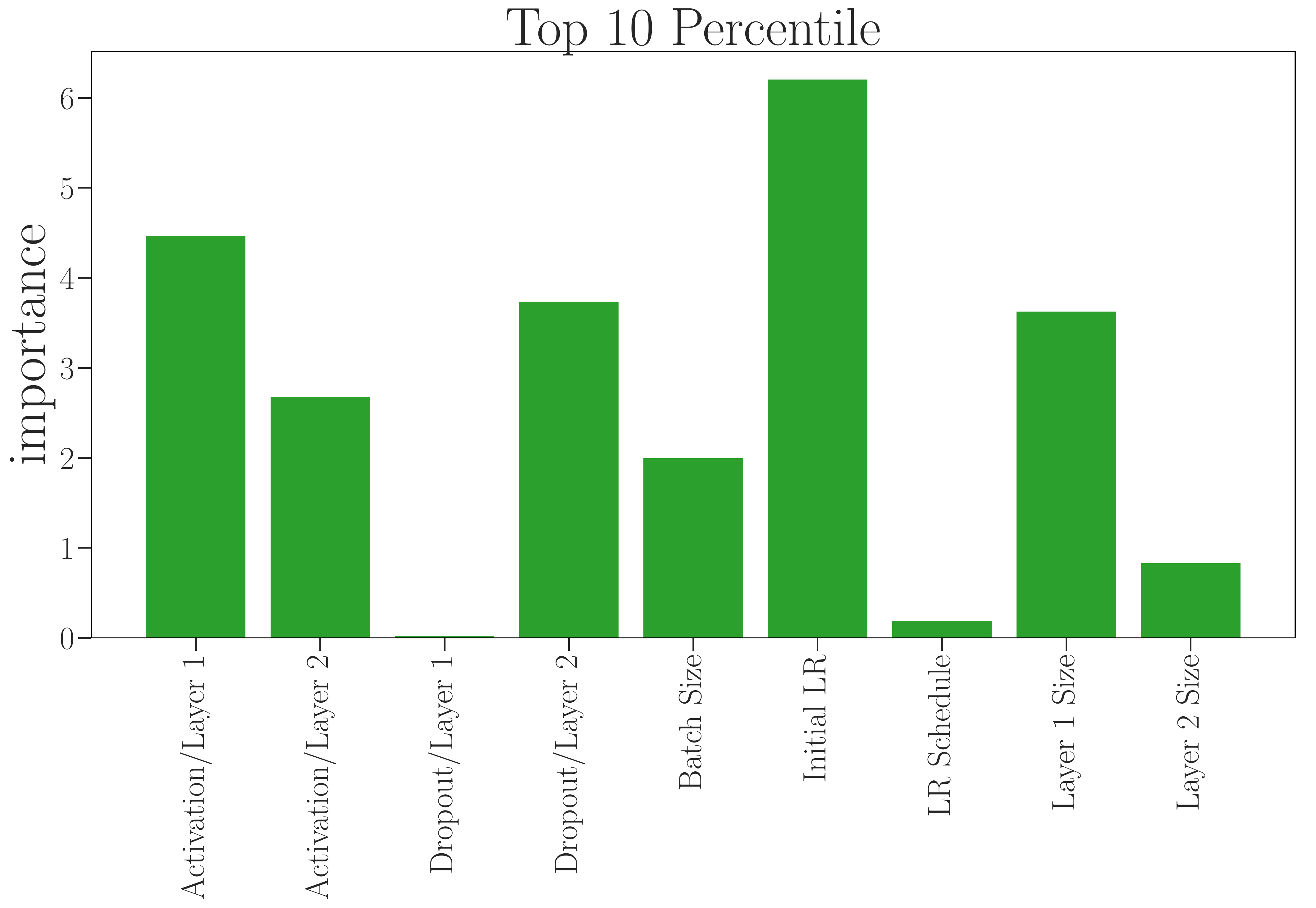}
\includegraphics[width=0.32\linewidth]{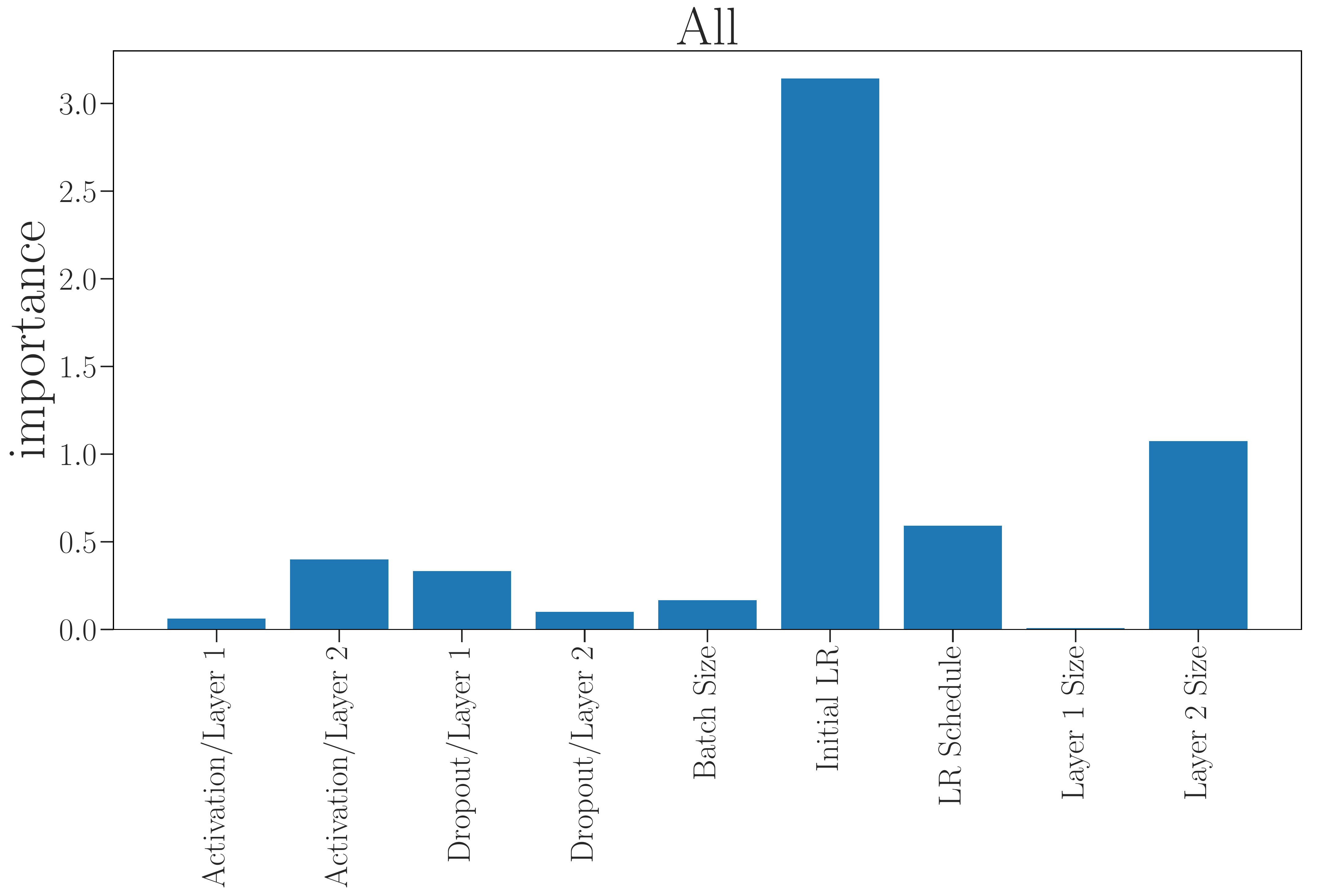}\\
\includegraphics[width=0.32\linewidth,valign=t]{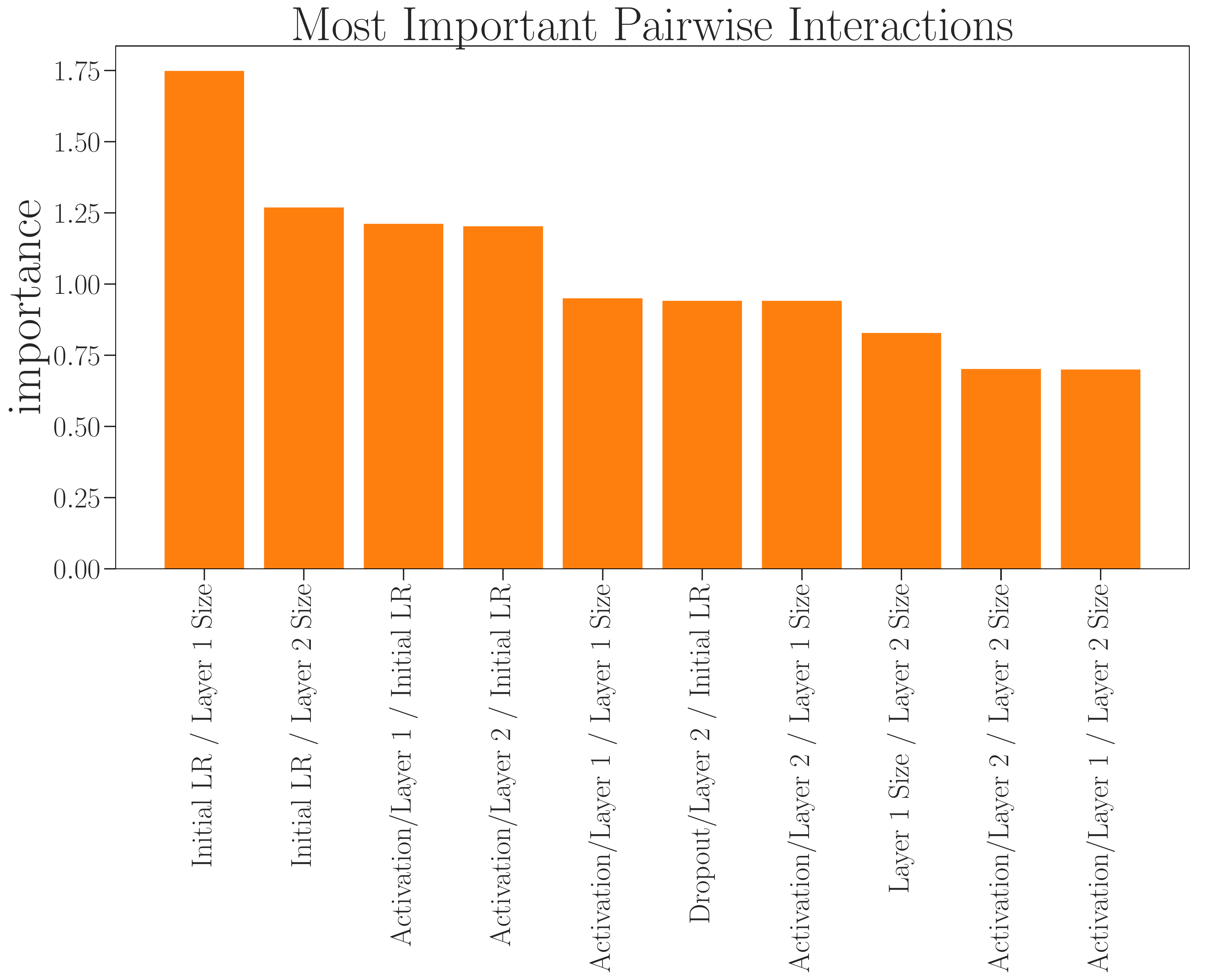}
\includegraphics[width=0.32\linewidth,valign=t]{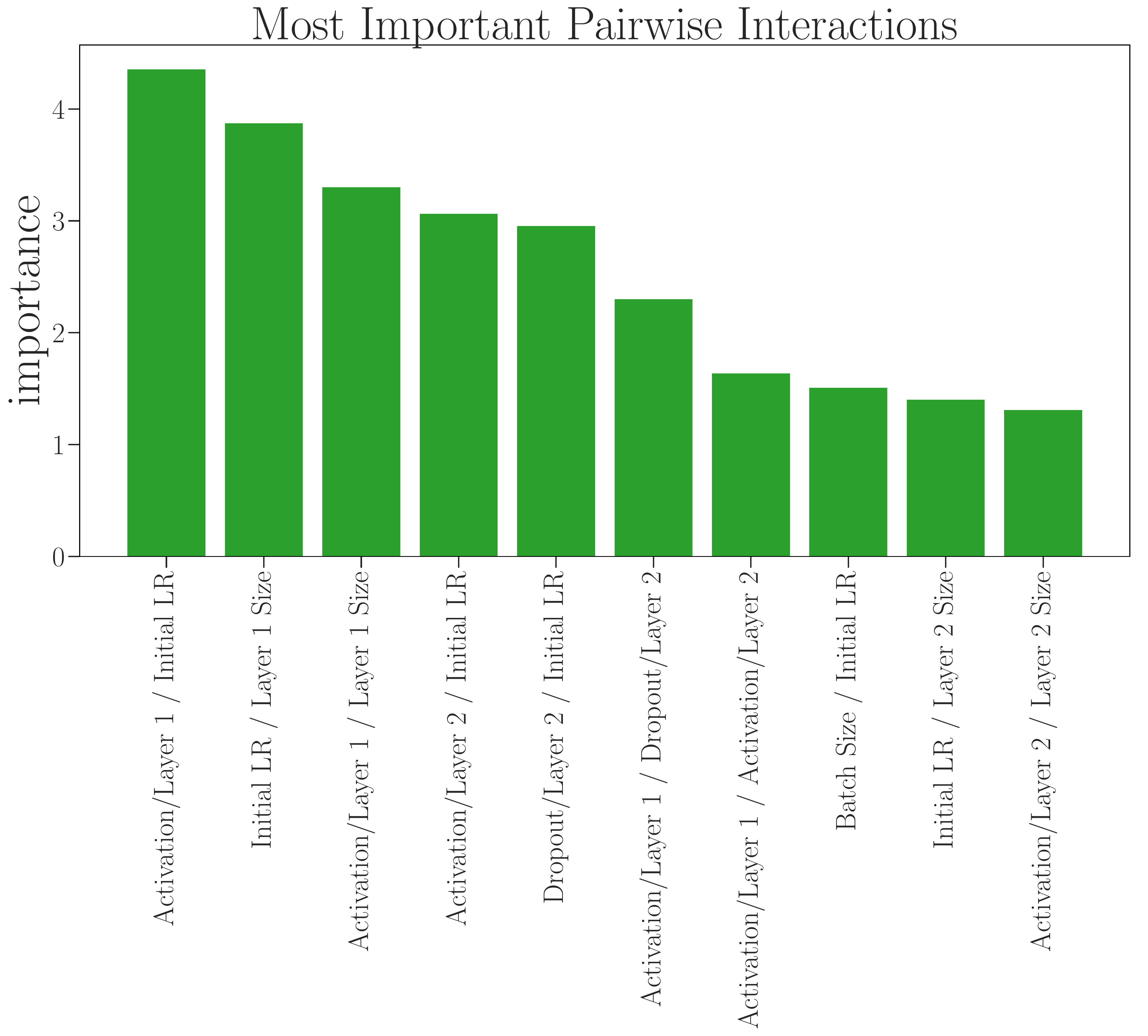}
\includegraphics[width=0.32\linewidth,valign=t]{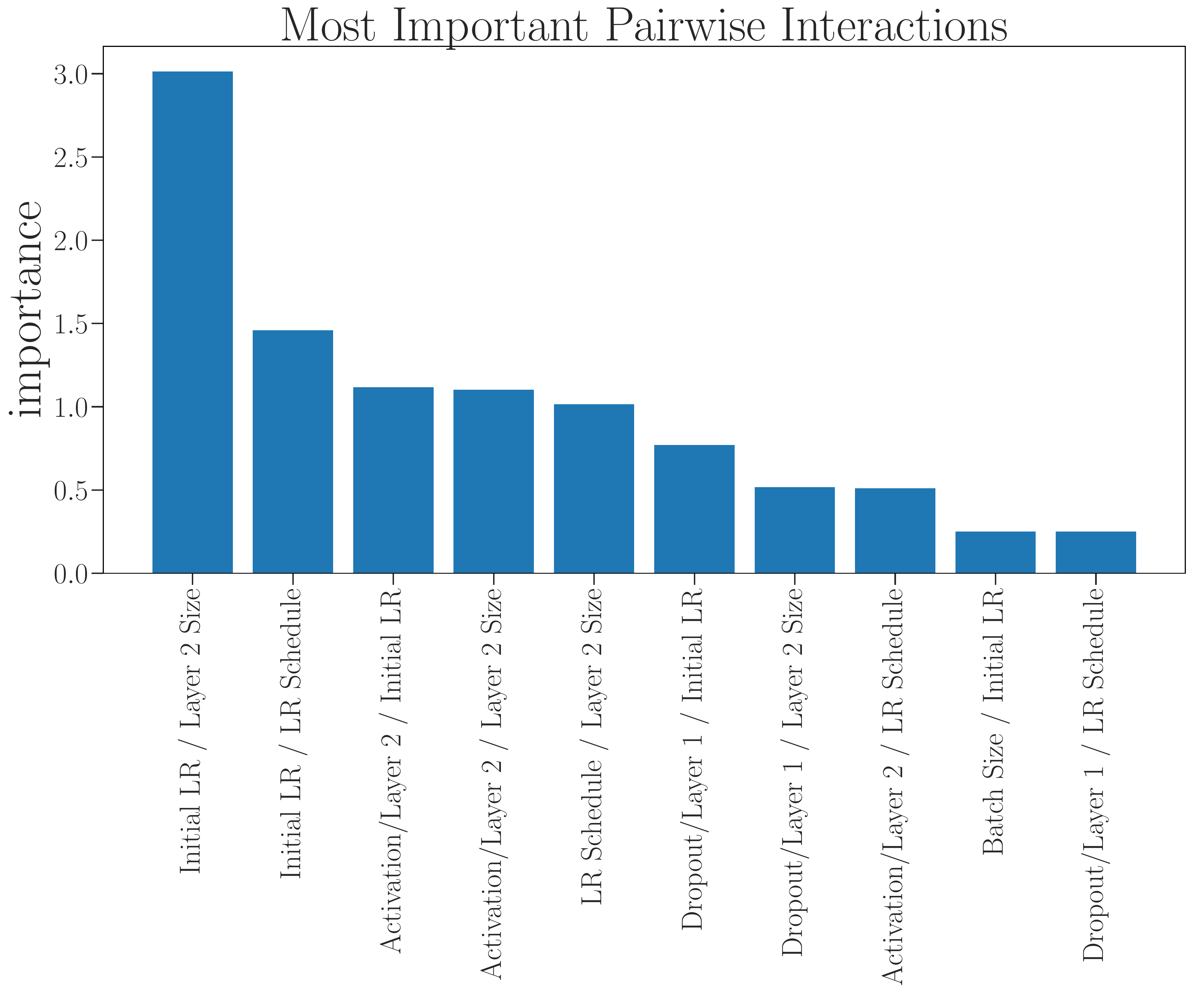}
\caption[fANOVA HPOBench]{Top row: Importance of the different hyperparameter based on the fANOVA for: (left) only the top $1\%$ ; (middle) top $10\%$ ; (right) all configurations. Bottom row: most important hyperparameter pairs with (left) only the top $1\%$ ; (middle) top $10\%$ ; (right) all configurations.}
\label{fig:importance}
\end{figure}

As can be seen in Figure~\ref{fig:importance} (upper right), on average across the entire configuration space, the initial learning rate obtained the highest importance value. However, the importance of individual hyperparameters is very small due to a few outliers with very high errors, which only happen for a few combinations of several hyperparameter values.
We also computed the importance values of hyperparameter configuration pairs (see Figure~\ref{fig:importance} lower right for the ten most important pairs). 
These general small values for single and pairwise hyperparameters indicates that the benchmarks exhibit higher order interaction effects.
Unfortunately, computing higher than second order interaction effect is computational infeasible.

A better estimate of hyperparameter importance in a region of the configuration space with reasonable performance can be obtained by only using the best performing configuration for the fANOVA.
Figure~\ref{fig:importance} (left) shows the results of this procedure with the $1$ percentile and Figure~\ref{fig:importance} (middle) with the $10$ percentile of all configuration. This shows that in this more interesting part of the configuration space, other hyperparameters also become important.

\subsection{Local Neighbourhood}

While the fANOVA takes the whole configuration space into account, we now focus on a more local view around the best configuration (incumbent) to see how robust it is against small perturbations.
We show in Table~\ref{tab:neighbors} the change in performance if we flip single hyperparameters of the incumbent while keeping all other hyperparameters fixed.
Additionally, we also show in the rightmost column the relative change $\frac{y_{new} - y_{\star}}{y_{\star}}$ between the error of the incumbent $y_{\star}$ and the new observed error $y_{new}$.

Interestingly, the highest drop in performance occurs by changing the activation function of the first and the second layer from relu to tanh.
This is despite the fact that tanh is a much more common activation function for regression than relu.
In contrast, increasing the batch size only has a marginal effect on the performance.

\begin{table}[h]
  \center
  \caption[Local neighborhood of the incumbent]{Performance change if single hyperparameters of the incumbent (average test error 0.2153) are flipped.}
  \begin{tabular}{|c|c|c|c|}
    \hline
	Hyperparameter & Change & Test Error & Relative Change \\
	\hline

Batch Size & $8 \rightarrow 16$ & 0.2163 & 0.0042 \\
Initial LR & $0.0005 \rightarrow 0.001$ & 0.2169 & 0.0072 \\
Layer 2 Size & $512 \rightarrow 256$ & 0.2203 & 0.0231 \\
Layer 1 & $512 \rightarrow 256$ & 0.2216 & 0.0288 \\
Dropout/Layer 2 & $0.3 \rightarrow 0.6$ & 0.2257 & 0.0478 \\
LR Schedule & $cosine \rightarrow const$ & 0.2269 & 0.0534 \\
Dropout/Layer 2 & $0.3 \rightarrow 0.0$ & 0.2280 & 0.0587 \\
Dropout/Layer 1 & $0.0 \rightarrow 0.3$ & 0.2307 & 0.0711 \\
Activation/Layer 2 & $relu \rightarrow tanh$ & 0.2875 & 0.3351 \\
Activation/Layer 1 & $relu \rightarrow tanh$ & 0.3012 & 0.3987 \\

	\hline
  \end{tabular}
  \label{tab:neighbors}
\end{table}

\subsection{Ranking across Datasets}

\begin{figure}[t]
\centering
\includegraphics[width=0.32\linewidth]{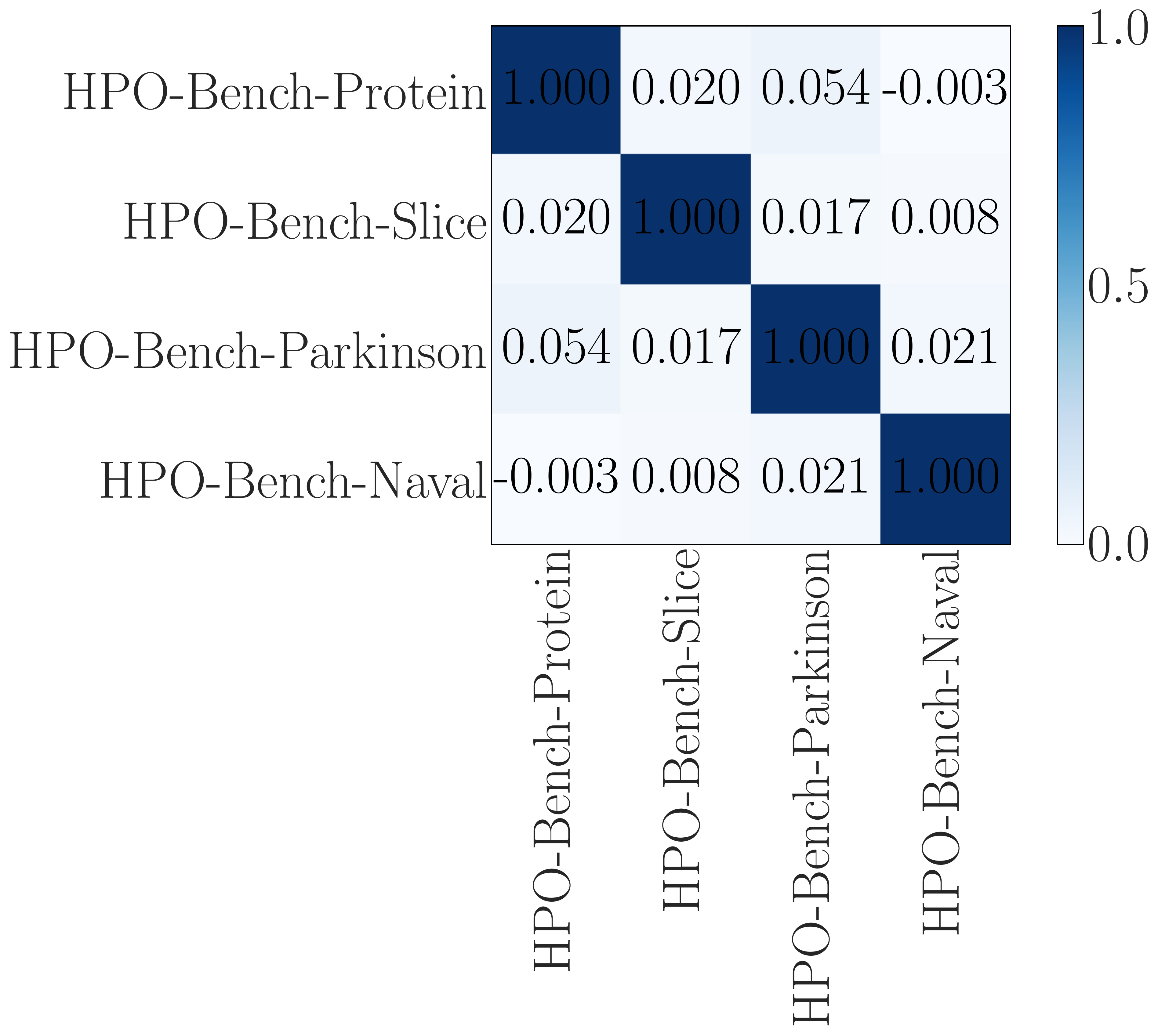}
\includegraphics[width=0.32\linewidth]{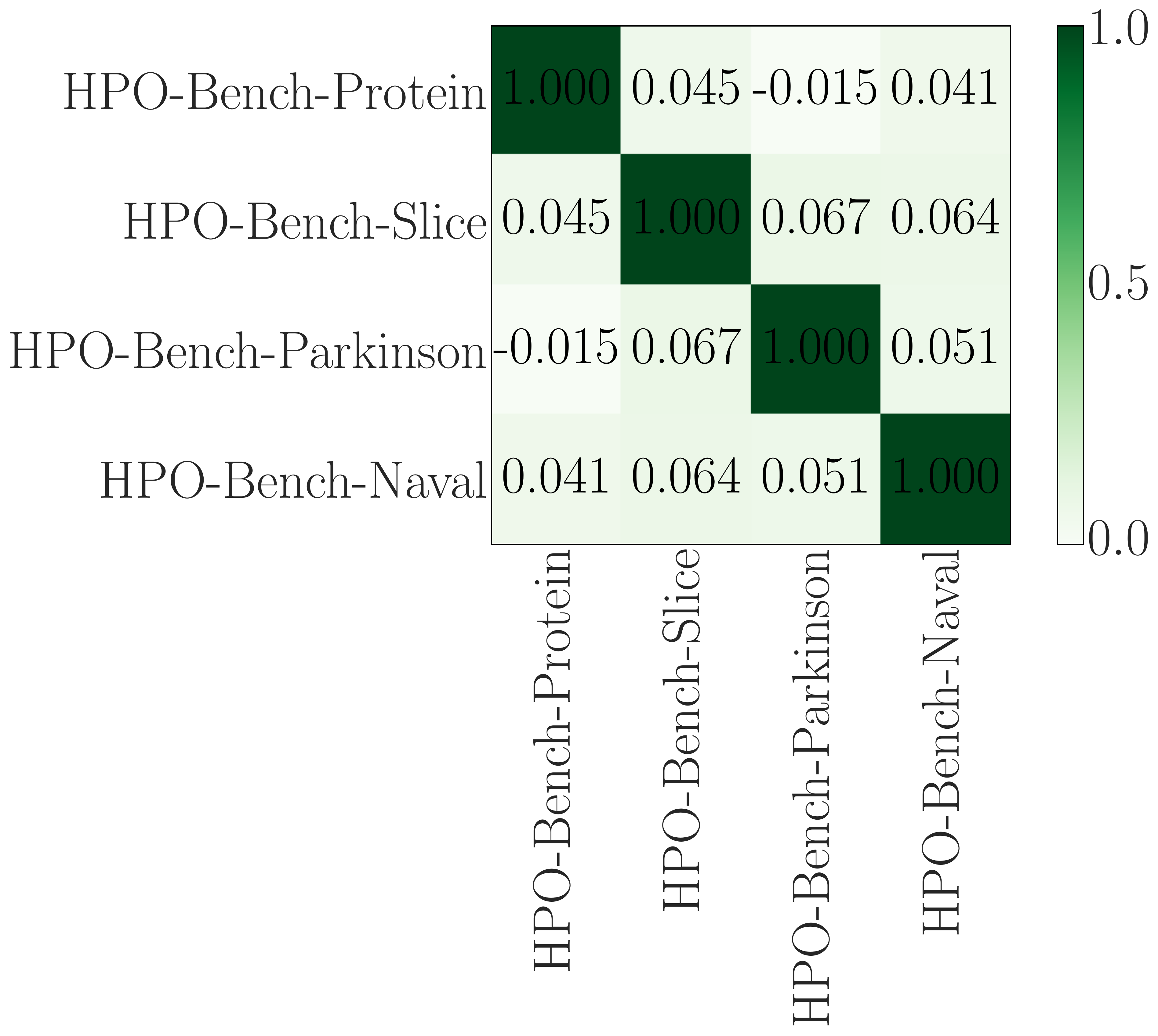}
\includegraphics[width=0.32\linewidth]{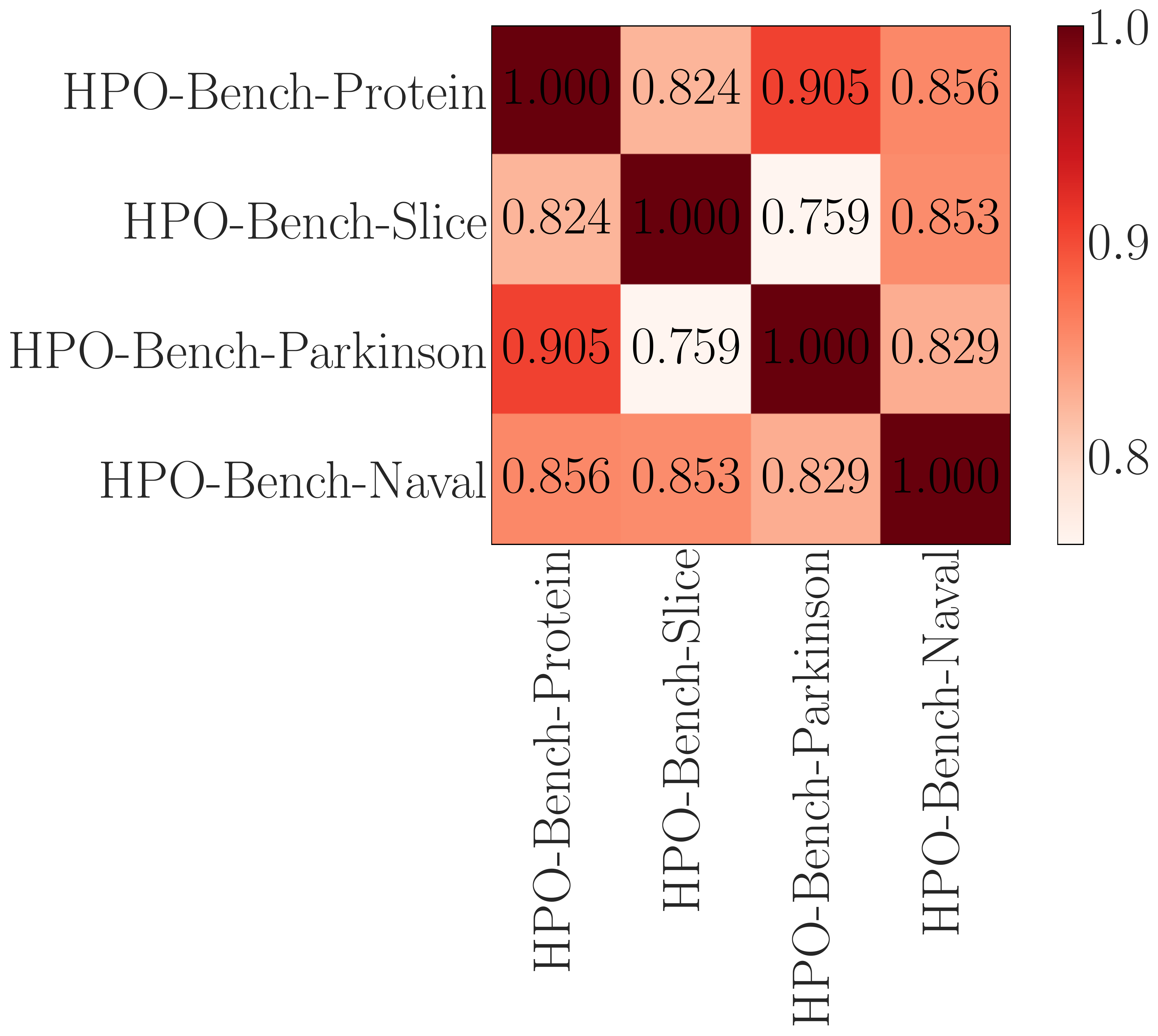}
\caption[Rank correlation across HPOBench datasets]{Correlation of the ranks for (left) top-$1\%$ / (middle) top-$10\%$ and all hyperparameter configurations across all four datasets.}
\label{fig:ranking}
\end{figure}

We now analyse hyperparameter configurations across the four different datasets.
In Table~\ref{tab:global_opt_fcnet} we can see that the best configuration in terms of average test error changes only slightly across datasets.
For some hyperparameters, such as the learning rate, a certain value, $0.0005$ in this case, can be used for all all datasets whereas other hyperparameters, for example the activation functions, need to be set differently.

To see how the performance of all hyperparameter configurations correlates across datasets, we computed for every configuration on every dataset its rank in terms of final average test performance.
Figure~\ref{fig:ranking} shows the Spearman rank correlation between the different datasets if we consider the first percentile (left), the $10$th percentile (middle) or all configurations (right).
The correlation decreases if we only consider the best-performing configurations, which implies that it does not suffice to reuse a good configuration from a different datasets to achieve top performance on a new dataset.
Nevertheless, the correlation for all configurations is high which indicates that multi-task methods could be able to exploit previously collected data.

\begin{table}[h!]
  \center
  \scriptsize
  \caption[Best configurations FC-Net HPOBench]{Best configurations in terms of average test error for each dataset}
  \begin{tabular}{c c c c c}
	\toprule
	Hyperparameters & \hpobench-Protein & \hpobench-Slice & \hpobench-Naval & \hpobench-Parkinson\\
	\midrule
	Initial LR & 0.0005 & 0.0005 & 0.0005 & 0.0005 \\
	Batch Size & 8 & 32 & 8 & 8 \\
	LR Schedule & cosine & cosine & cosine & cosine  \\
	Activation/Layer 1 & relu & relu & tanh & tanh  \\
	Activation/Layer 2 & relu & tanh & relu & relu  \\
	Layer 1 Size & 512 & 512 & 128 & 128 \\
	Layer 2 Size & 512 & 512 & 512 & 512 \\
	Dropout/Layer 1 & 0.0 & 0.0 & 0.0 & 0.0 \\ 
	Dropout/Layer 2 & 0.3 & 0.0 & 0.0 & 0.0  \\

	\bottomrule
  \end{tabular}
  \label{tab:global_opt_fcnet}
\end{table}

\section{Comparison}\label{sec:comparison_hpobench}

In this section we use the generated benchmarks to evaluate different HPO methods. 
To mimic the randomness that comes with evaluating a configuration, in each function evaluation we randomly sample one of the four performance values.
To obtain a realistic estimate of the wall-clock time required for each optimizer, we accumulated the stored runtime of each configuration the optimizer evaluated.
We do not take the additional overhead of the optimizer into account since it is negligible compared to the training time of the neural network.
After each function evaluation we estimate the incumbent as the configuration with the lowest observed error and compute the regret between the incumbent and the globally best configuration in terms of test error. 
We performed 500 independent runs of each method and report the median and the 25th and 75th quantile.

\subsection{Performance over Time}

\begin{figure}[h!]
\centering
\includegraphics[width=.48\textwidth]{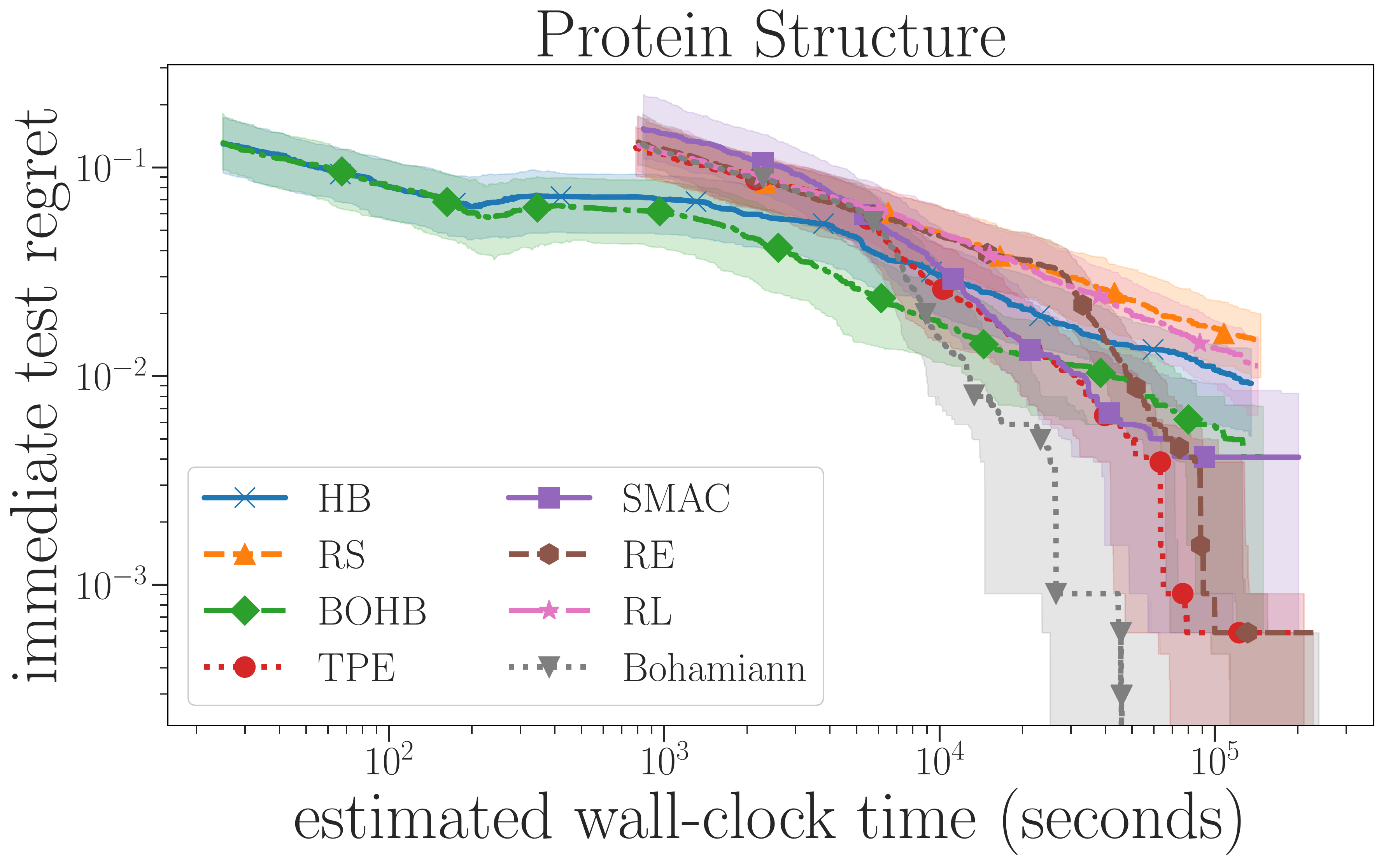}
\includegraphics[width=.48\textwidth]{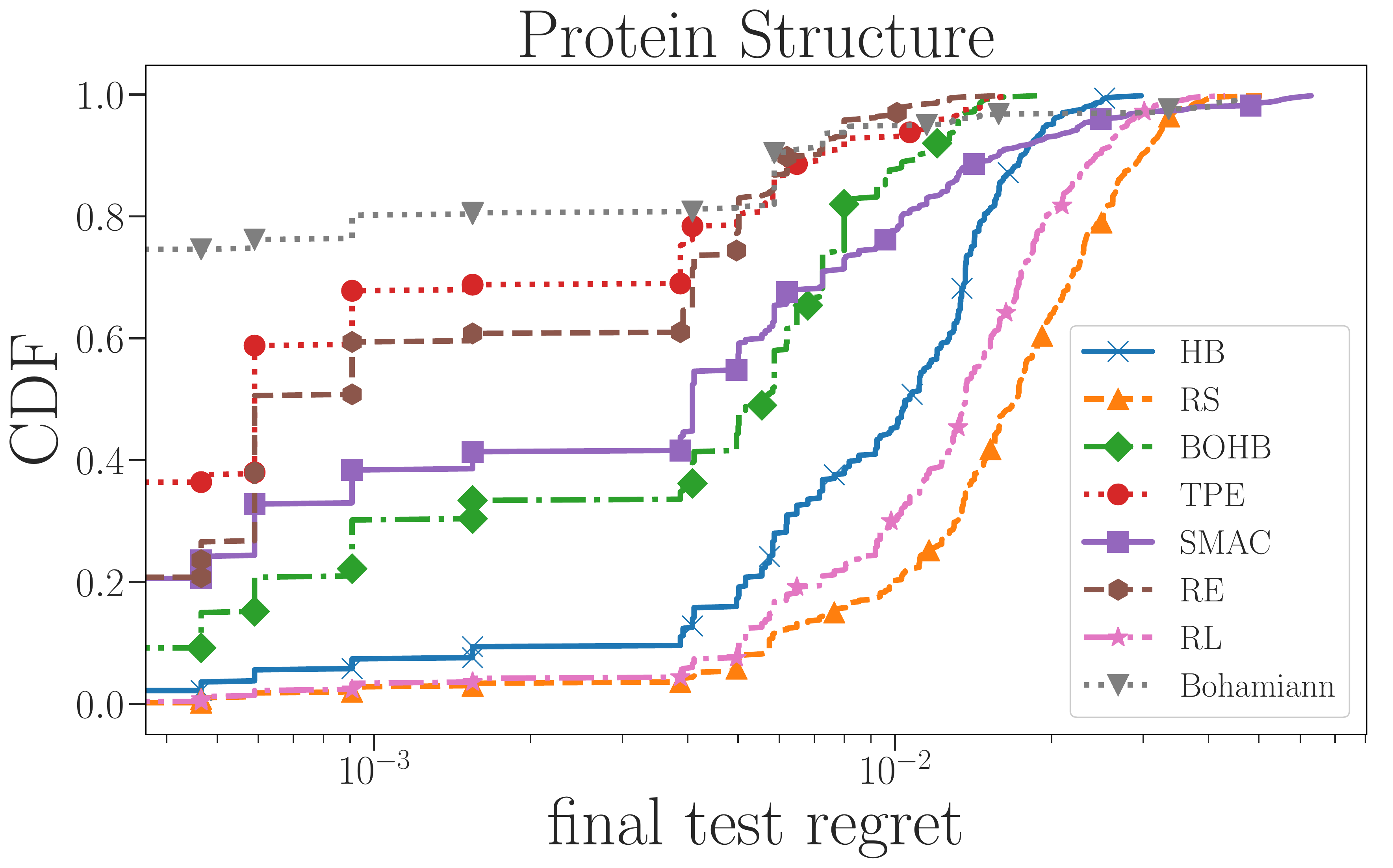}
\caption[Comparison \hpobench-Protein]{\label{fig:comparison} Left: Comparison of various HPO methods on the \hpobench-Protein datasets. For each method, we plot the median and the 25th and 75th quantile (shaded area) of the test regret of the incumbent (determined based on the validation performance) across 500 independent runs. Right: Empirical cumulative distribution of the final performance over all runs of each methods after $10^5$ seconds.}
\end{figure}

We compared the following HPO methods from the literature (see Figure~\ref{fig:comparison}): random search~\citep{bergstra-jmlr12a}, SMAC~\citep{hutter-lion11a}\footnote{We used SMAC3 from \url{https://github.com/automl/SMAC3}}, Tree Parzen Estimator (TPE)~\citep{bergstra-nips11a}\footnote{We used Hyperopt from \url{https://github.com/hyperopt/hyperopt}}, Bohamiann~\citep{springenberg-nips16}\footnote{We used the implementation from \citet{klein-bayesopt17}}, Regularized Evolution~\citep{real-aaai19}, Hyperband (HB)~\citep{li-iclr17} and BOHB~\citep{falkner-icml18}\footnote{For both HB and BOHB we used the implementation from \url{https://github.com/automl/HpBandSter}}.
Inspired by the recent success of reinforcement learning for neural architecture search~\citep{zoph-iclr17a}, we also include a similar reinforcement learning strategy (RL), which however does not use an LSTM as controller but instead uses REINFORCE~\citep{williams-ml92} to optimize the probability of each categorical variable directly~\citep{ying-arxiv19}.
Each method that operates on the full budget of 100 epochs was allowed to perform 500 function evaluations.
For BOHB and HB we set the minimum budget to 3 epochs, the maximum budget to 100, $\eta$ to 3 and the number of successive halving iterations to 125 (which leads to roughly the same amount of function evaluation time as the other methods).
More details about the meta-parameters of the different optimizers are described in Appendix~\ref{sec:supp_hpobench_comparison}.

Figure~\ref{fig:comparison} left show the performance over time for all methods. Results for the other datasets can be found in Appendix~\ref{sec:supp_hpobench_comparison}.
We can make the following observations:
\begin{itemize}
  \item As expected, Bayesian optimization methods, \ie{} SMAC, TPE and Bohamiann worked as well as RS in the beginning but started to perform superior once they obtained a meaningful model.
    Interestingly, while all Bayesian optimization methods start improving at roughly the same time, they converge to different optima, which we attribute to their different internal models.
  \item The same holds for BOHB, which is in the beginning as good as HB but starts outperforming it as soon as it obtains a meaningful model.
    Note that, compared to TPE, BOHB uses a multivariate KDE, instead of a univariate KDE, which is able to model interactions between hyperparameter configurations.
    We attributed TPE's superior performance over BOHB on these benchmarks to its very aggressive optimization of the acquisition function.
    BOHB's performance could be probably improved by optimizing its own meta-parameters since its default values were determined on continuous benchmarks~\citep{falkner-icml18} (where it outperformed TPE).

  \item HB achieved a reasonable performance relatively quickly but only slightly improves over simple RS eventually.
  \item RE needed more time than Bayesian optimization methods to outperform RS; however, it achieved the best final performance, since, compared to Bayesian optimization methods, it does not suffer from any model missmatch.
  \item RL requires even more time to improve upon RS than RE or Bayesian optimization and seems to be too sample inefficient for these tasks.
\end{itemize}

\subsection{Robustness}

Besides achieving good performance, we argue that robustness plays an important role in practice for HPO methods.
Figure~\ref{fig:comparison} shows the empirical cumulative distribution of the test regret for the final incumbent after $10^5$ seconds for \hpobench-Protein across all 500 runs of each method.

While RE achieves a lower mean test regret than TPE it seems to be less robust with respect to its internal randomness.
Interestingly, while all methods have non-zero probability to achieve a final test regret of $10^{-3}$ within $10^5$ seconds, only Bohamiann, RE and TPE are able to achieve this regret in more than $60\%$ of the cases. 
Also none of the methods is able to converge consistently to the same final regret.

\section{Conclusions}

We presented new tabular benchmarks for neural architecture and hyperparameter search that are cheap to evaluate but still recover the original optimization problem, enabling us to rigorously compare various methods from the literature.
Based on the data we generated for these benchmarks, we had a closer look at the difficulty of the optimization problem and the importance of different hyperparameters.

In future work, we will generate more of these benchmarks for other architectures and datasets. 
Ultimately, we hope that such benchmarks will help the community to easily reproduce experiments and evaluate new developed methods without spending enormous compute resources.

\newpage

\bibliographystyle{apa}

\bibliography{lib}

\newpage

\appendix 

\section{Dataset Statistics}\label{sec:supp_hpobench_dataset}

\begin{figure}[h!]
\begin{center}
 \includegraphics[width=0.24\linewidth]{plots/cdf_mse_protein_structure.pdf}
 \includegraphics[width=0.24\linewidth]{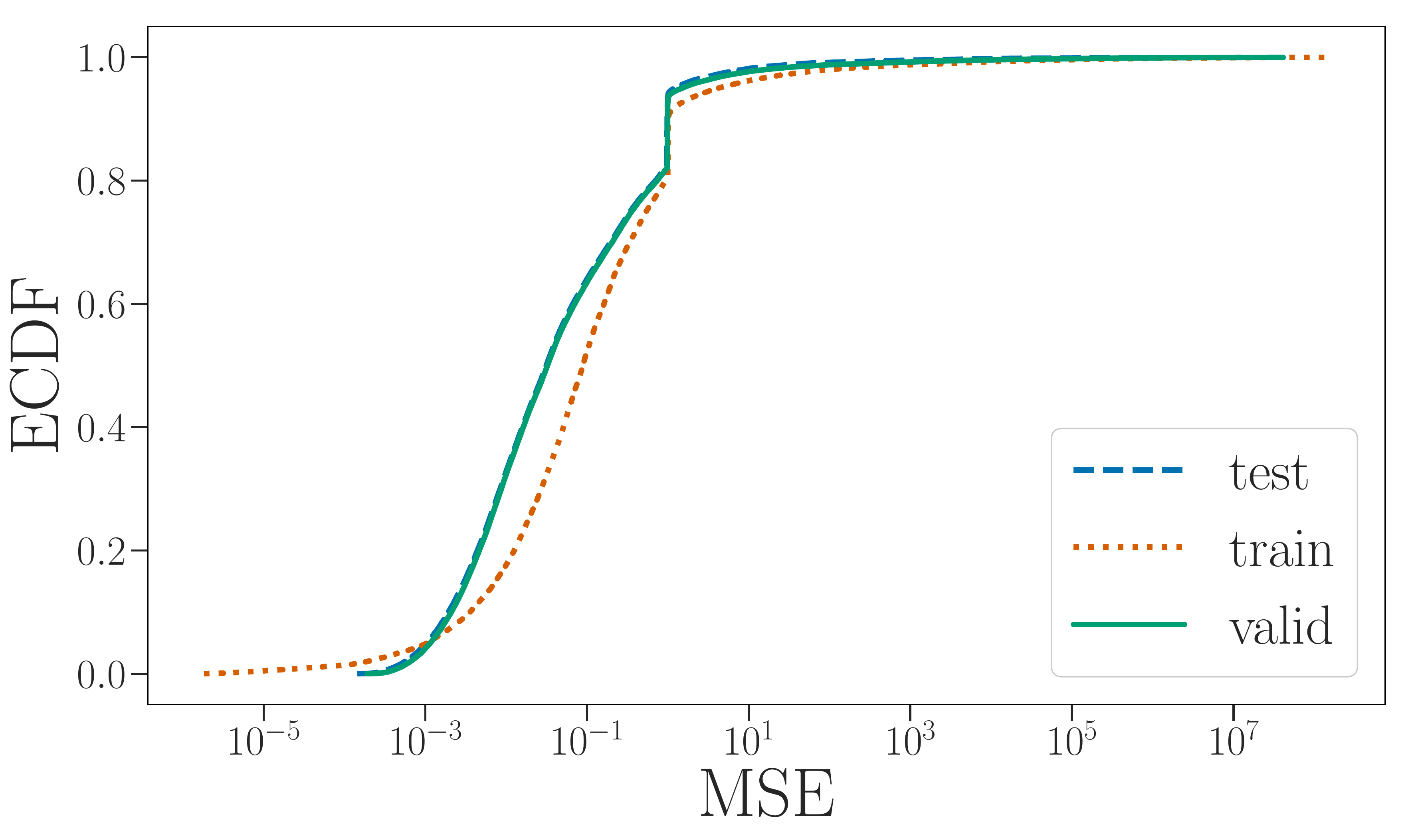}
 \includegraphics[width=0.24\linewidth]{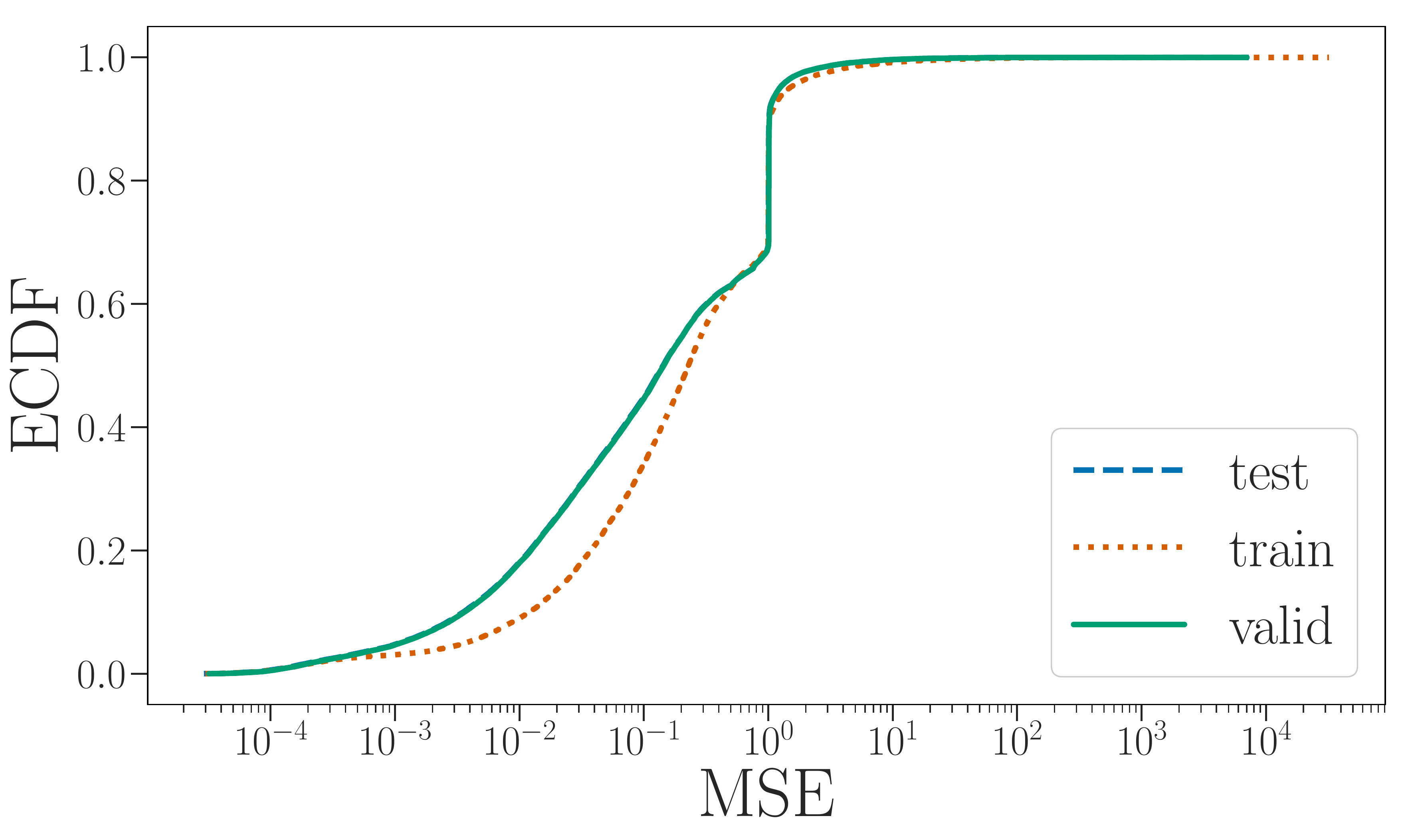}
 \includegraphics[width=0.24\linewidth]{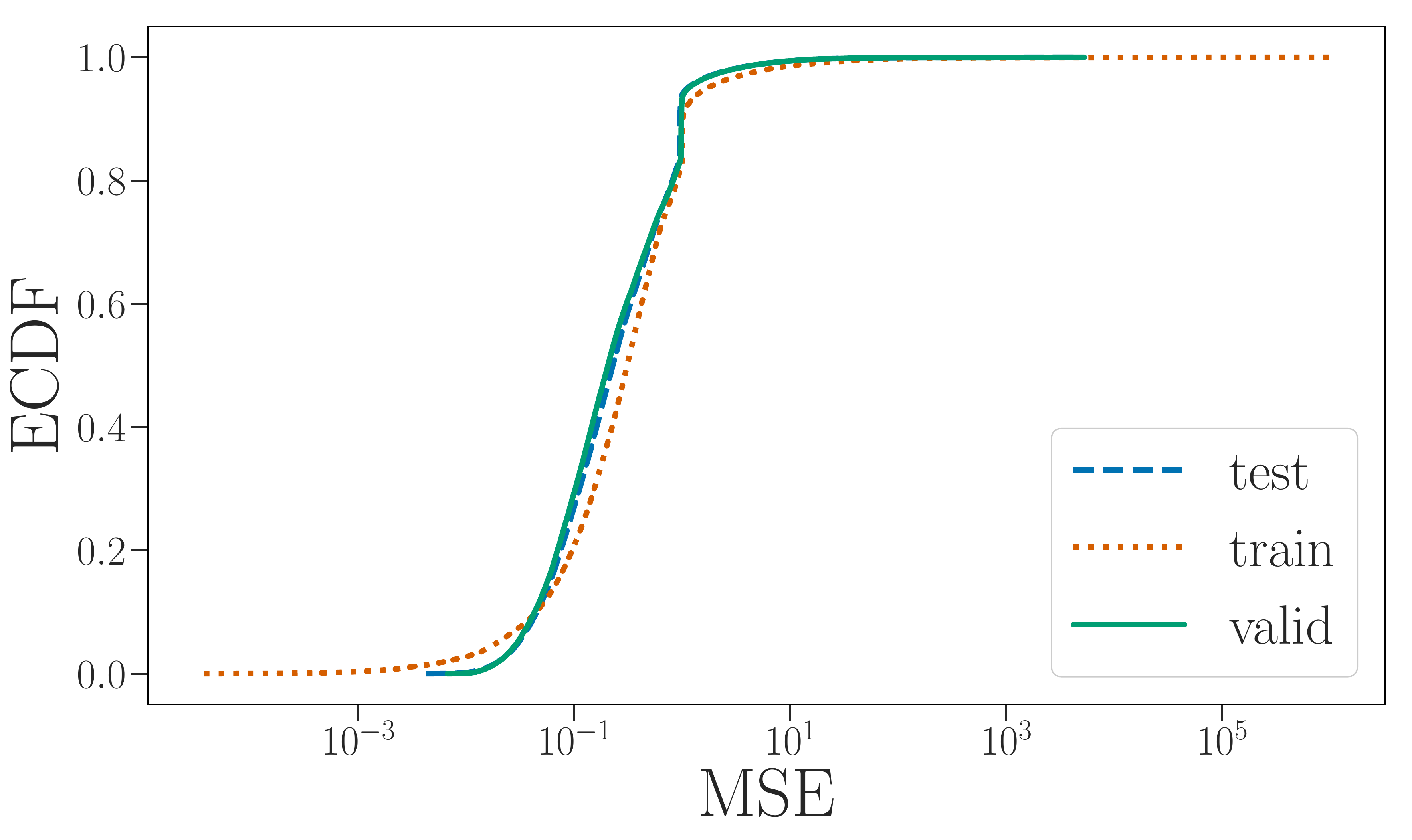}

 \caption[Empirical cumulative distributions accuracy]{The empirical cumulative distribution (ECDF) of the average train/valid/test error after 100 epochs of training, computed on \hpobench-Protein (left), \hpobench-Slice (left middle), \hpobench-Naval (right middle) and \hpobench-Parkinson (right).}
 \label{fig:ecdf_mse_all}
\end{center}
\end{figure}

\begin{figure}[h!]
\begin{center}
 \includegraphics[width=0.24\linewidth]{plots/cdf_parameters_protein_structure.pdf}
 \includegraphics[width=0.24\linewidth]{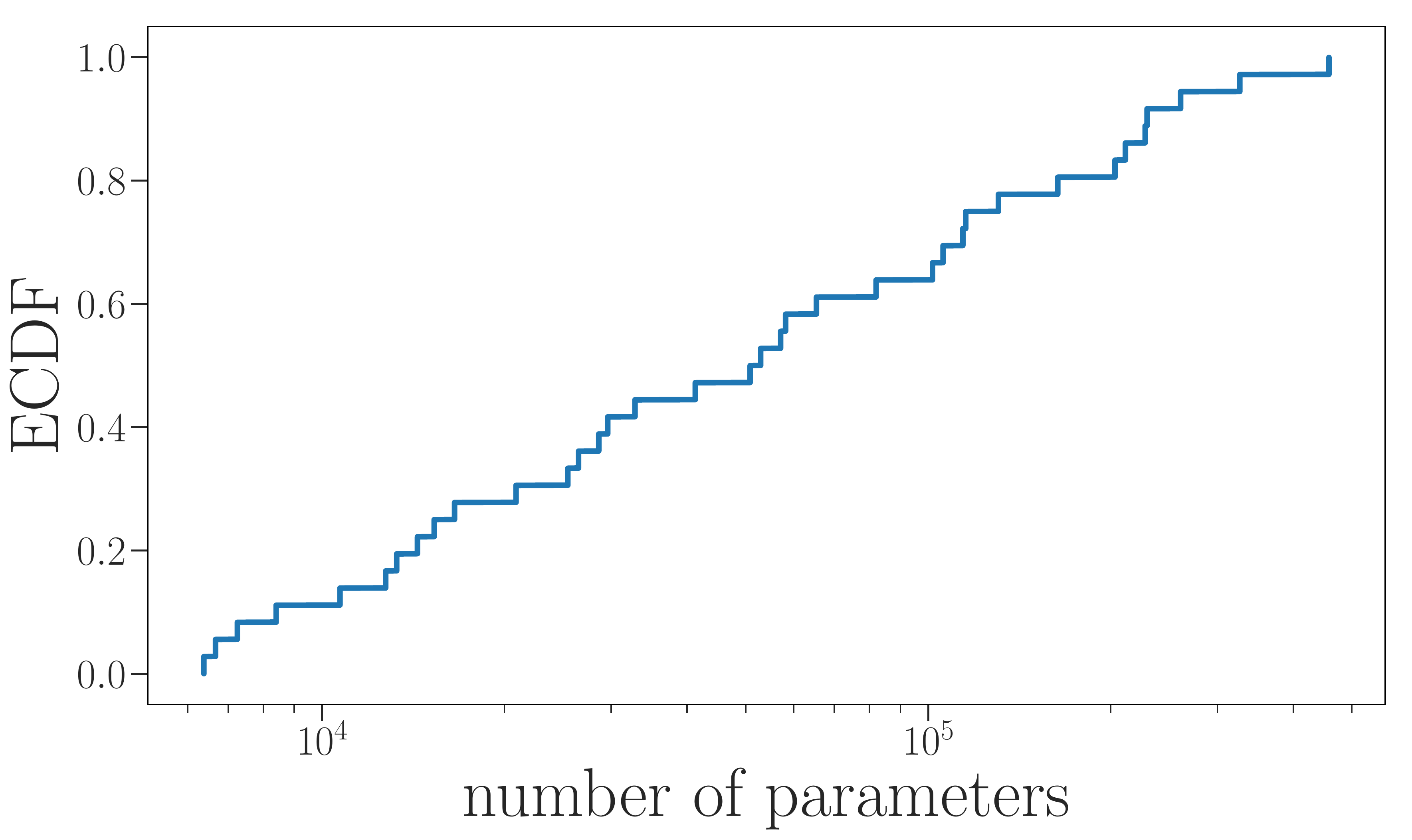}
 \includegraphics[width=0.24\linewidth]{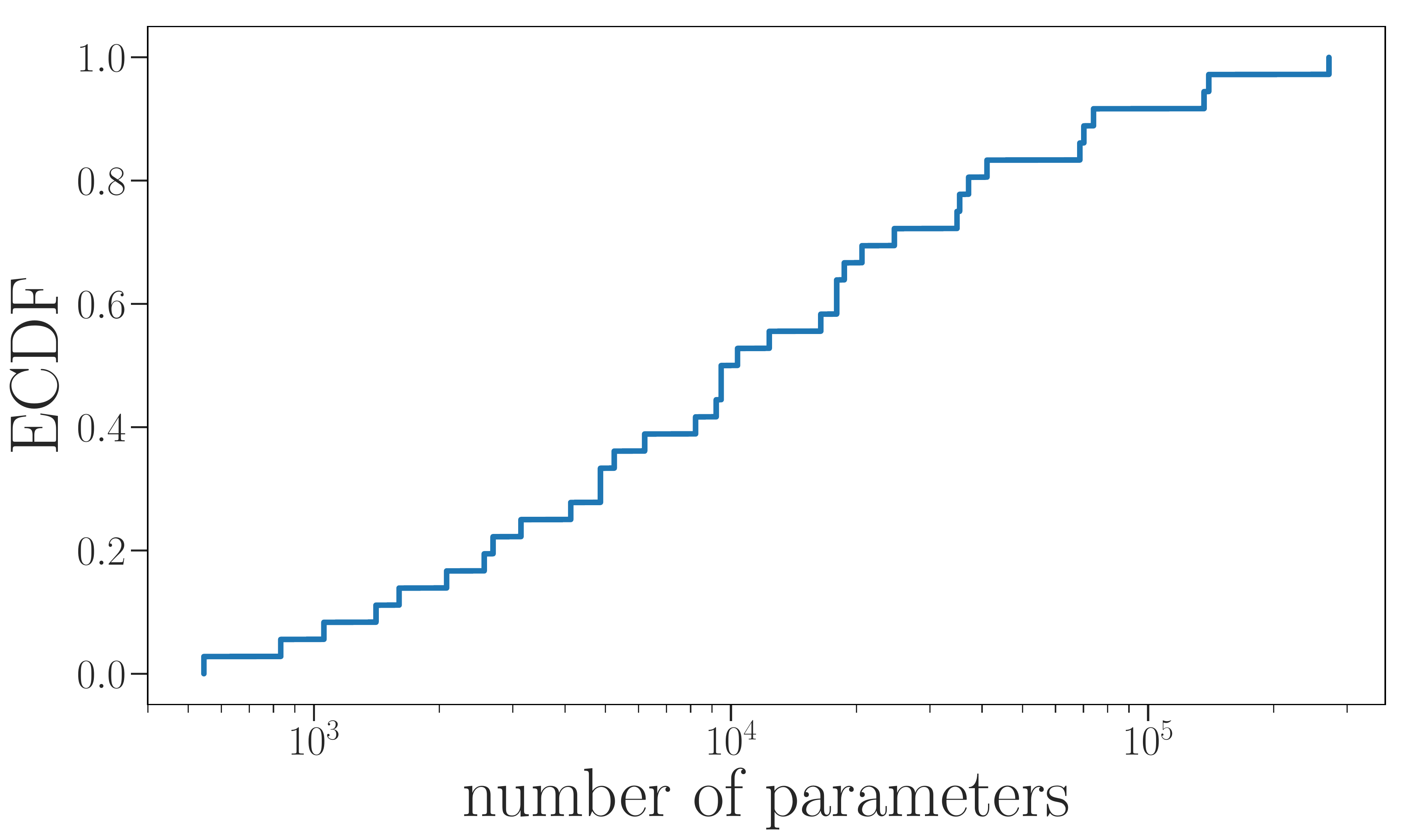}
 \includegraphics[width=0.24\linewidth]{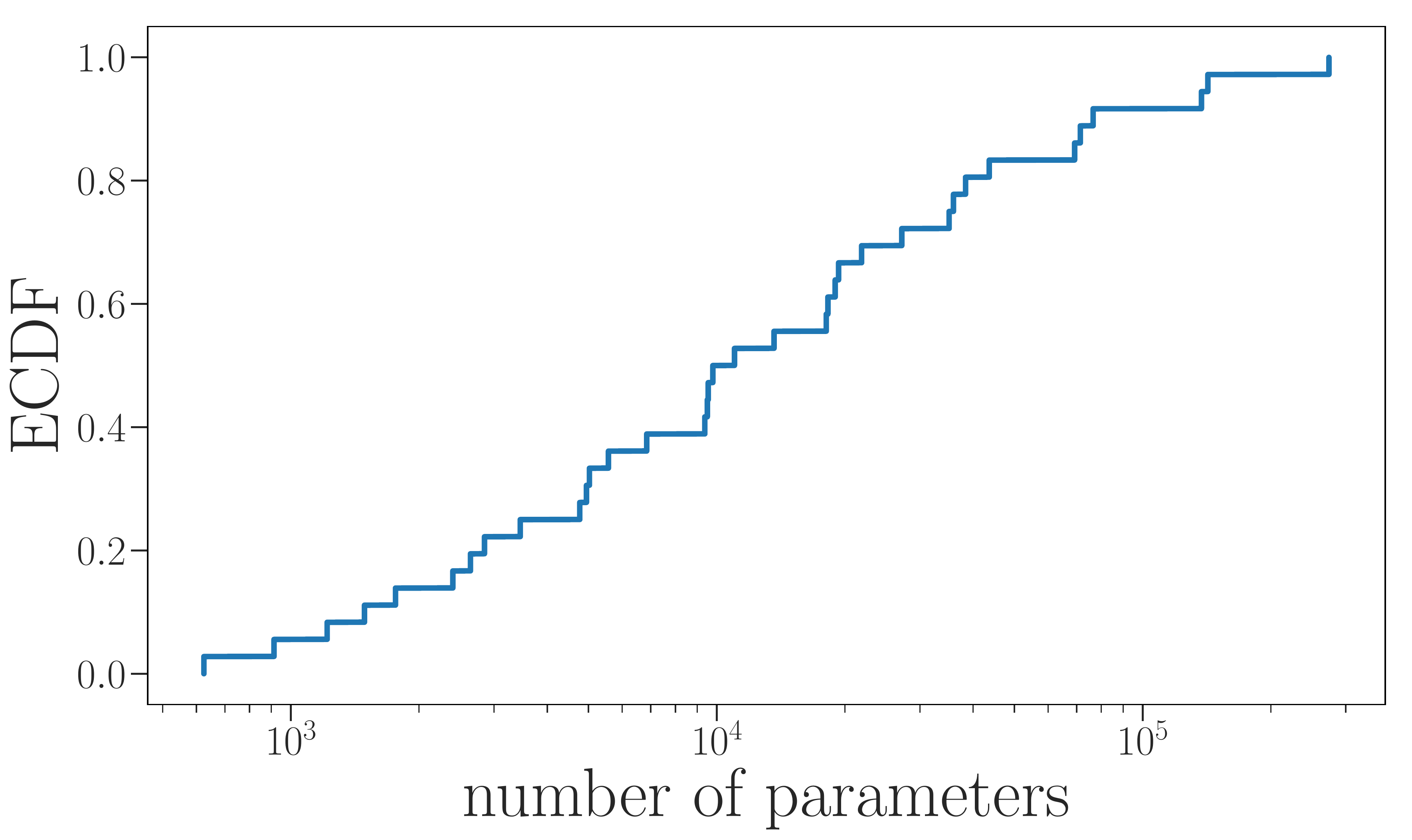}

 \caption[Empirical cumulative distributions params]{The empirical cumulative distribution (ECDF) of the number of parameters, computed on \hpobench-Protein (left), \hpobench-Slice (left middle), \hpobench-Naval (right middle) and \hpobench-Parkinson (right).}
 \label{fig:ecdf_params_all}
\end{center}
\end{figure}

\begin{figure}[h!]
\begin{center}
 \includegraphics[width=0.24\linewidth]{plots/cdf_runtime_protein_structure.pdf}
 \includegraphics[width=0.24\linewidth]{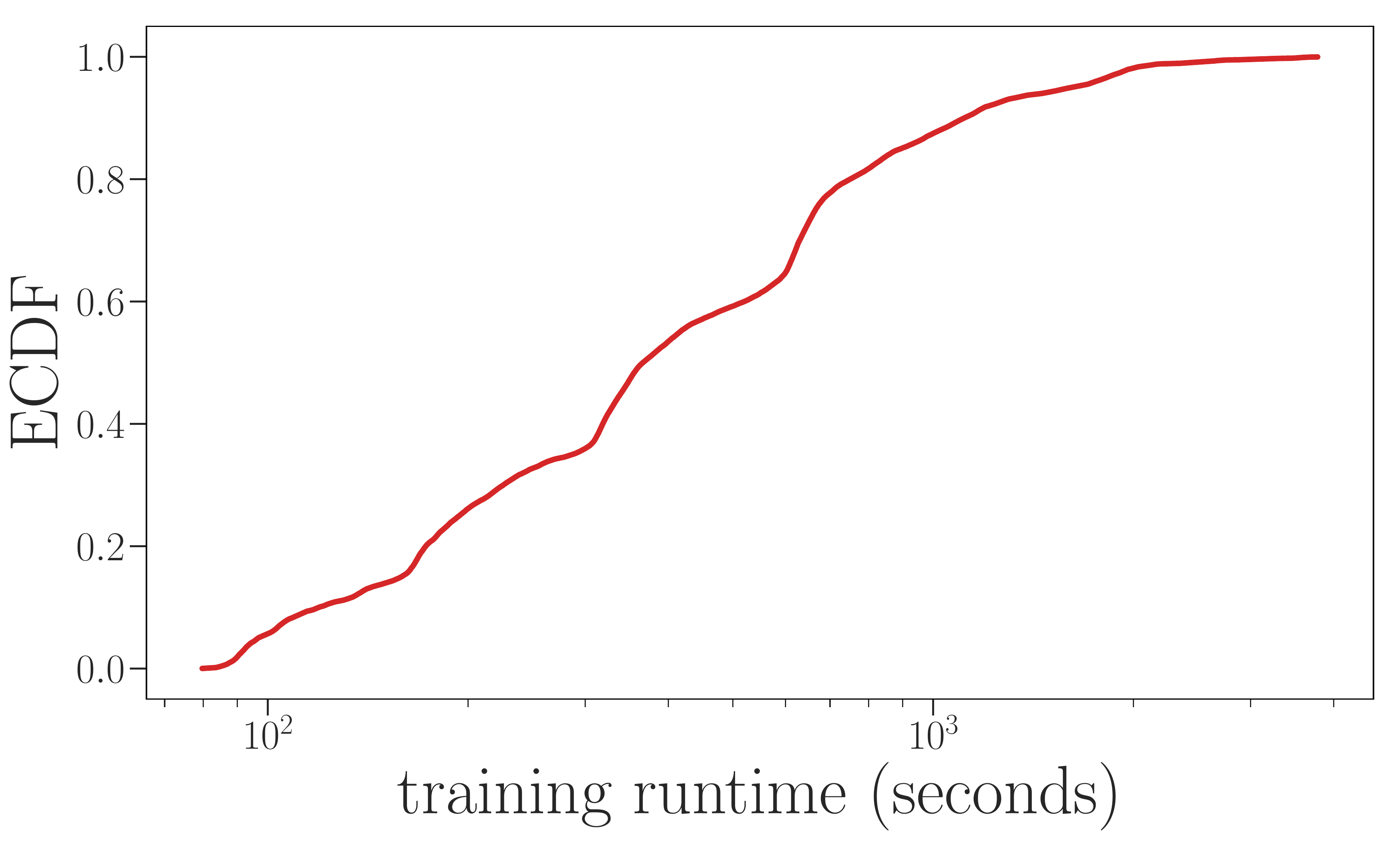}
 \includegraphics[width=0.24\linewidth]{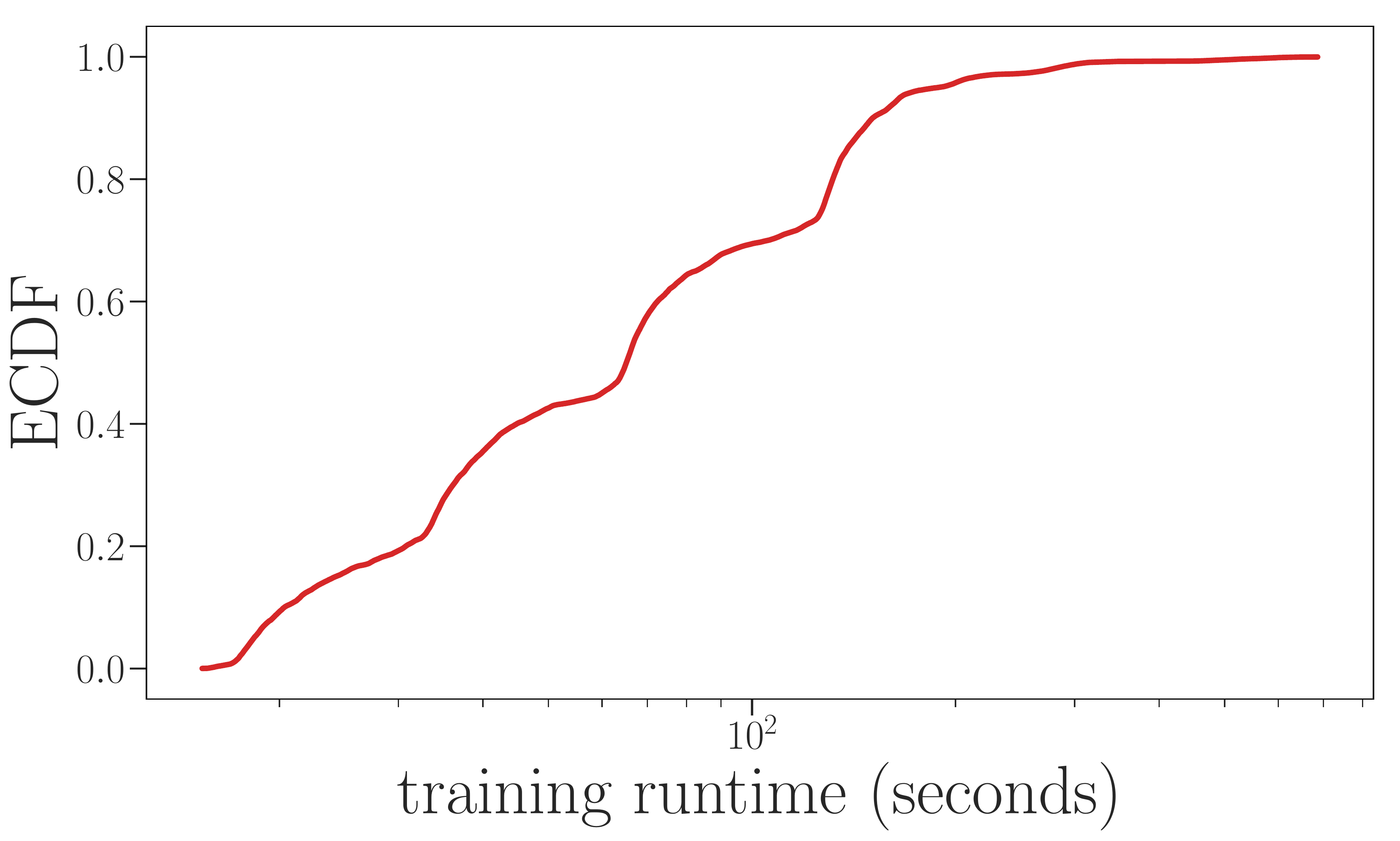}
 \includegraphics[width=0.24\linewidth]{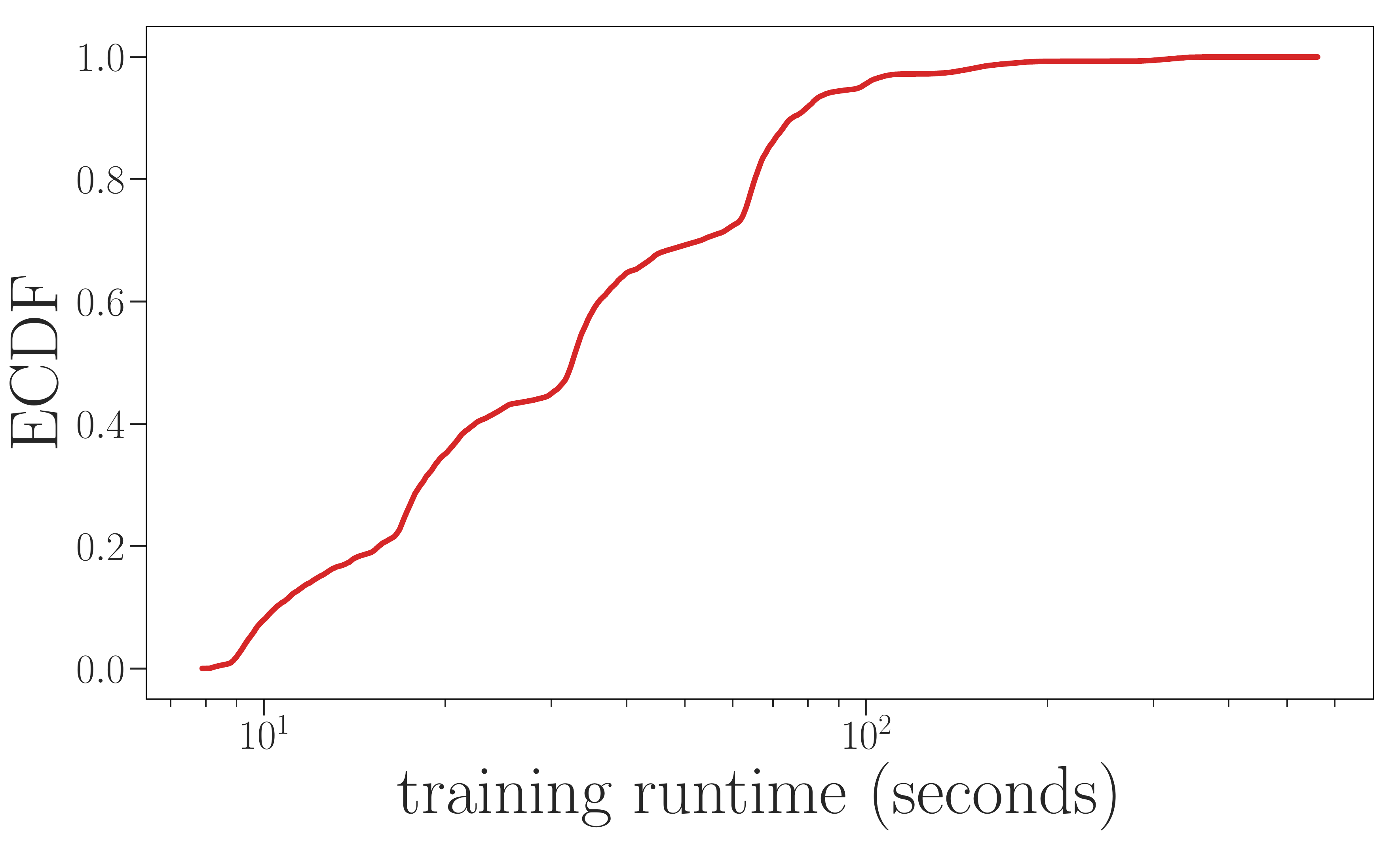}

 \caption[Empirical cumulative distributions time]{The empirical cumulative distribution (ECDF) of the training runtime, computed on \hpobench-Protein (left), \hpobench-Slice (left middle), \hpobench-Naval (right middle) and \hpobench-Parkinson (right).}
 \label{fig:ecdf_time_all}
\end{center}
\end{figure}

\begin{figure}[h!]
\begin{center}
 \includegraphics[width=0.24\linewidth]{plots/cdf_noise_protein_structure.pdf}
 \includegraphics[width=0.24\linewidth]{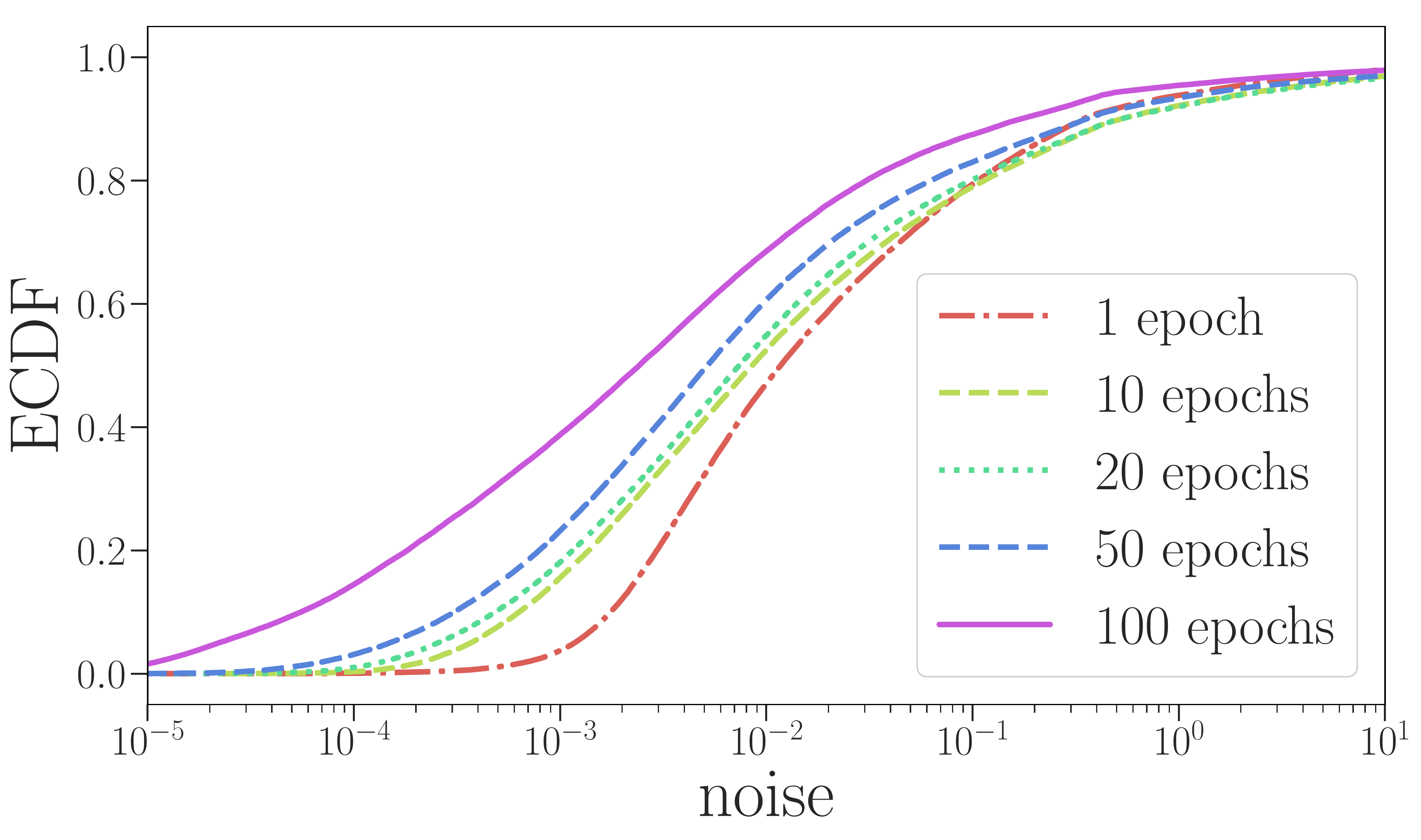}
 \includegraphics[width=0.24\linewidth]{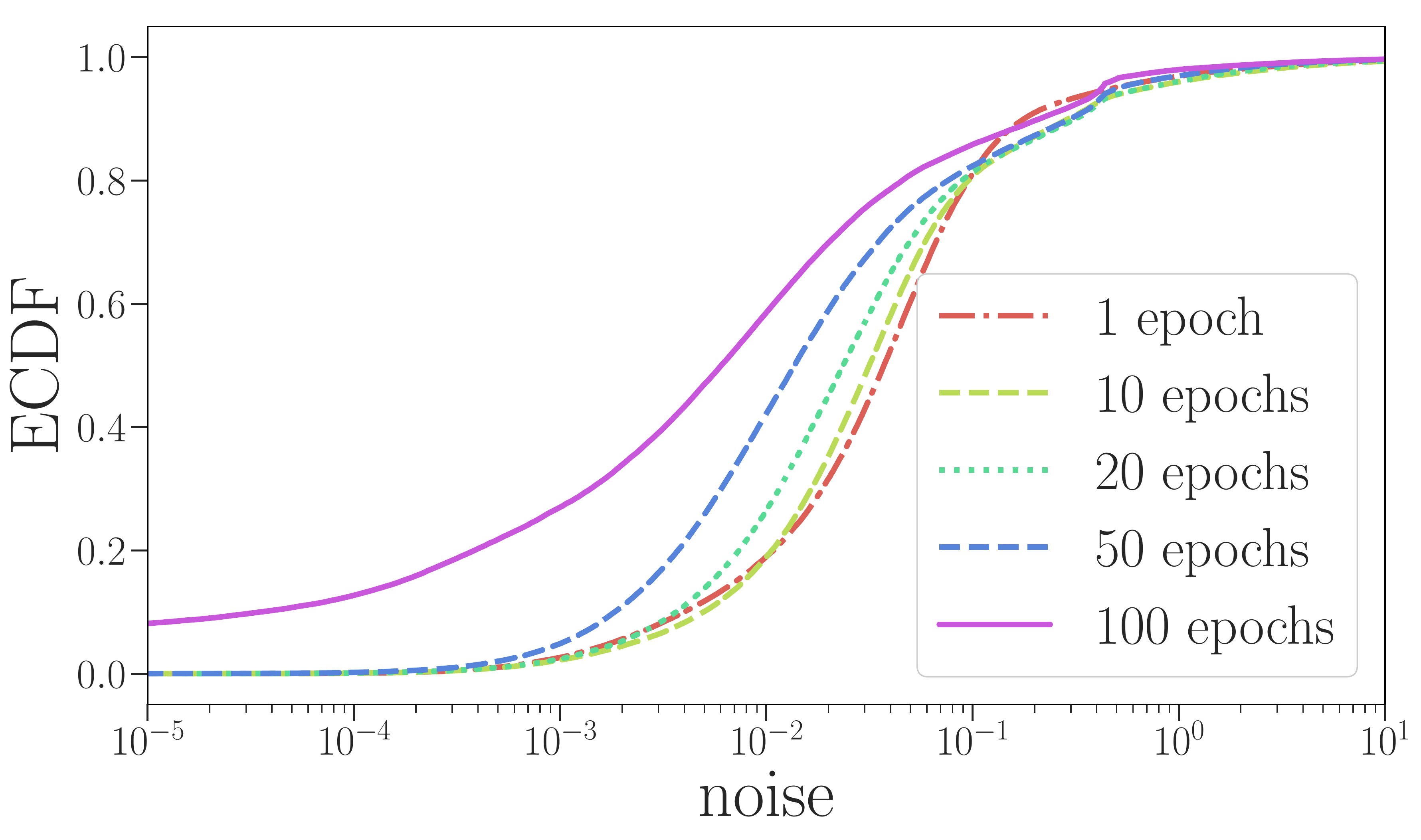}
 \includegraphics[width=0.24\linewidth]{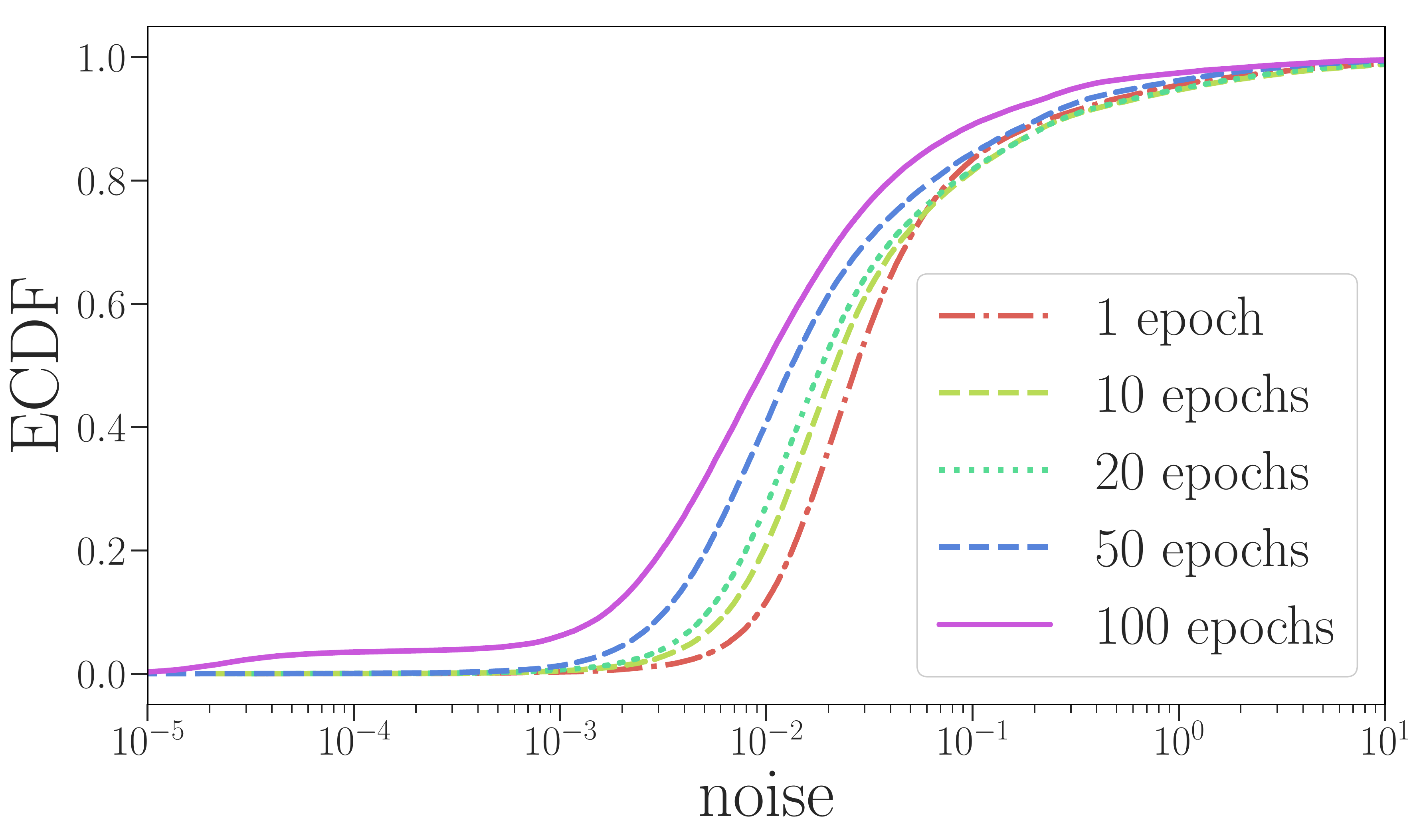}

 \caption[Empirical cumulative distributions noise]{The empirical cumulative distribution (ECDF) of the noise across the 4 repeated training processes for each configuration, computed on \hpobench-Protein (left), \hpobench-Slice (left middle), \hpobench-Naval (right middle) and \hpobench-Parkinson (right).}
 \label{fig:ecdf_noise_all}
\end{center}
\end{figure}

We now show the empirical cumulative distribution (ECDF) of all four datasets for: the mean squared error for training, validation and test (Figure~\ref{fig:ecdf_mse_all}), the number of parameters (Figure~\ref{fig:ecdf_params_all}), the measured wall-clock time for training (Figure~\ref{fig:ecdf_time_all}) and the noise, defined as standard deviation between the individual trials of each configuration (Figure~\ref{fig:ecdf_noise_all}) .
Note that we computed the ECDF of the mean squared error and the runtime based on the average over the four trials.

\begin{figure}[h!]
\begin{center}
 \includegraphics[width=0.24\linewidth]{plots/rank_correlation_protein_structure_all.pdf}
 \includegraphics[width=0.24\linewidth]{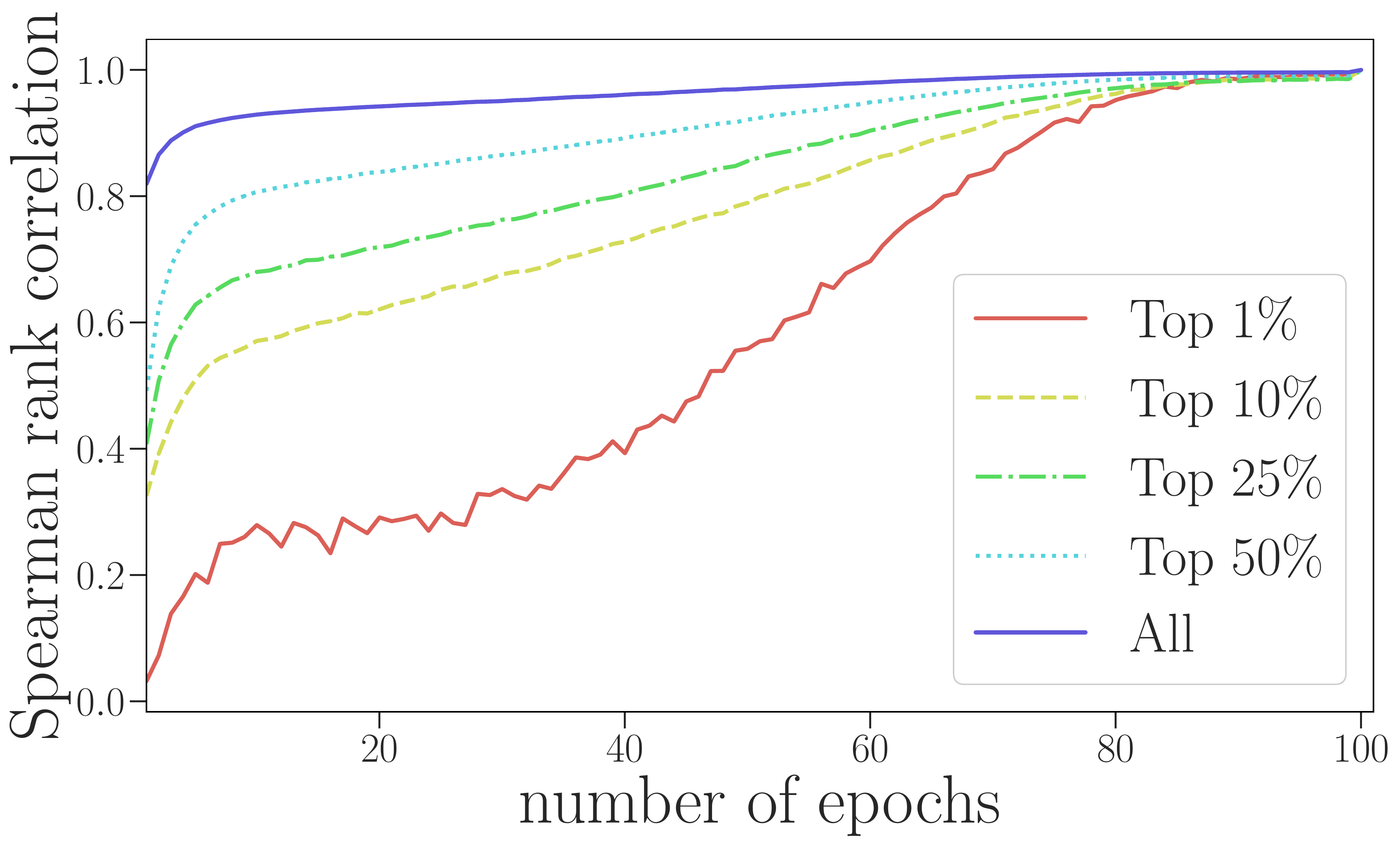}
 \includegraphics[width=0.24\linewidth]{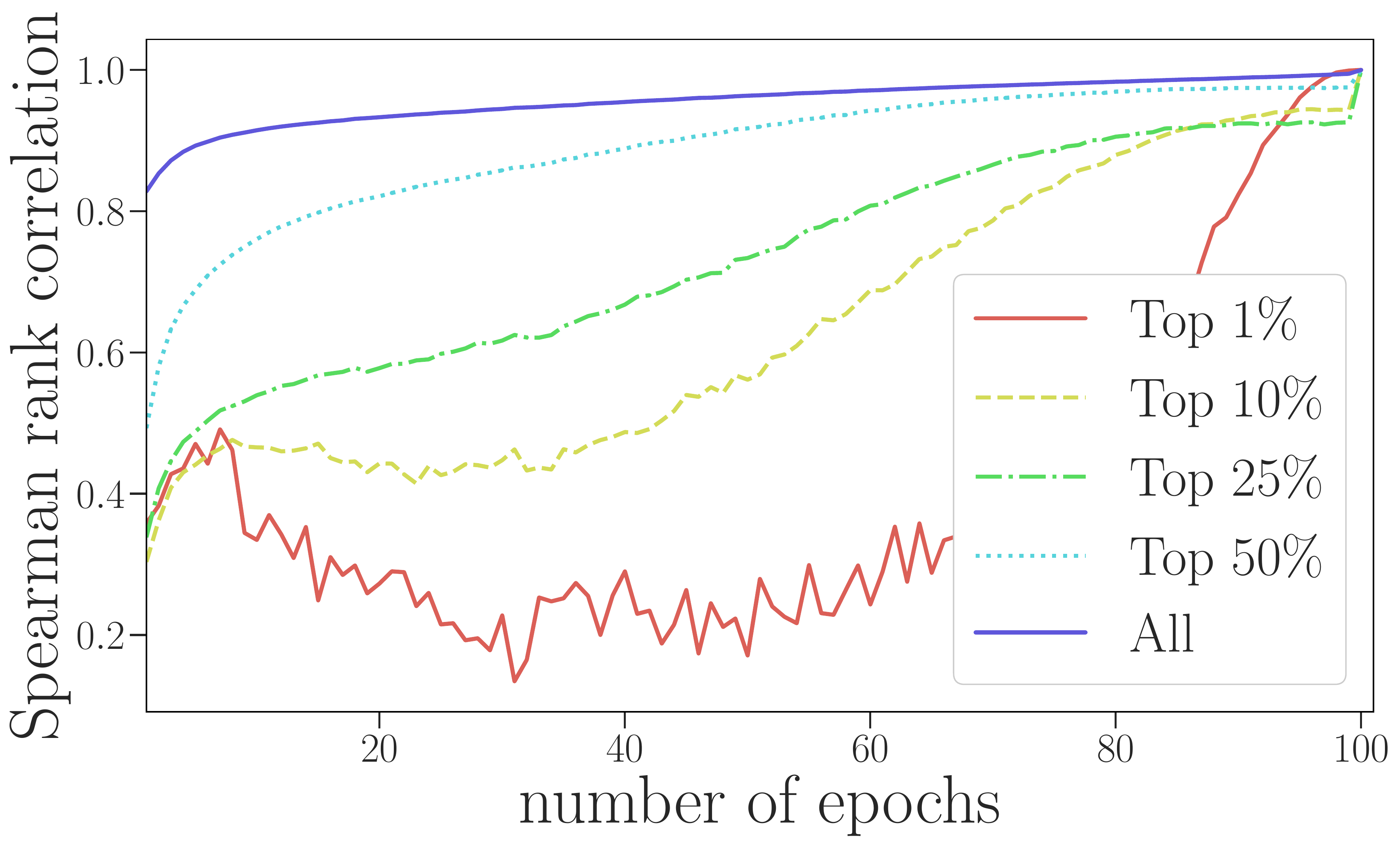}
 \includegraphics[width=0.24\linewidth]{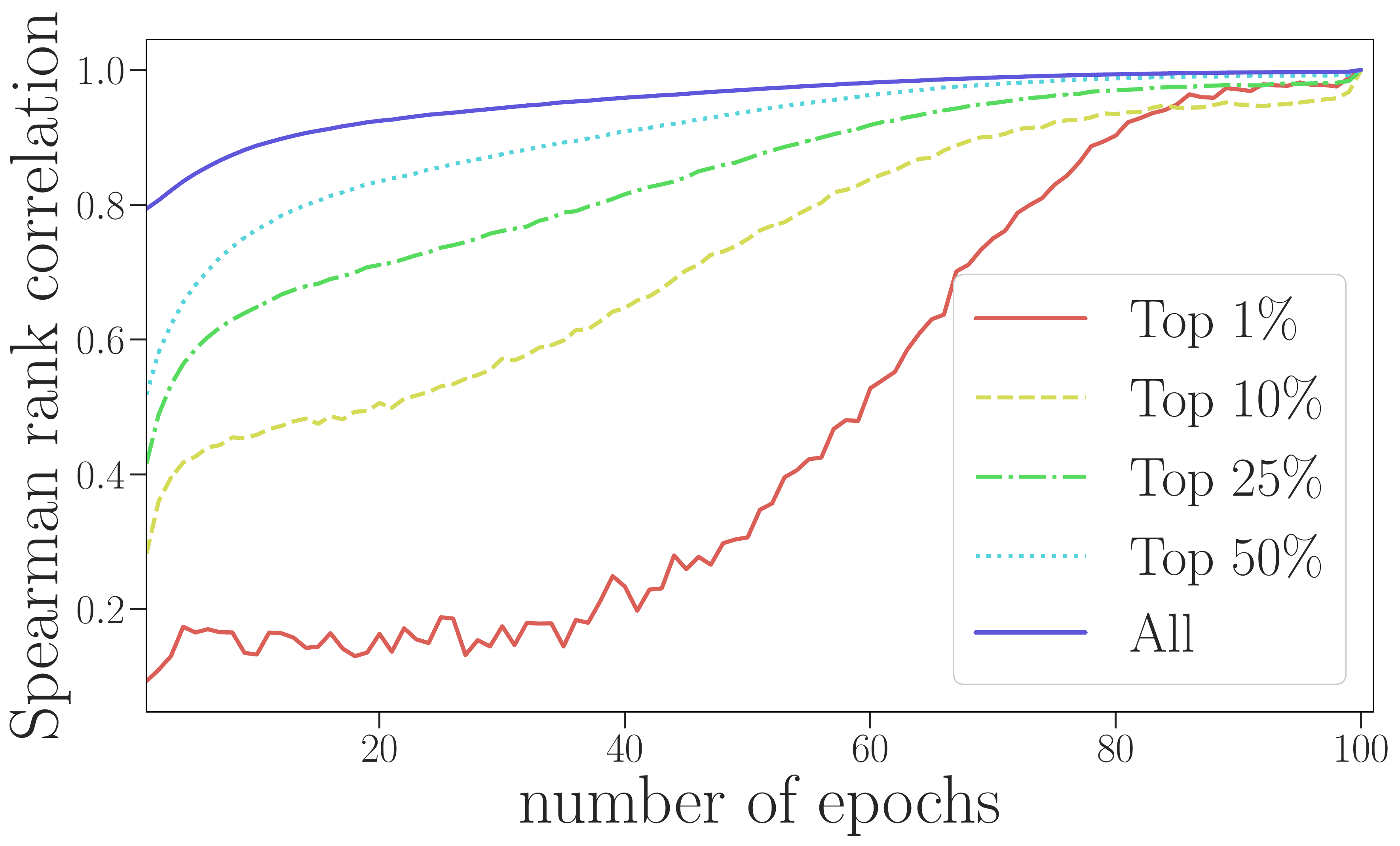}
 \caption[Rank correlation across budgets]{The Spearman rank correlation between different number of epochs for the \hpobench-Protein (left), \hpobench-Slice (left middle), \hpobench-Naval (right middle) and \hpobench-Parkinson (right) when we consider all configurations or only the top $1\%$, $10\%$, $20\%$, and $50\%$ of all configurations based on their test error.}
 \label{fig:rank_correlation_all}
\end{center}
\end{figure}

  Figure~\ref{fig:rank_correlation_all} shows the Spearman rank correlation between the performance of a hyperparameter configuration after training for the final budget of 100 epochs and its performance after training for the corresponding number of epochs on the x-axis.
  We also show the correlation if only the top $1\%, 10\%, 25\%$ and $50\%$ configurations are taken into account.

\section{Hyperparameter Importance}\label{sec:supp_hpobench_importance}

\begin{figure}[t]
\centering
\includegraphics[width=0.32\linewidth]{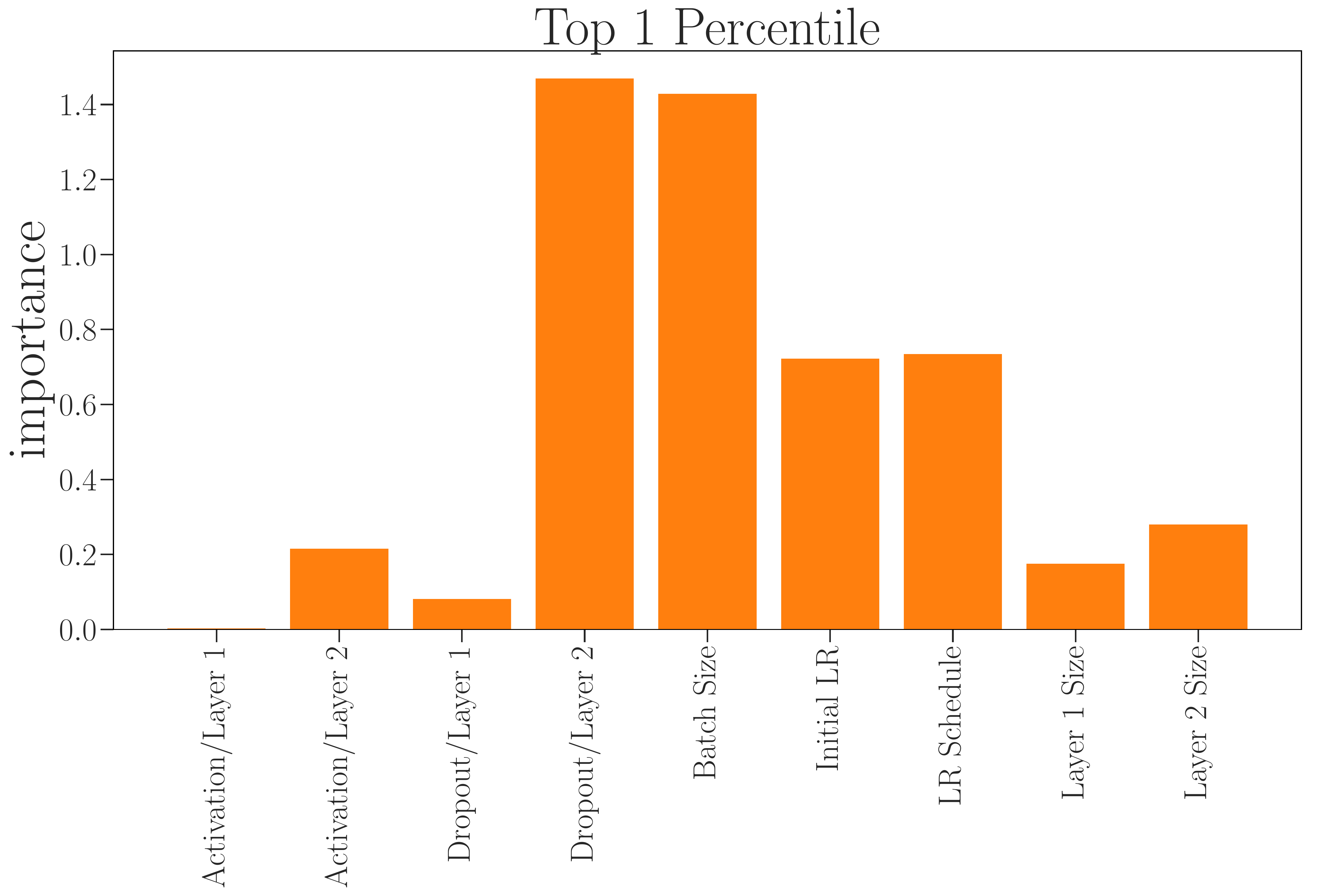}
\includegraphics[width=0.32\linewidth]{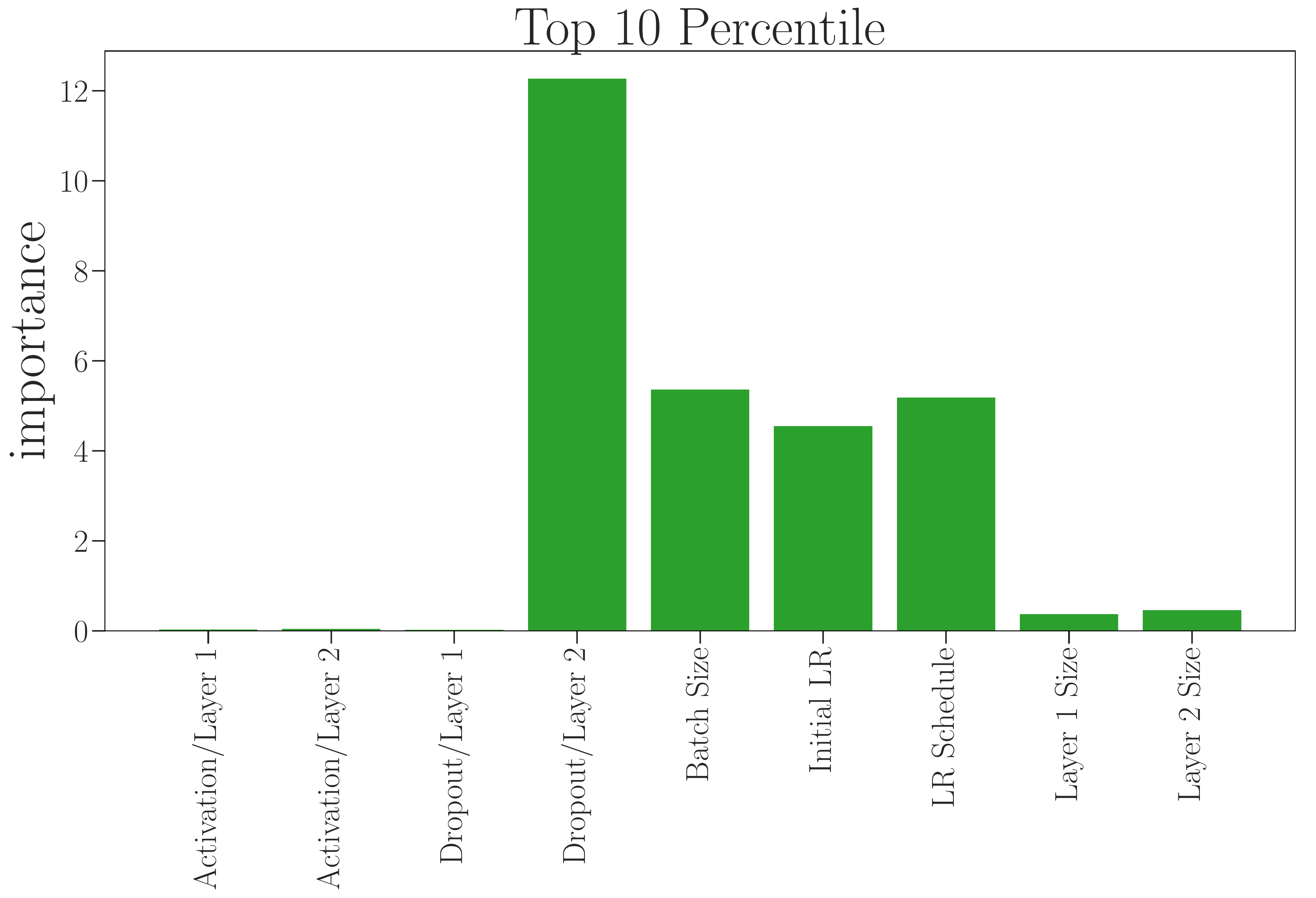}
\includegraphics[width=0.32\linewidth]{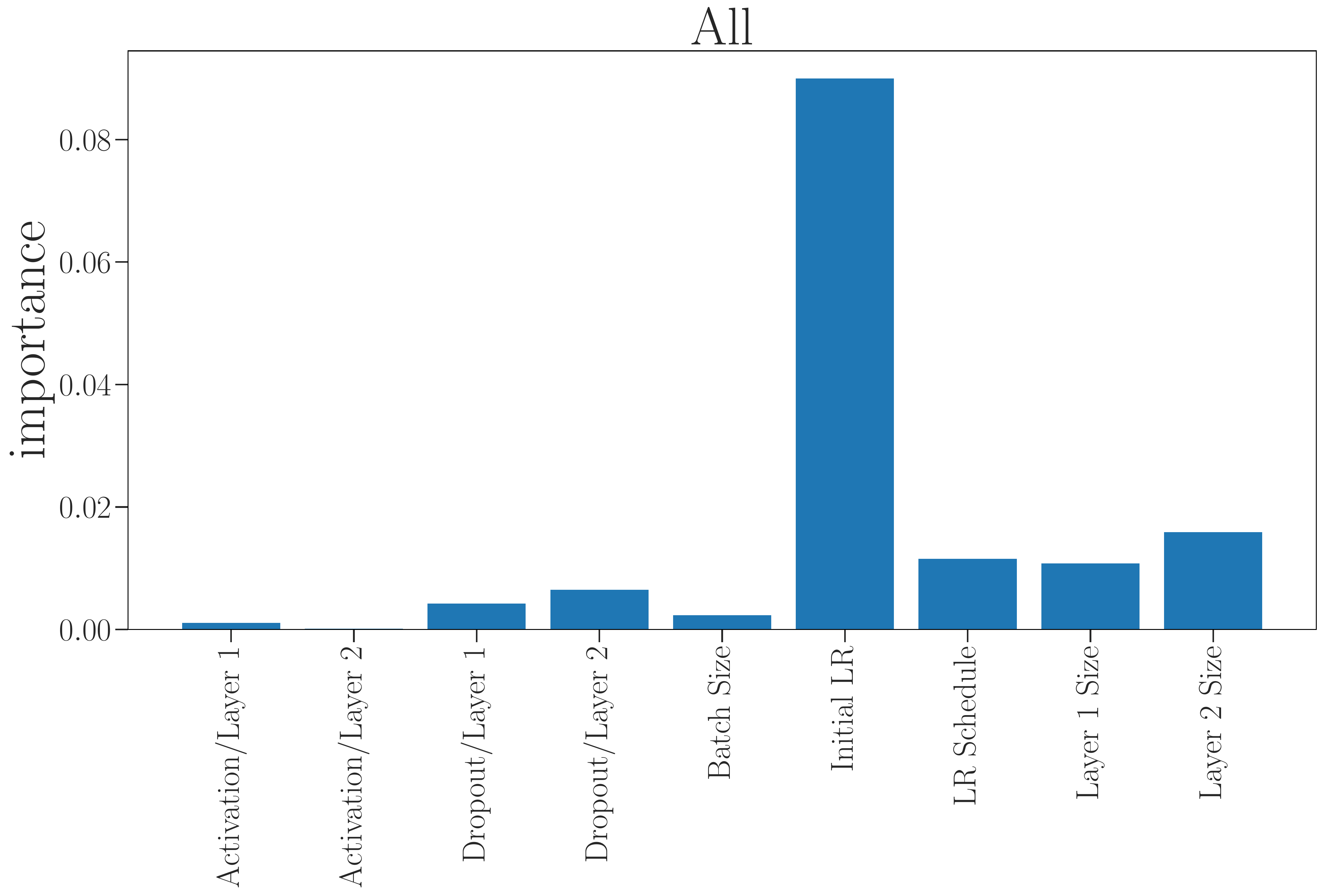}\\
\includegraphics[width=0.32\linewidth,valign=t]{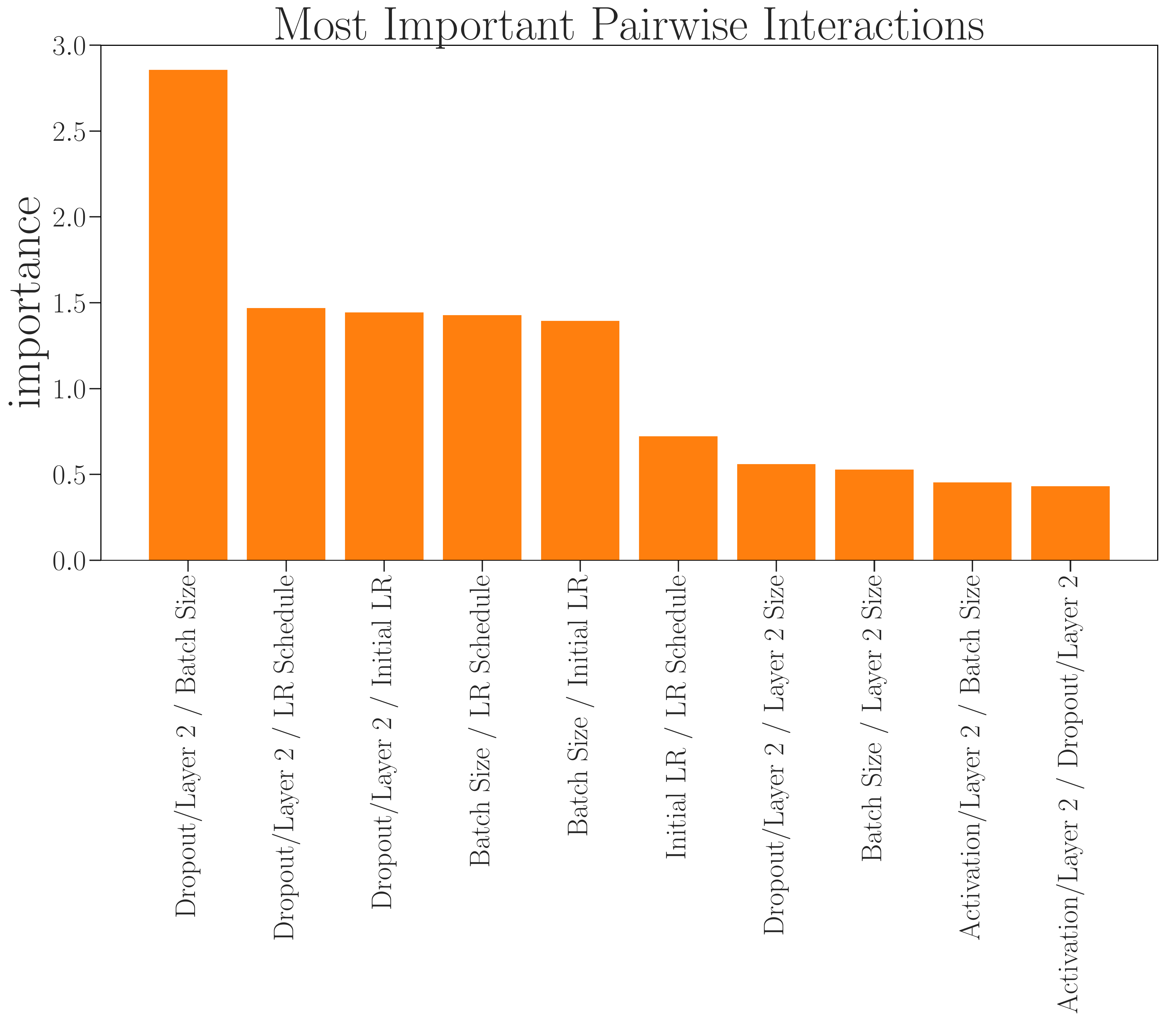}
\includegraphics[width=0.32\linewidth,valign=t]{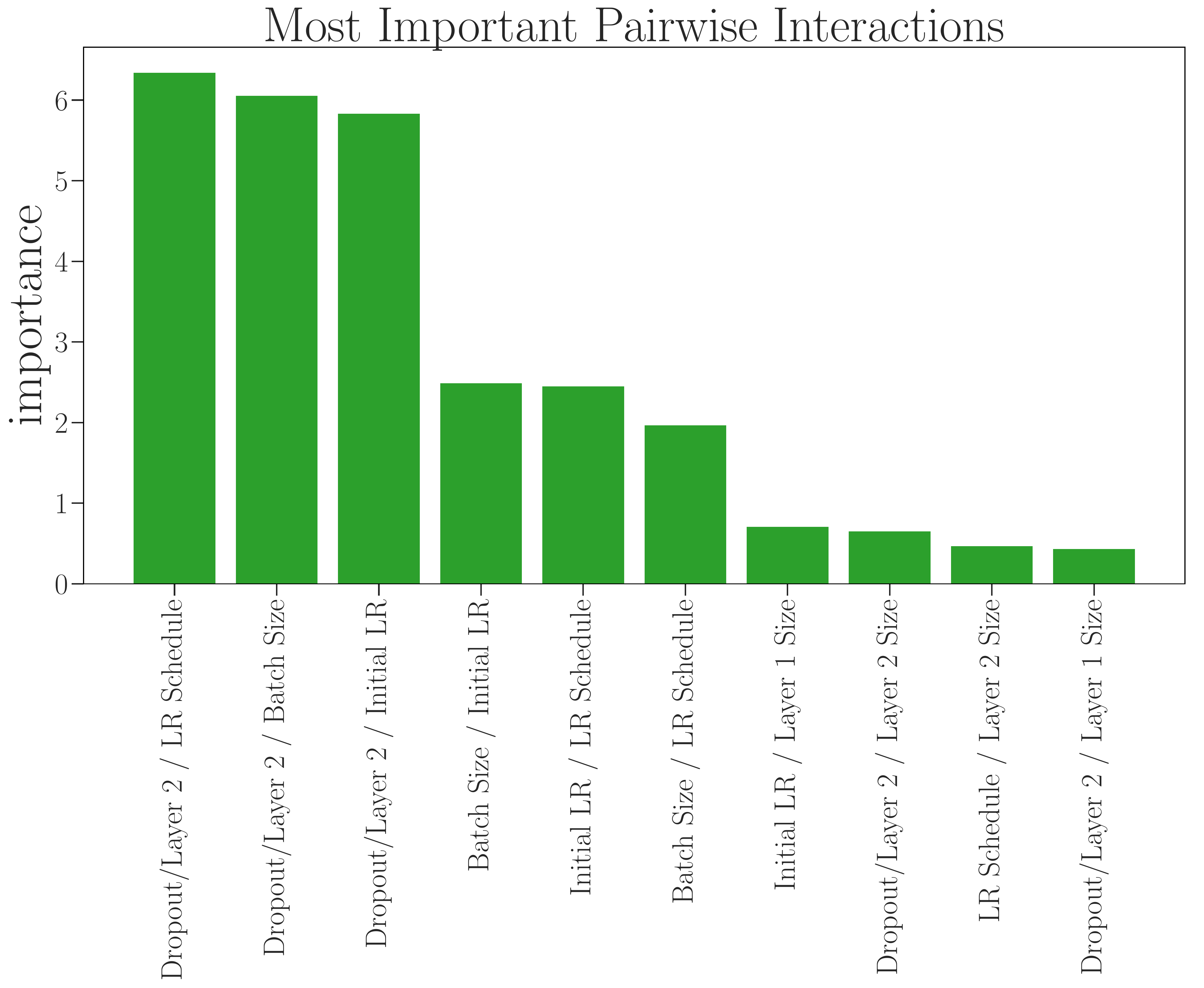}
\includegraphics[width=0.32\linewidth,valign=t]{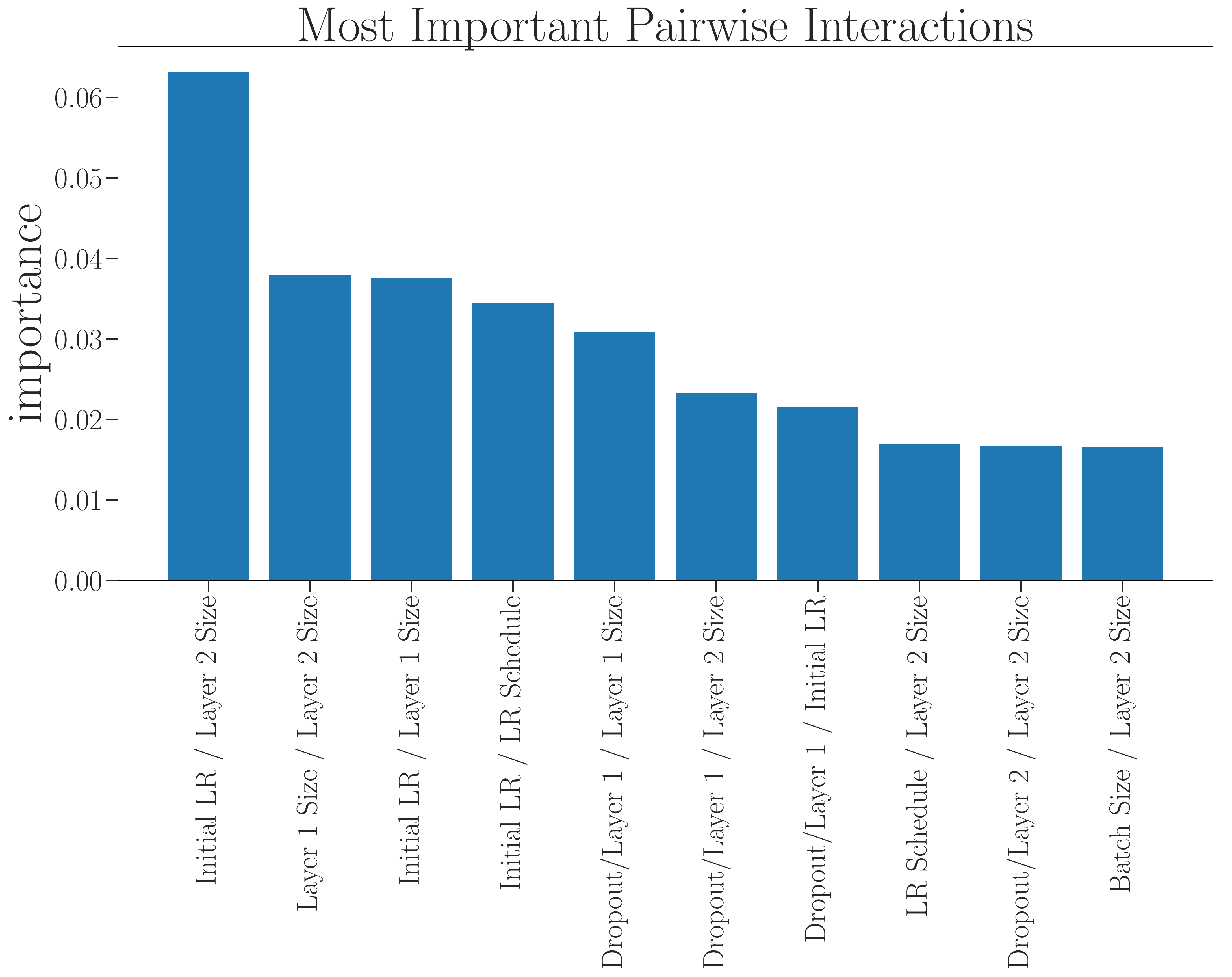}
\caption[fANOVA HPOBench]{HPOBench-Naval. Top row: Importance of the different hyperparameter based on the fANOVA for: (left) only the top $1\%$ ; (middle) top $10\%$ ; (right) all configurations. Bottom row: most important hyperparameter pairs with (left) only the top $1\%$ ; (middle) top $10\%$ ; (right) all configurations.}
\label{fig:importance_naval}
\end{figure}

\begin{figure}[t]
\centering
\includegraphics[width=0.32\linewidth]{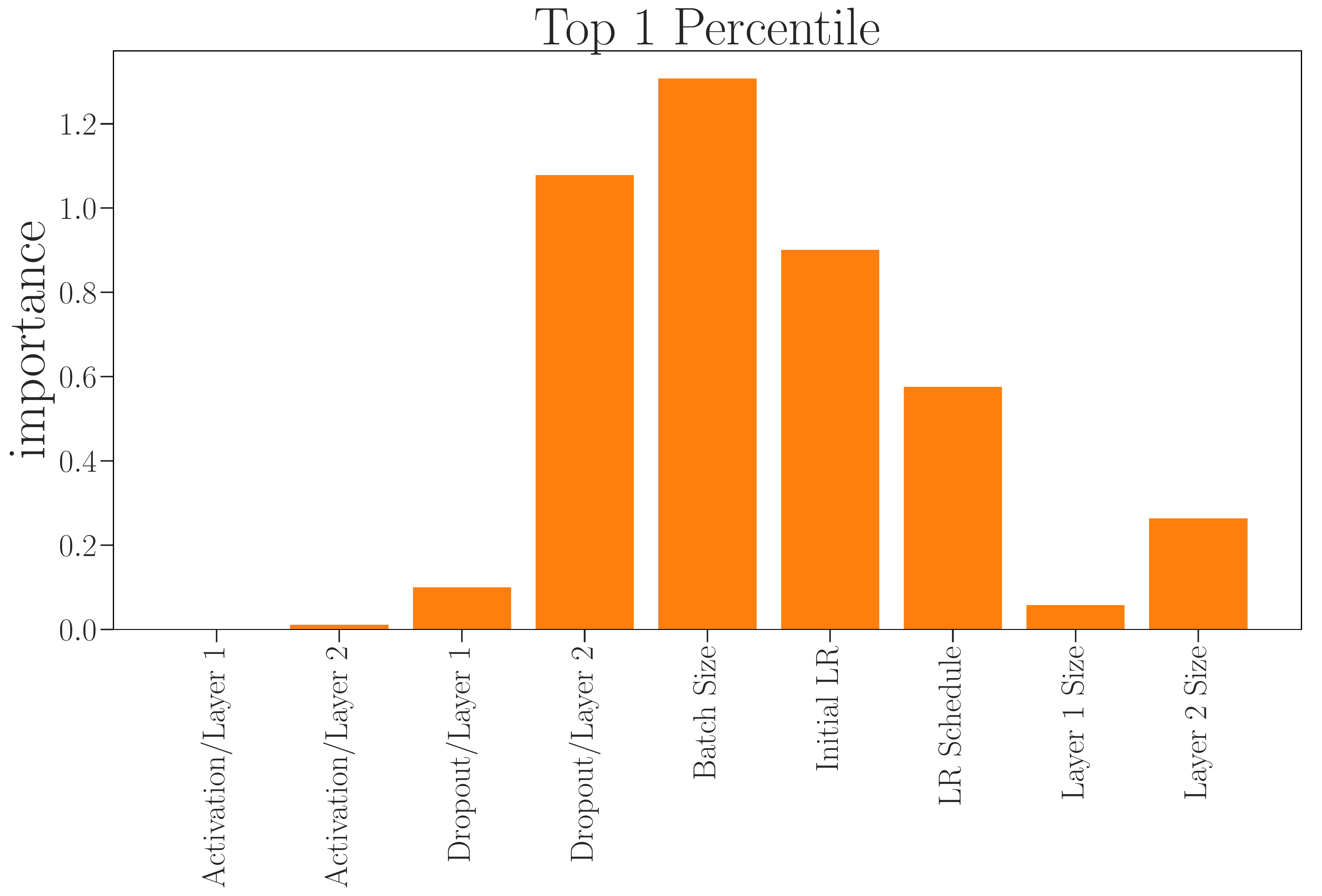}
\includegraphics[width=0.32\linewidth]{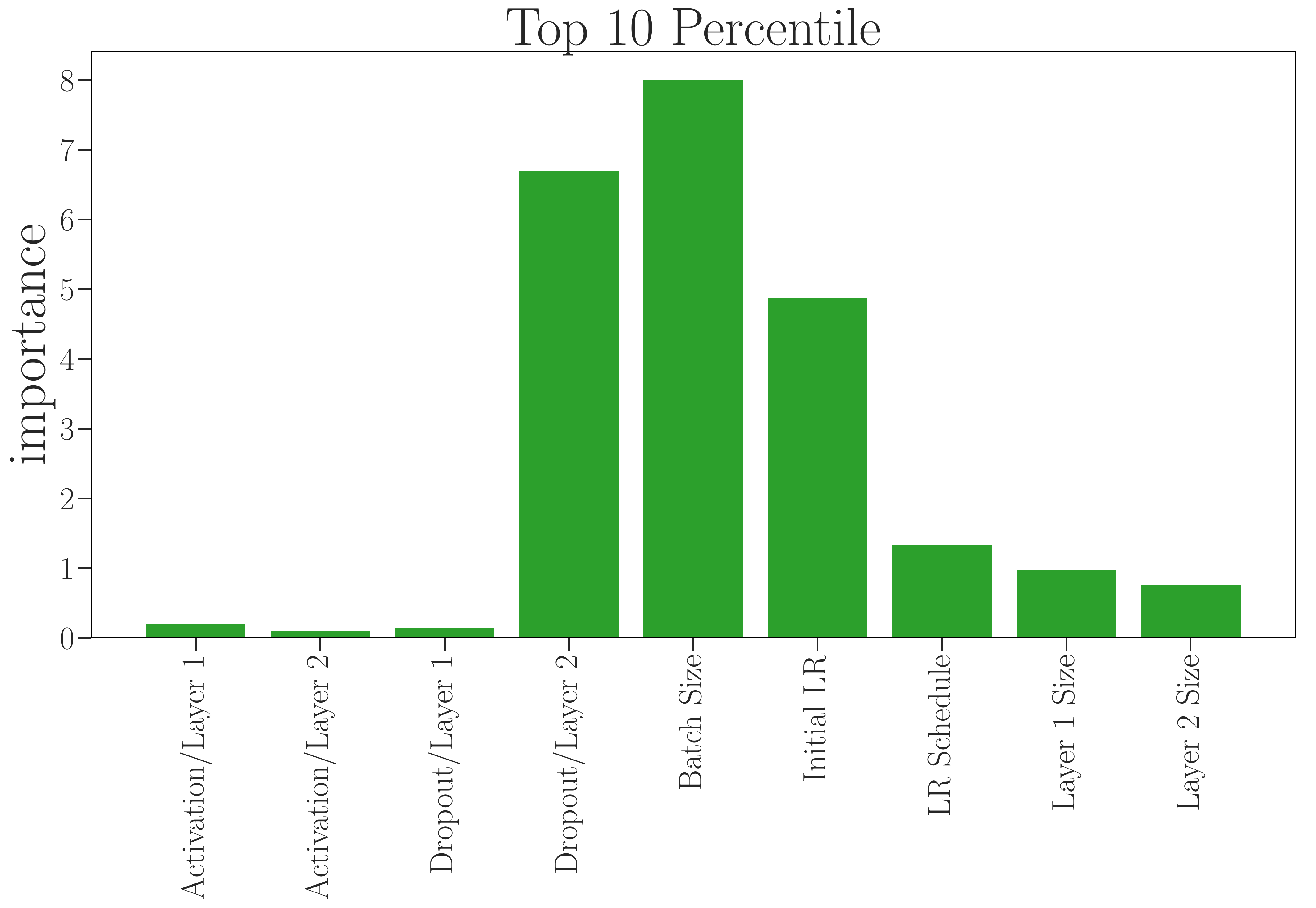}
\includegraphics[width=0.32\linewidth]{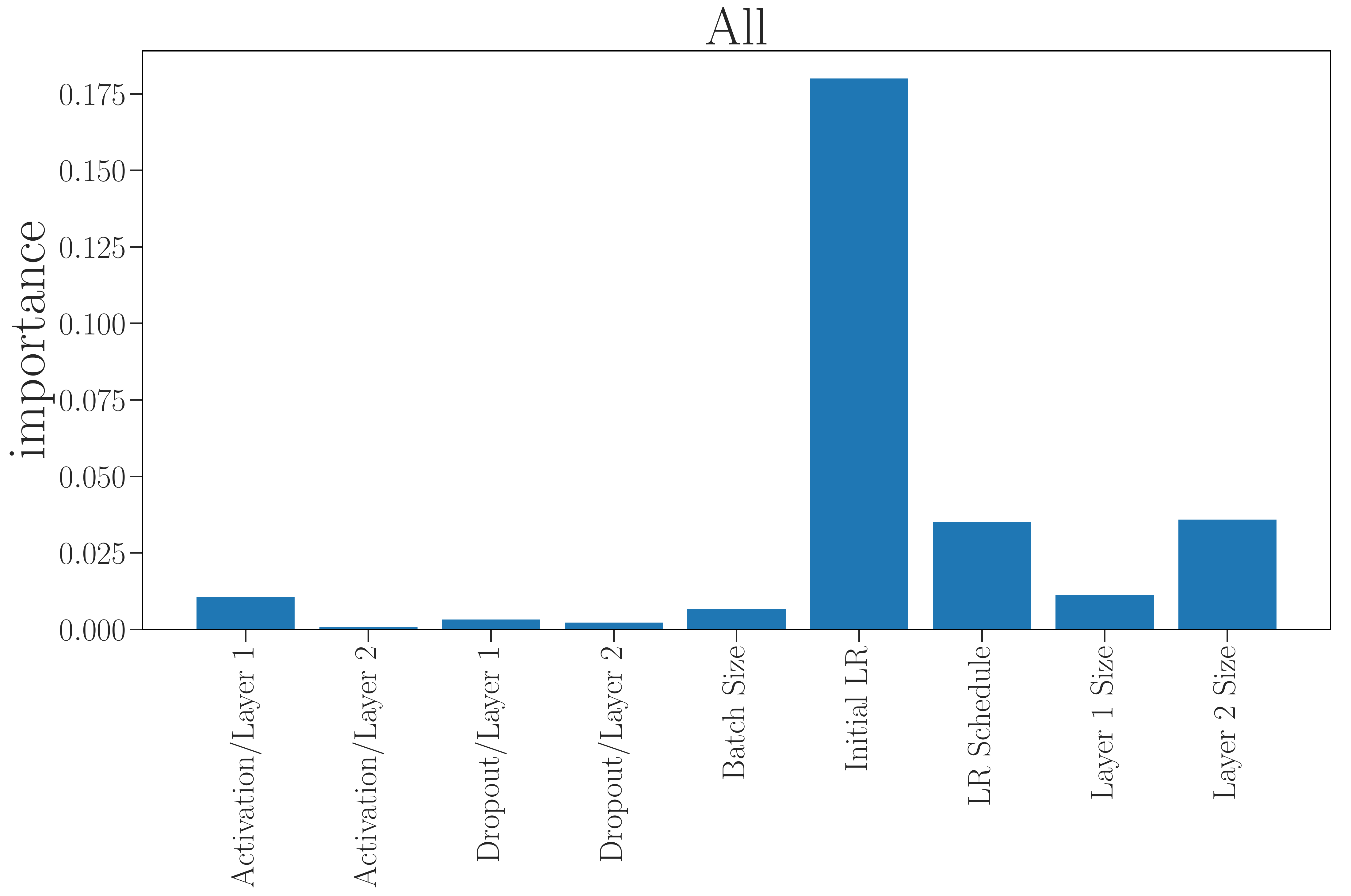}\\
\includegraphics[width=0.32\linewidth,valign=t]{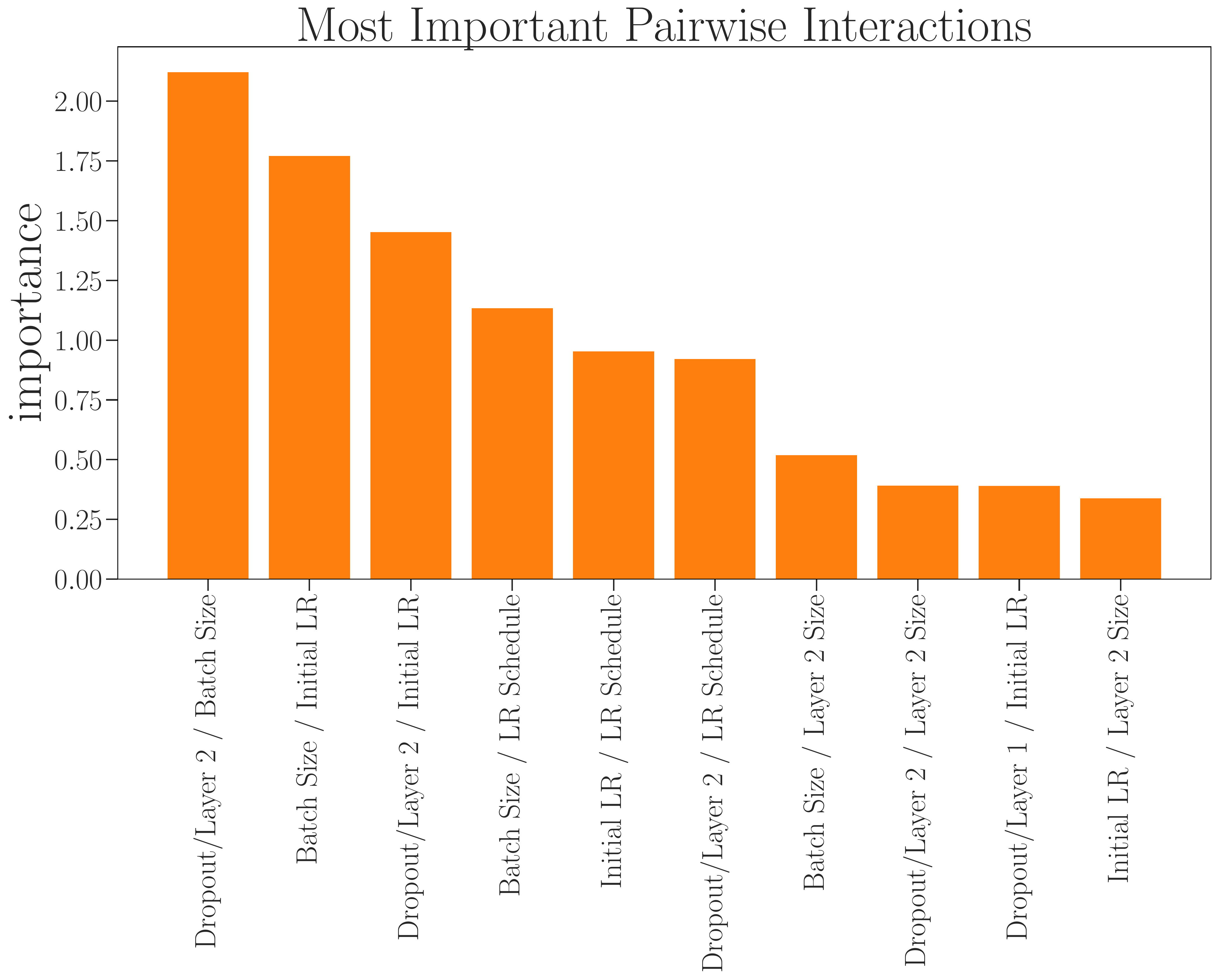}
\includegraphics[width=0.32\linewidth,valign=t]{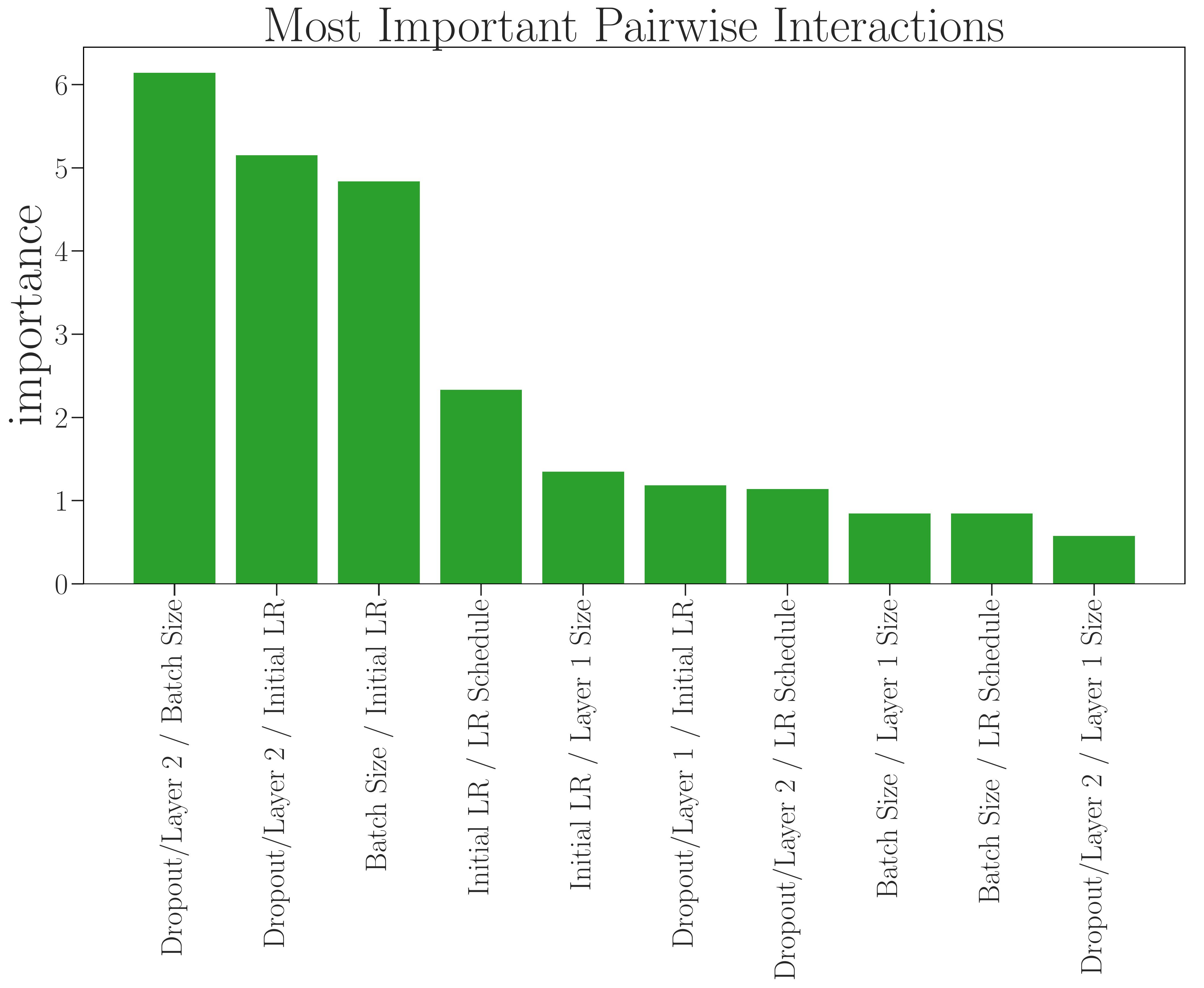}
\includegraphics[width=0.32\linewidth,valign=t]{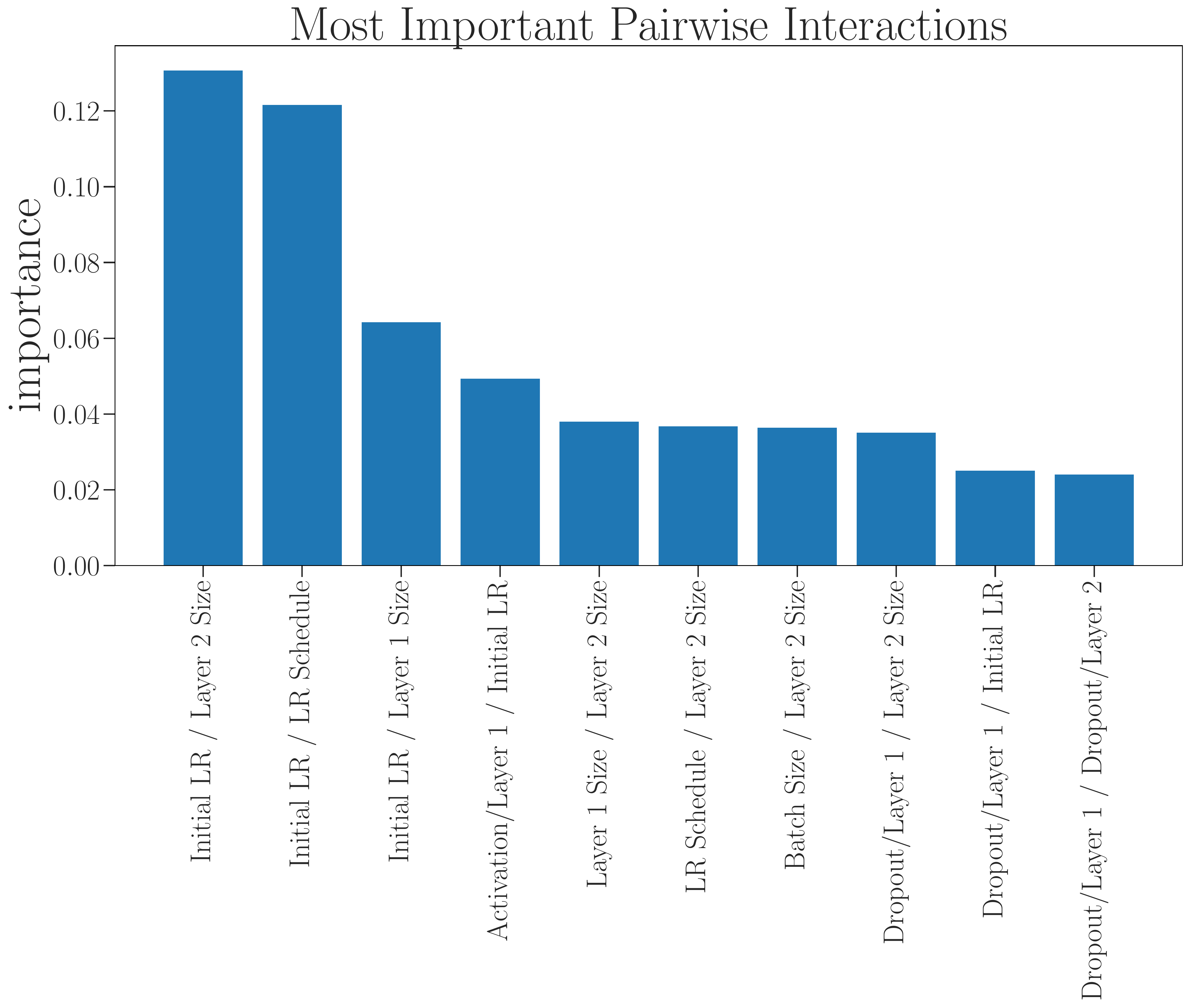}
\caption[fANOVA HPOBench]{HPOBench-Parkinson. Top row: Importance of the different hyperparameter based on the fANOVA for: (left) only the top $1\%$ ; (middle) top $10\%$ ; (right) all configurations. Bottom row: most important hyperparameter pairs with (left) only the top $1\%$ ; (middle) top $10\%$ ; (right) all configurations.}
\label{fig:importance_parkinson}
\end{figure}

\begin{figure}[t]
\centering
\includegraphics[width=0.32\linewidth]{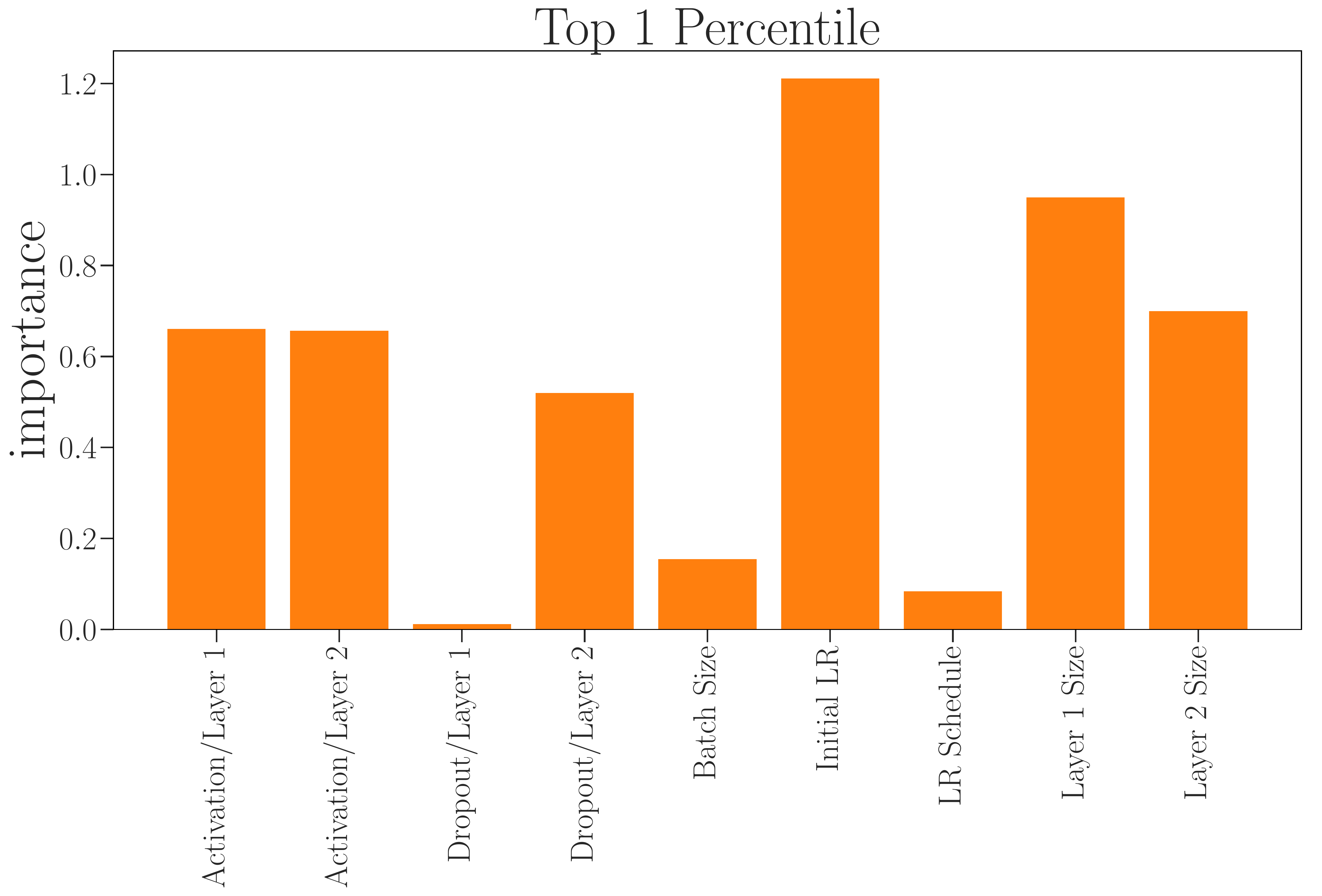}
\includegraphics[width=0.32\linewidth]{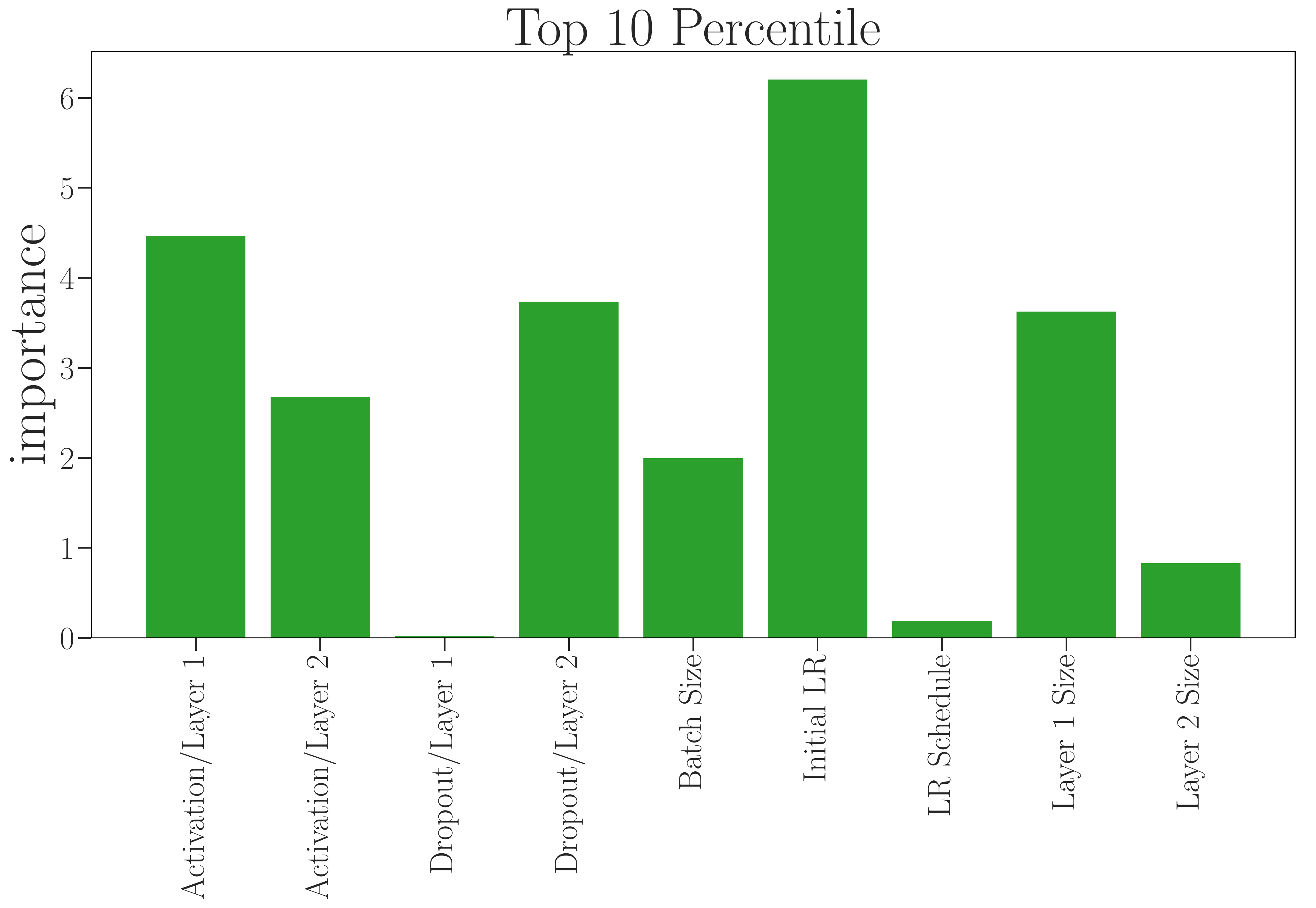}
\includegraphics[width=0.32\linewidth]{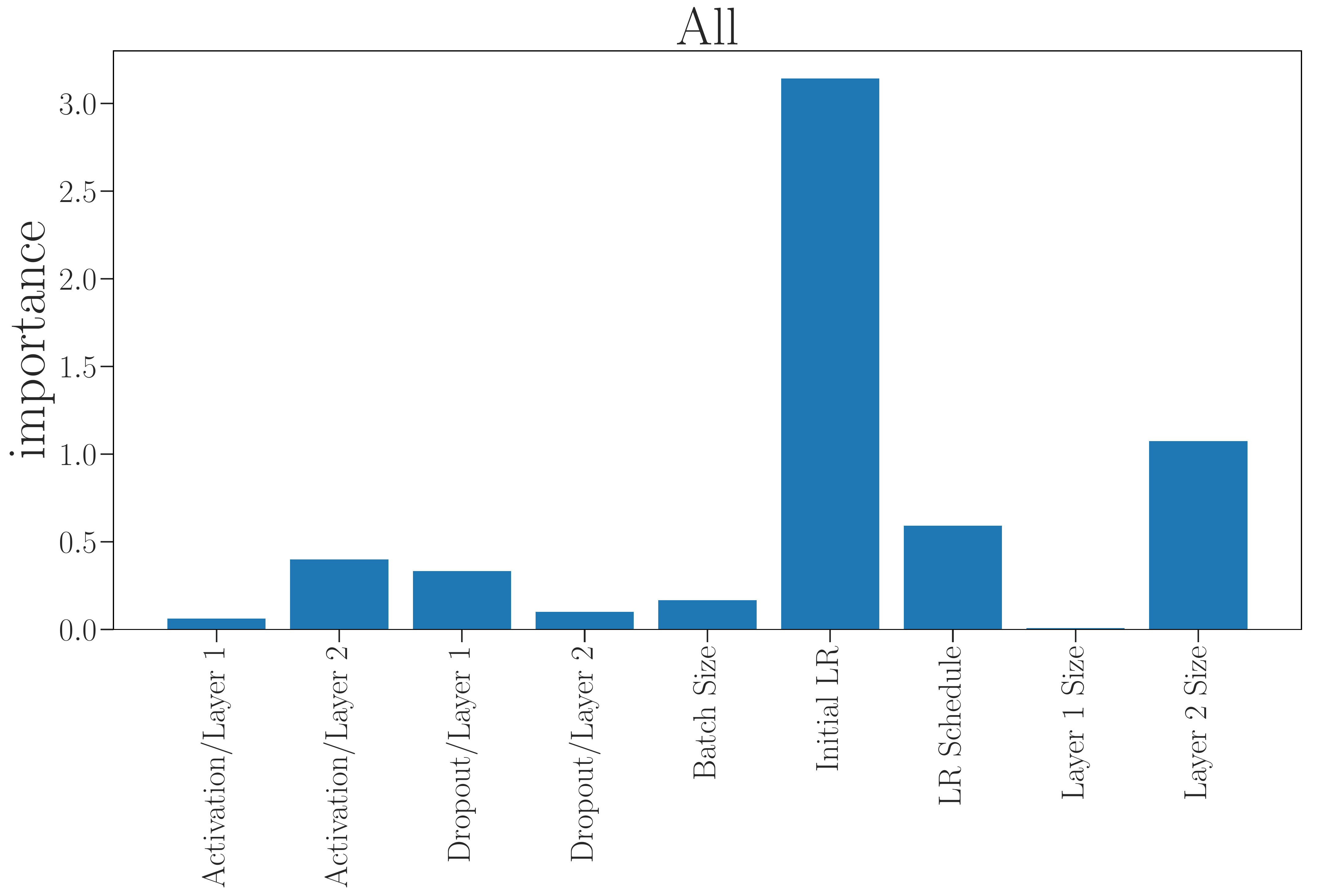}\\
\includegraphics[width=0.32\linewidth,valign=t]{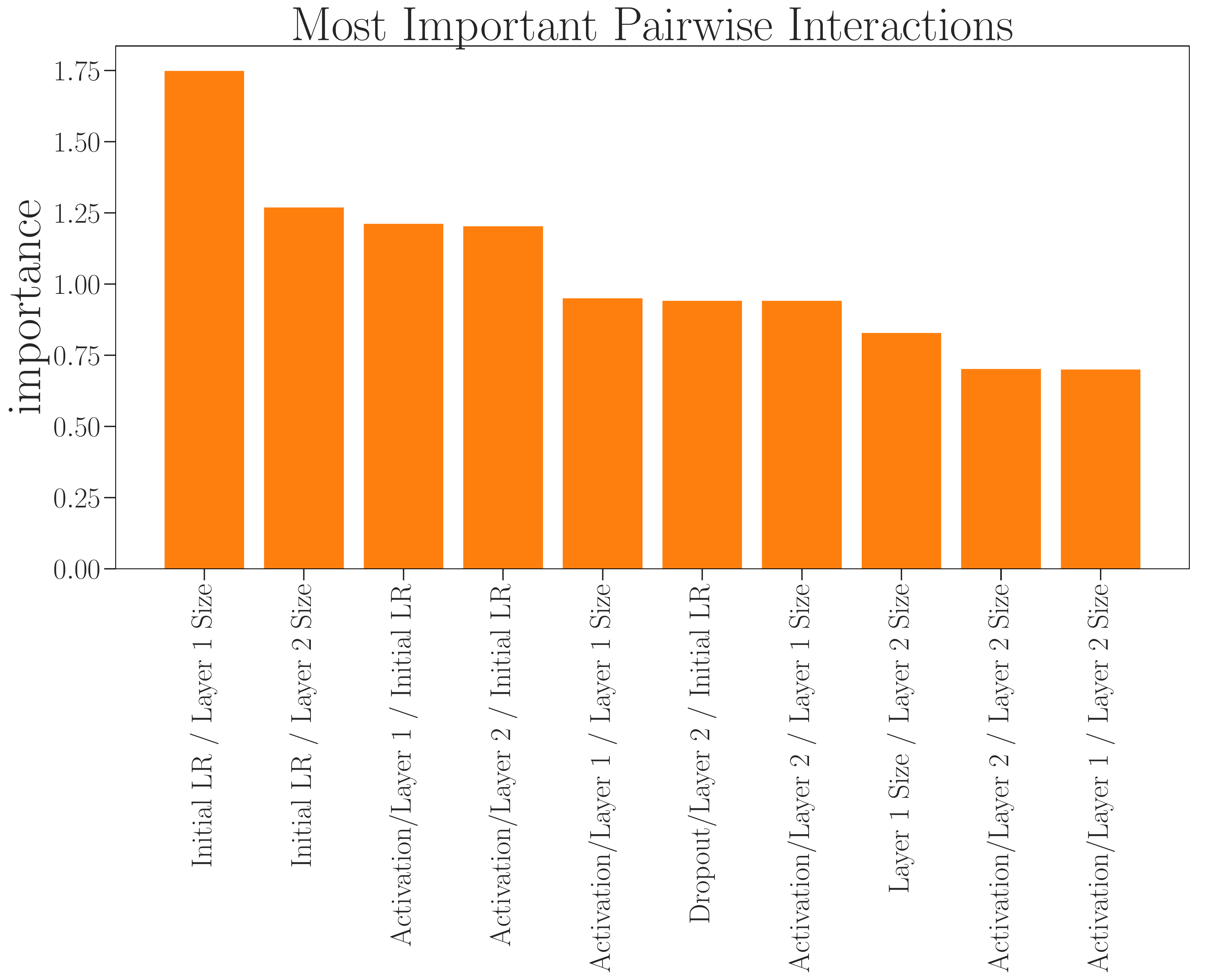}
\includegraphics[width=0.32\linewidth,valign=t]{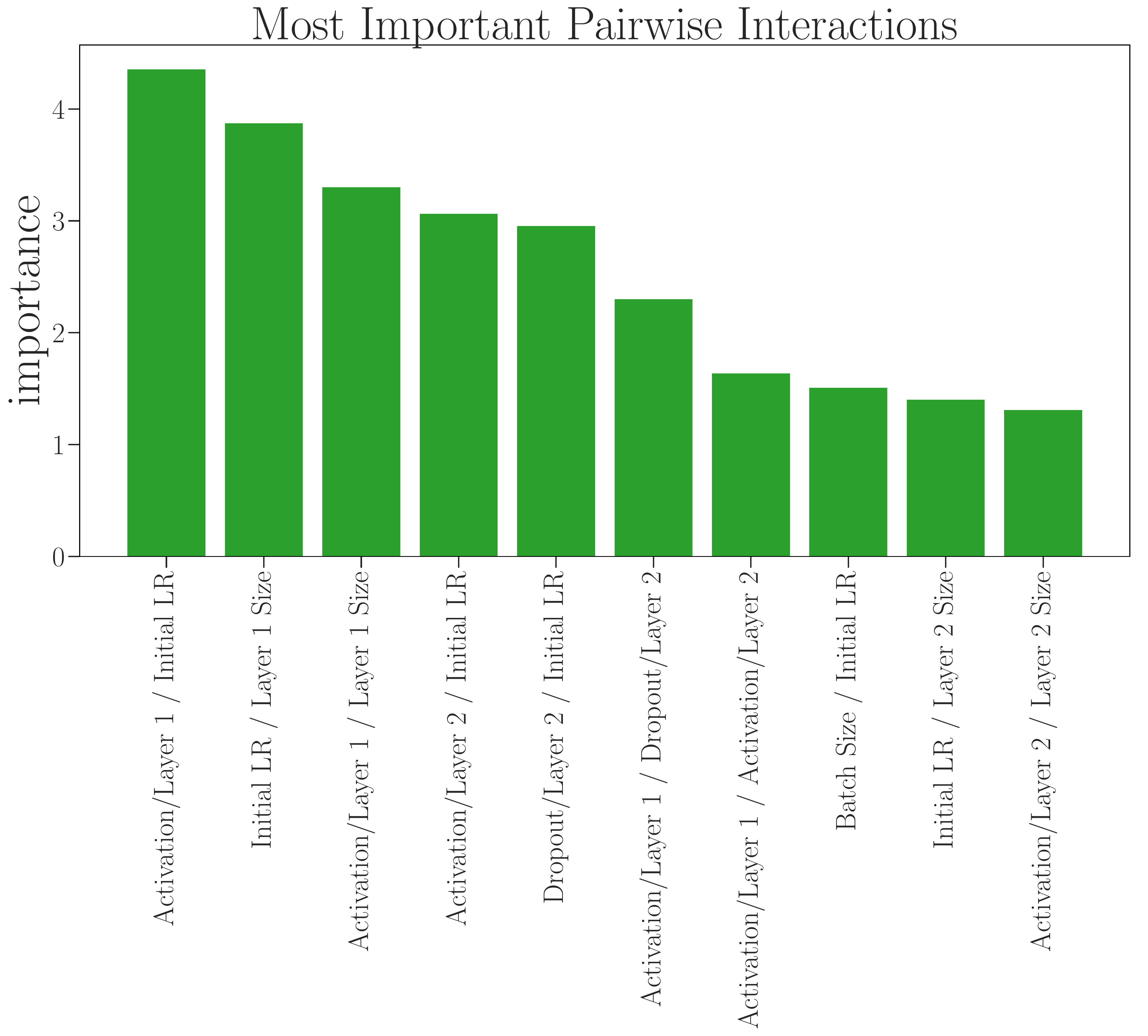}
\includegraphics[width=0.32\linewidth,valign=t]{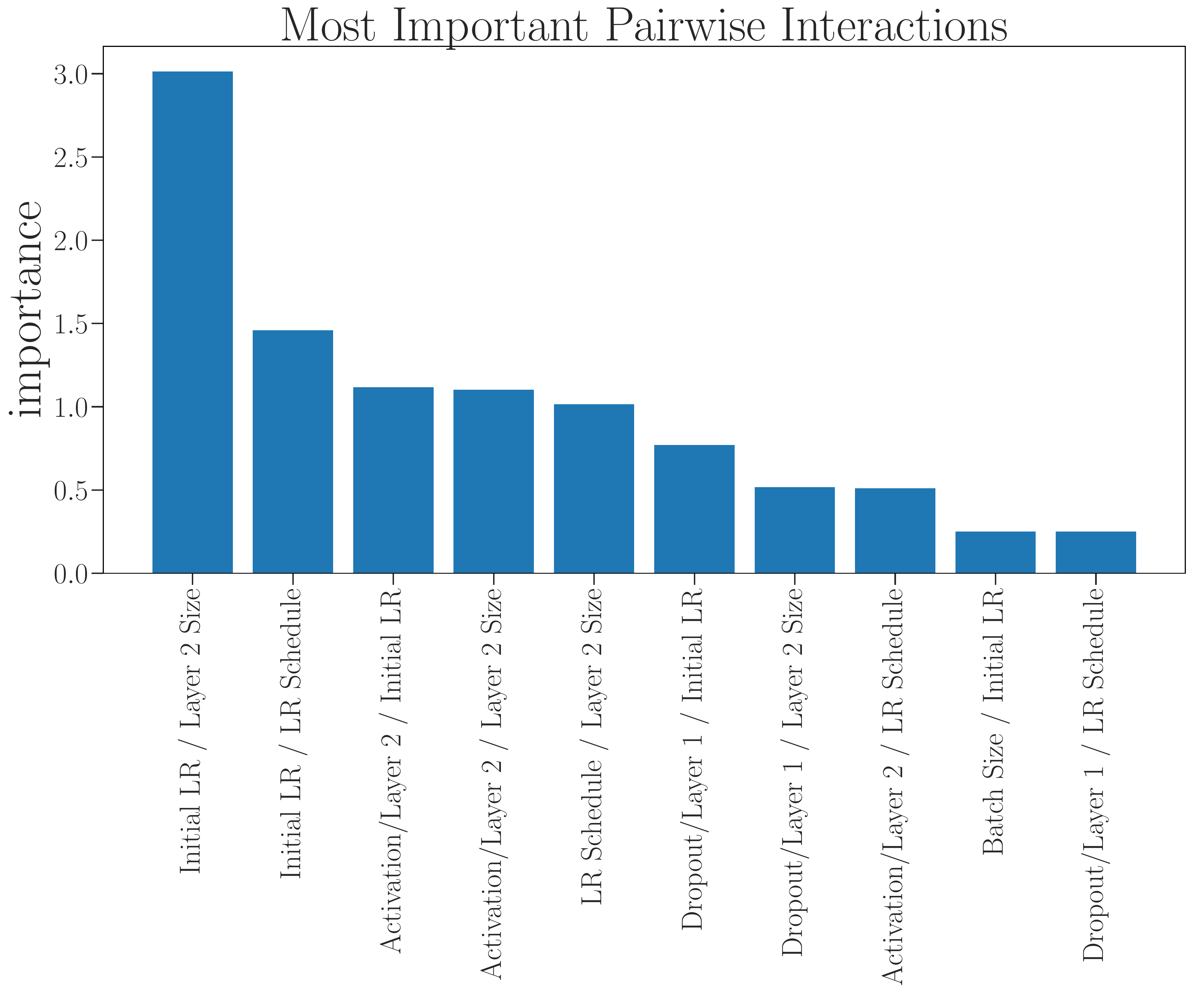}
\caption[fANOVA HPOBench]{HPOBench-Protein. Top row: Importance of the different hyperparameter based on the fANOVA for: (left) only the top $1\%$ ; (middle) top $10\%$ ; (right) all configurations. Bottom row: most important hyperparameter pairs with (left) only the top $1\%$ ; (middle) top $10\%$ ; (right) all configurations.}
\label{fig:importance_protein}
\end{figure}

\begin{figure}[t]
\centering
\includegraphics[width=0.32\linewidth]{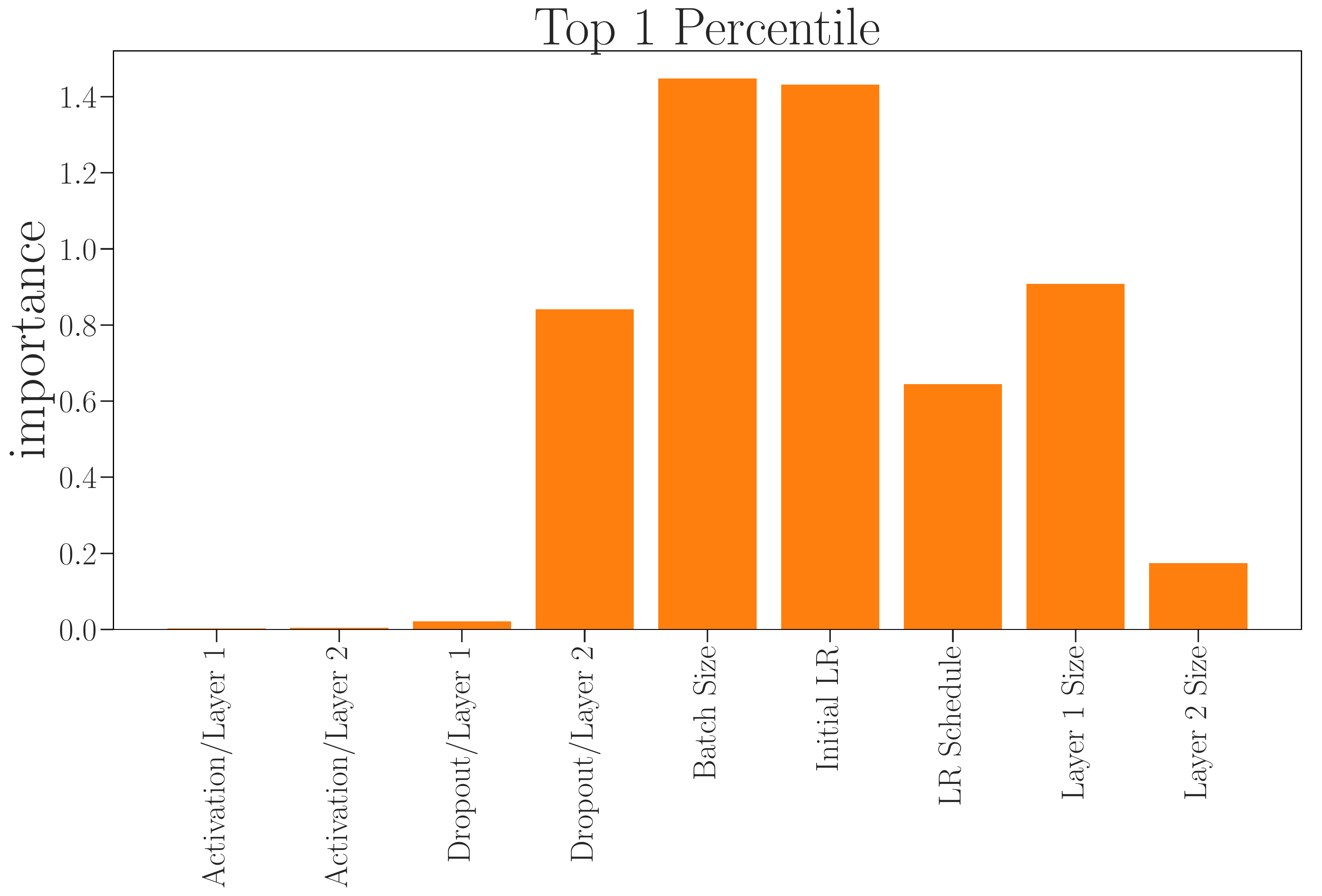}
\includegraphics[width=0.32\linewidth]{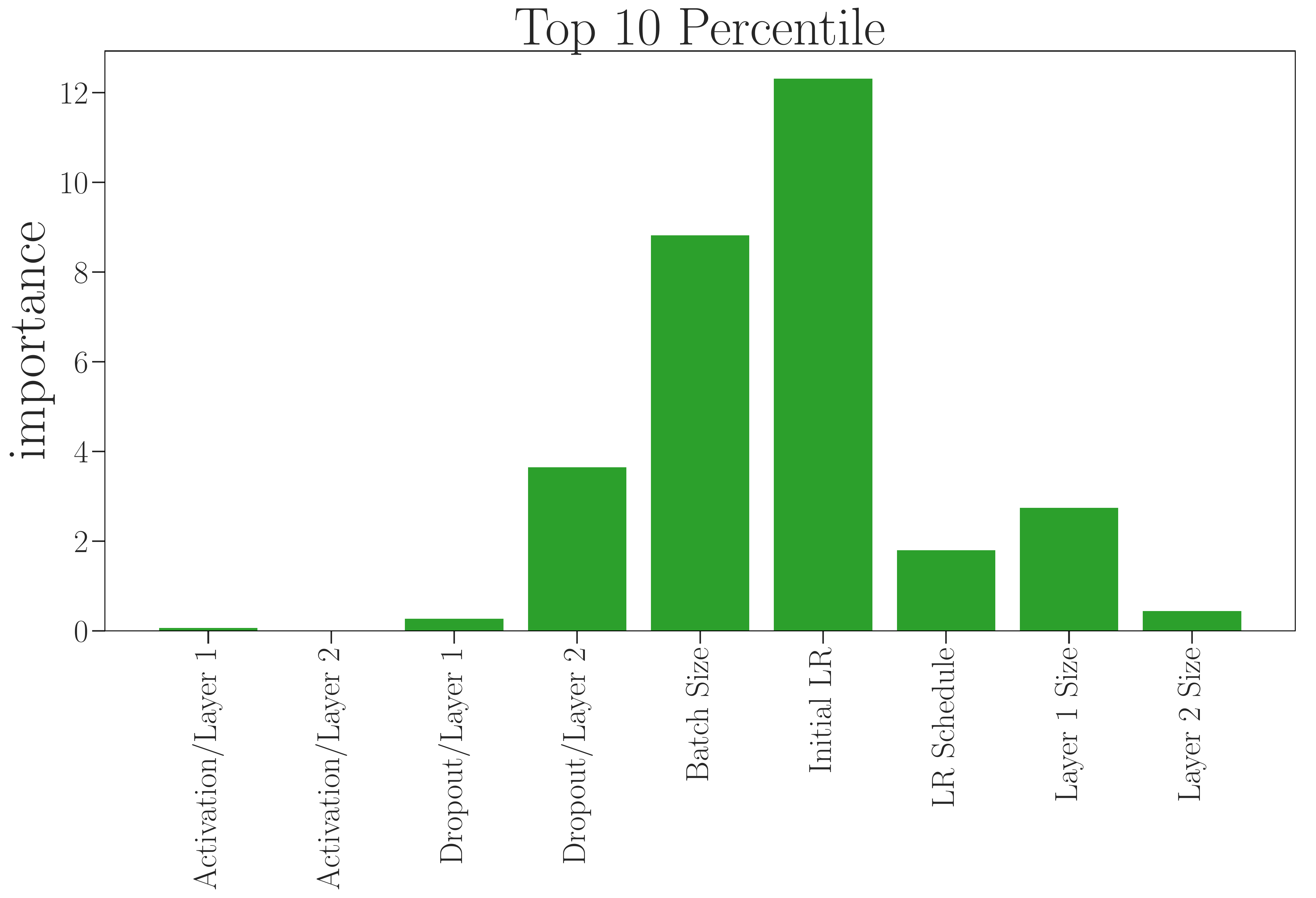}
\includegraphics[width=0.32\linewidth]{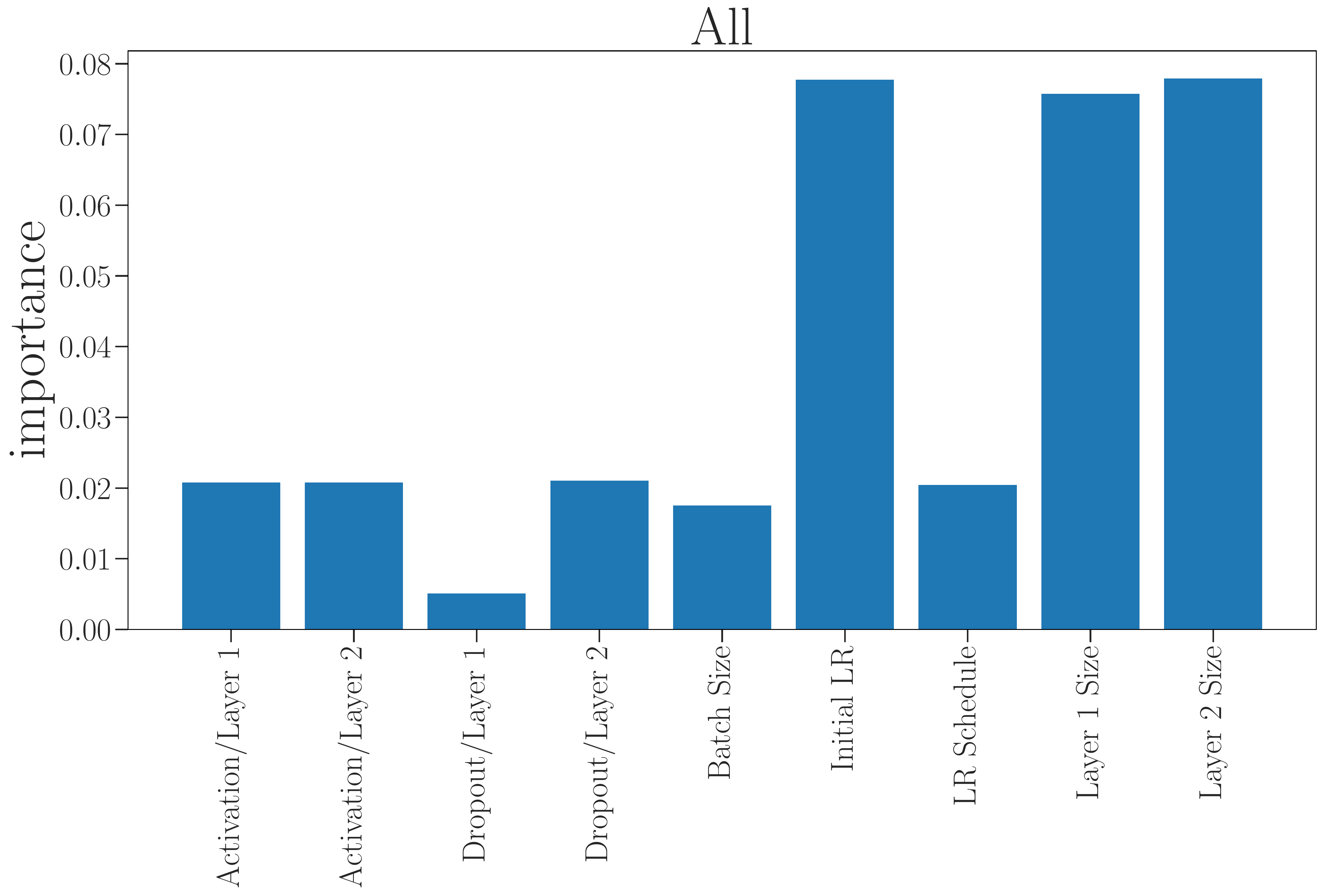}\\
\includegraphics[width=0.32\linewidth,valign=t]{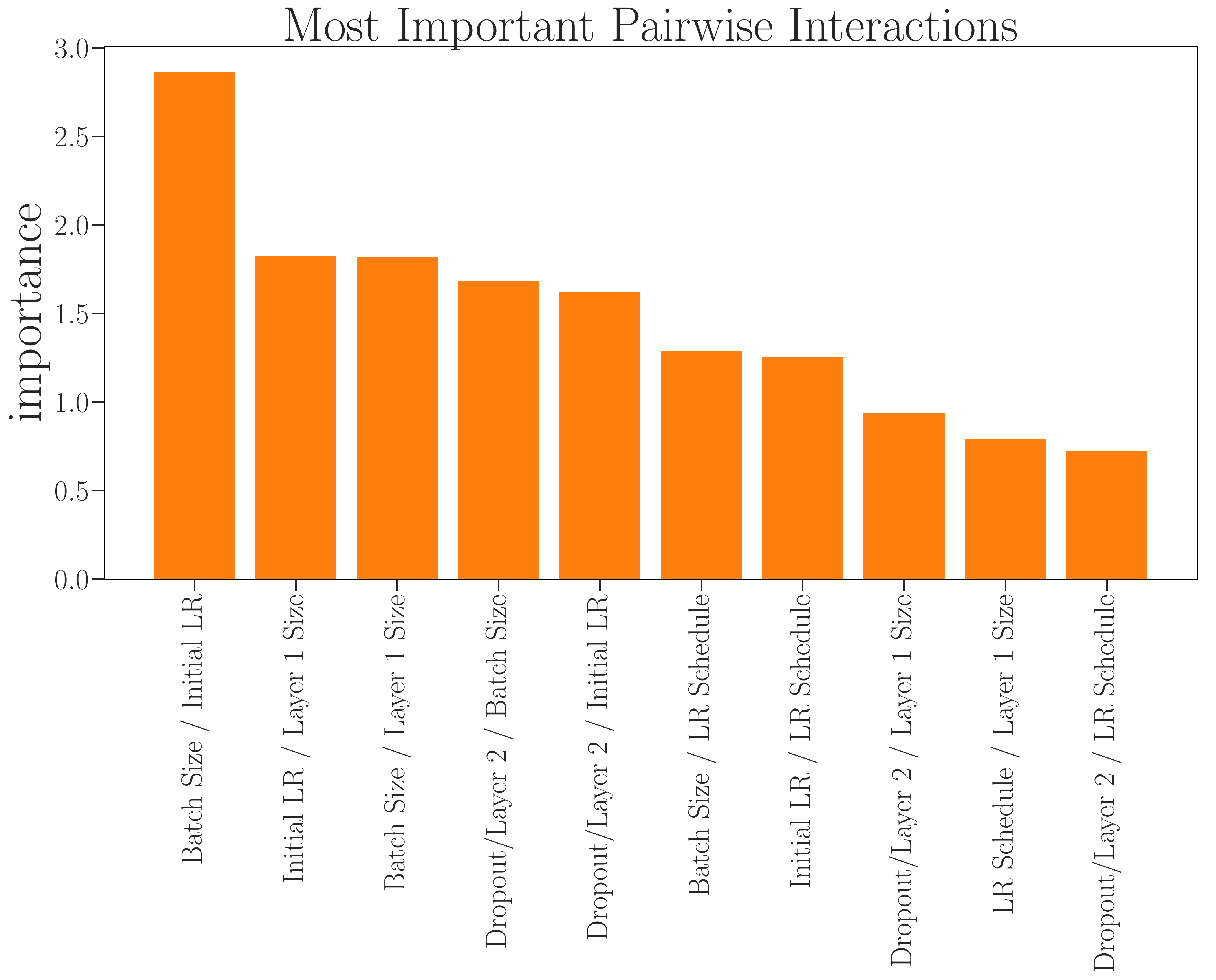}
\includegraphics[width=0.32\linewidth,valign=t]{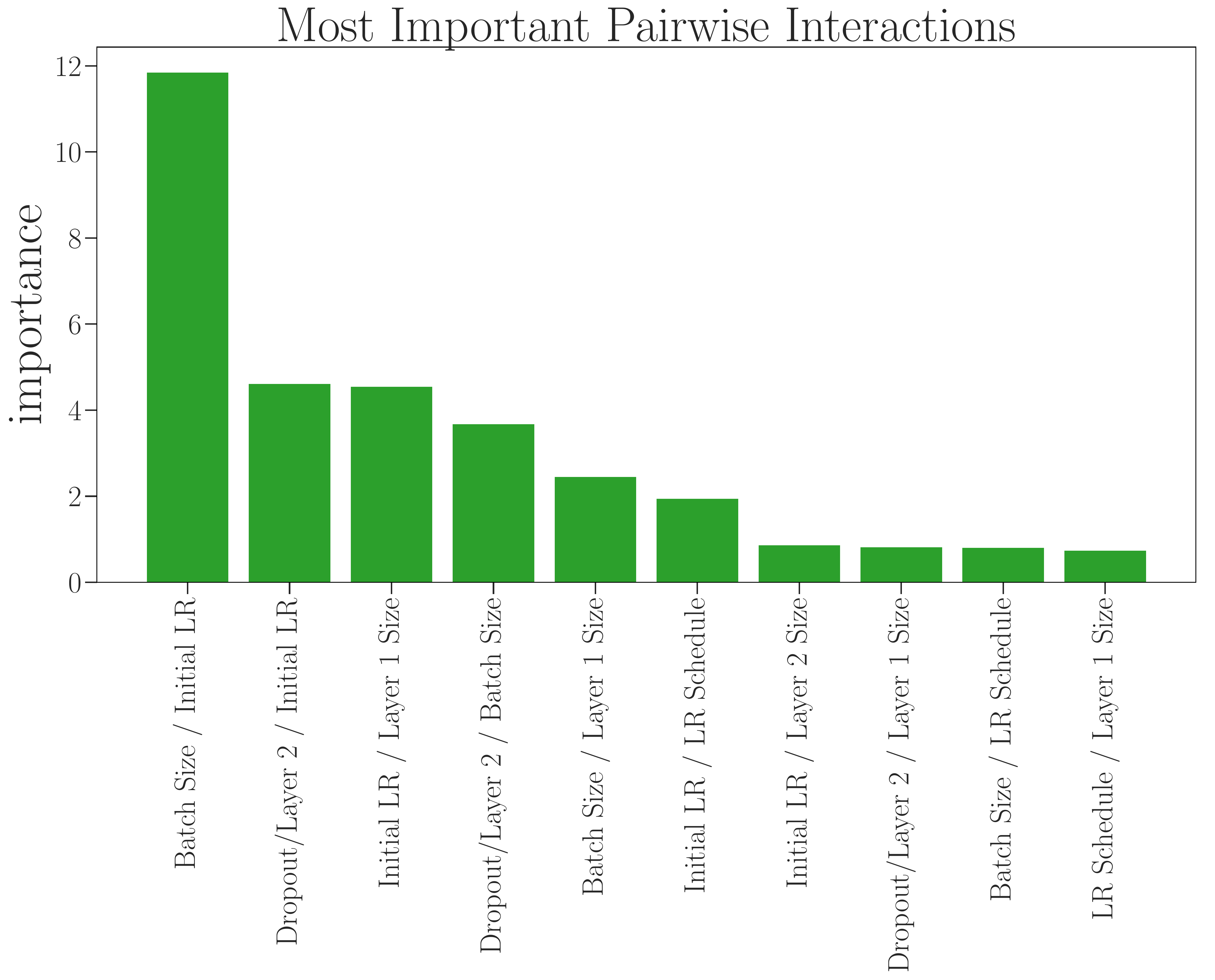}
\includegraphics[width=0.32\linewidth,valign=t]{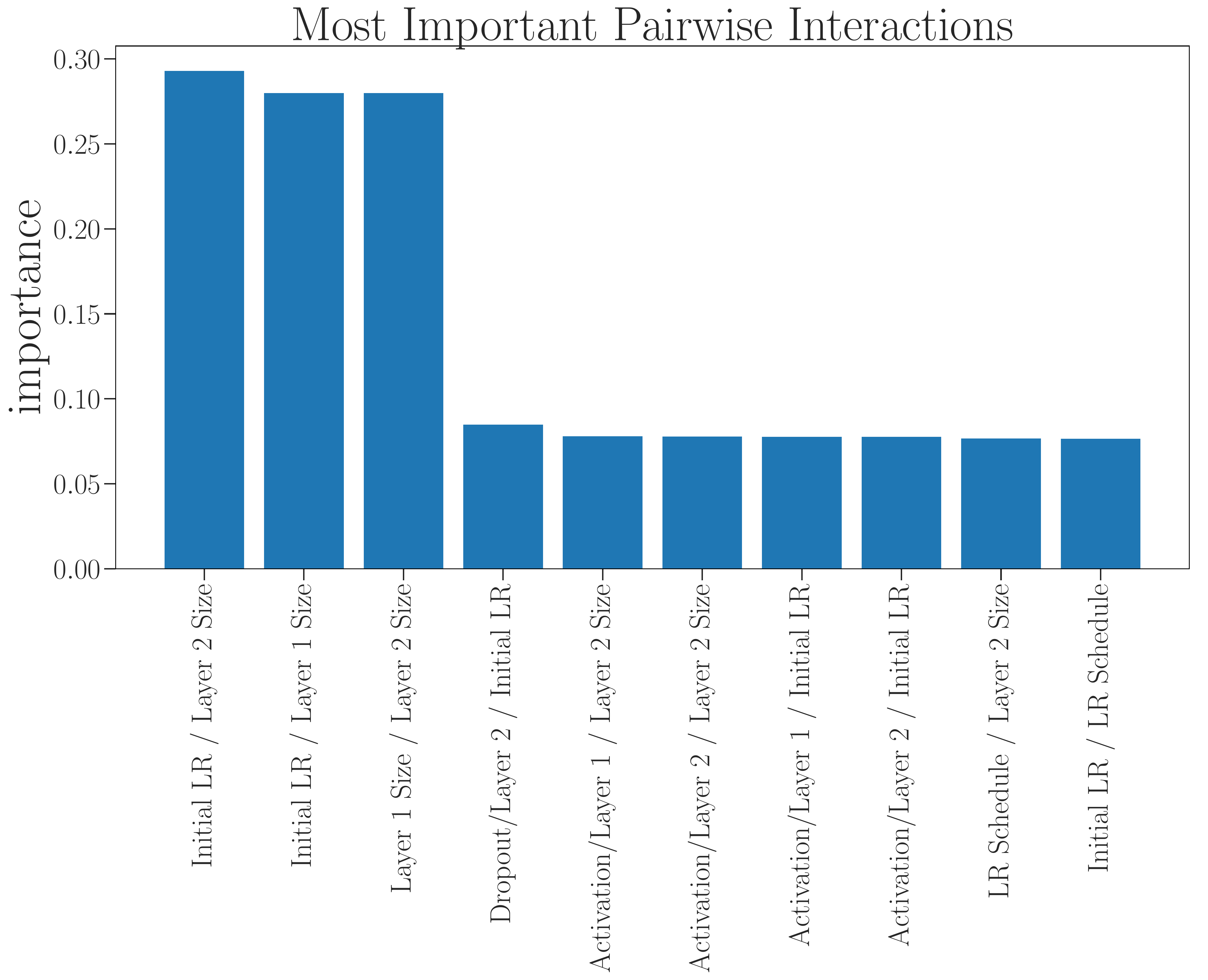}
\caption[fANOVA HPOBench]{HPOBench-Slice. Top row: Importance of the different hyperparameter based on the fANOVA for: (left) only the top $1\%$ ; (middle) top $10\%$ ; (right) all configurations. Bottom row: most important hyperparameter pairs with (left) only the top $1\%$ ; (middle) top $10\%$ ; (right) all configurations.}
\label{fig:importance_slice}
\end{figure}

Figure~\ref{fig:importance_naval}, \ref{fig:importance_parkinson}, \ref{fig:importance_protein} and \ref{fig:importance_slice} show the importance values based on the fANOVA tool for the top $1\%$ , top $10\%$ and all configurations as well as the most important pair-wise plots for \hpobench-Naval, \hpobench-Parkinson, \hpobench-Protein and \hpobench-Slice, respectively.
Table~\ref{tab:neighbors_naval}, \ref{tab:neighbors_parkinson}, \ref{tab:neighbors_protein} and \ref{tab:neighbors_slice} show the local neighbourhood for \hpobench-Naval, \hpobench-Parkinson, \hpobench-Protein and \hpobench-Slice.

\begin{table}[h]
  \center
  \caption[Local neighborhood of the incumbent]{\hpobench-Naval: Performance change if single hyperparameters of the incumbent (average test error 0.000029) are flipped.}
  \begin{tabular}{|c|c|c|c|}
    \hline
	Hyperparameter & Change & Test Error & Relative Change \\
	\hline
Layer 1 Size  & $128 \rightarrow 256$ & 0.0000 & 0.1331 \\
Initial LR & $0.0005 \rightarrow 0.001$ & 0.0000 & 0.1751 \\
Layer 1 Size  & $128 \rightarrow 64$ & 0.0000 & 0.2196 \\
Layer 2 Size  & $512 \rightarrow 256$ & 0.0000 & 0.4929 \\
Batch Size & $8 \rightarrow 16$ & 0.0000 & 0.5048 \\
Activation/Layer 1 & $tanh \rightarrow relu$ & 0.0000 & 0.6933 \\
Activation/Layer 2 & $relu \rightarrow tanh$ & 0.0002 & 4.9685 \\
Dropout/Layer 2 & $0.0 \rightarrow 0.3$ & 0.0004 & 11.1872 \\
Dropout/Layer 1 & $0.0 \rightarrow 0.3$ & 0.0010 & 34.3490 \\
LR Schedule & $cosine \rightarrow const$ & 0.0063 & 217.0092 \\

	\hline
  \end{tabular}
  \label{tab:neighbors_naval}
\end{table}

\begin{table}[h]
  \center
  \caption[Local neighborhood of the incumbent]{\hpobench-Parkinson: Performance change if single hyperparameters of the incumbent (average test error 0.004239) are flipped.}
  \begin{tabular}{|c|c|c|c|}
    \hline
	Hyperparameter & Change & Test Error & Relative Change \\
	\hline

Layer 1 Size  & $512 \rightarrow 256$ & 0.0051 & 0.2142 \\
Layer 2 Size  & $512 \rightarrow 256$ & 0.0054 & 0.2740 \\
Batch Size & $16 \rightarrow 32$ & 0.0059 & 0.3962 \\
Dropout/Layer 1 & $0.0 \rightarrow 0.3$ & 0.0081 & 0.9012 \\
Batch Size & $16 \rightarrow 8$ & 0.0085 & 1.0068 \\
Activation/Layer 1 & $tanh \rightarrow relu$ & 0.0106 & 1.5100 \\
Initial LR & $0.005 \rightarrow 0.001$ & 0.0111 & 1.6268 \\
Activation/Layer 2 & $tanh \rightarrow relu$ & 0.0178 & 3.1980 \\
Initial LR & $0.005 \rightarrow 0.01$ & 0.0189 & 3.4530 \\
Dropout/Layer 2 & $0.0 \rightarrow 0.3$ & 0.0216 & 4.0912 \\
LR Schedule & $cosine \rightarrow const$ & 0.1407 & 32.1805 \\

	\hline
  \end{tabular}
  \label{tab:neighbors_parkinson}
\end{table}

\begin{table}[h]
  \center
  \caption[Local neighborhood of the incumbent]{\hpobench-Protein: Performance change if single hyperparameters of the incumbent (average test error 0.2153) are flipped.}
  \begin{tabular}{|c|c|c|c|}
    \hline
	Hyperparameter & Change & Test Error & Relative Change \\
	\hline

Batch Size & $8 \rightarrow 16$ & 0.2163 & 0.0042 \\
Initial LR & $0.0005 \rightarrow 0.001$ & 0.2169 & 0.0072 \\
Layer 2 Size & $512 \rightarrow 256$ & 0.2203 & 0.0231 \\
Layer 1 Size & $512 \rightarrow 256$ & 0.2216 & 0.0288 \\
Dropout/Layer 2 & $0.3 \rightarrow 0.6$ & 0.2257 & 0.0478 \\
LR Schedule & $cosine \rightarrow const$ & 0.2269 & 0.0534 \\
Dropout/Layer 2 & $0.3 \rightarrow 0.0$ & 0.2280 & 0.0587 \\
Dropout/Layer 1 & $0.0 \rightarrow 0.3$ & 0.2307 & 0.0711 \\
Activation/Layer 2 & $relu \rightarrow tanh$ & 0.2875 & 0.3351 \\
Activation/Layer 1 & $relu \rightarrow tanh$ & 0.3012 & 0.3987 \\

	\hline
  \end{tabular}
  \label{tab:neighbors_protein}
\end{table}

\begin{table}[h]
  \center
  \caption[Local neighborhood of the incumbent]{\hpobench-Slice: Performance change if single hyperparameters of the incumbent (average test error 0.000144) are flipped.}
  \begin{tabular}{|c|c|c|c|}
    \hline
	Hyperparameter & Change & Test Error & Relative Change \\
	\hline

Layer 1 Size  & $512 \rightarrow 256$ & 0.0002 & 0.0831 \\
Batch Size & $32 \rightarrow 64$ & 0.0002 & 0.2014 \\
Dropout/Layer 1 & $0.0 \rightarrow 0.3$ & 0.0002 & 0.2514 \\
Layer 2 Size  & $512 \rightarrow 256$ & 0.0002 & 0.2535 \\
Batch Size & $32 \rightarrow 16$ & 0.0002 & 0.3087 \\
Activation/Layer 1 & $relu \rightarrow tanh$ & 0.0002 & 0.3383 \\
Activation/Layer 2 & $tanh \rightarrow relu$ & 0.0002 & 0.6326 \\
Initial LR & $0.0005 \rightarrow 0.001$ & 0.0003 & 0.7668 \\
Dropout/Layer 2 & $0.0 \rightarrow 0.3$ & 0.0006 & 3.3757 \\
LR Schedule & $cosine \rightarrow const$ & 0.0007 & 4.0016 \\

	\hline
  \end{tabular}
  \label{tab:neighbors_slice}
\end{table}

\section{Comparison HPOBench}\label{sec:supp_hpobench_comparison}

We now present a more detail discussion on how we set the meta-parameters of the individual optimizer for our comparison.
Code to reproduce the experiments is available at \url{https://github.com/automl/nas_benchmarks}. 

\textbf{Random Search (RS)}: We sample hyperparameter configurations from a uniform distribution over all possible hyperparameter configurations.

\textbf{Hyperband}: We set the $\eta=3$ which means that in each successive halving step only a third of the configurations are promoted to the next step. The minimum budget is set to 4 epochs and the maximum budget to 100 epochs of training.

\textbf{BOHB}: As for Hyperband we set $\eta=3$ and keep the same minimum and maximum budgets. 
The minimum possible bandwidth for the KDE is set to $0.3$ to prevent that the probability mass collapses to a single value.
The bandwidth factor is set to 3, the number of samples to optimize the acquisition function is 64,  and the fraction of random configurations is set to $\nicefrac{1}{3}$ which are the default values for BOHB.

\textbf{TPE}: We used all predefined meta-parameter values from the \href{https://github.com/hyperopt/hyperopt}{Hyperopt} package, since the python interface does not allow to change the meta-parameters.

\textbf{SMAC}: We set the maximum number of allowed function evaluations per configuration to 4. The number of trees for the random forest was set to 10 and the fraction of random configurations was set to $\nicefrac{1}{3}$ which are also the default values in the SMAC3 package.

\textbf{Regularized Evolution (RE)}: To mutate architectures, we first sample uniformly at random a hyperparameter and then sample a new value from the set of all possible values except the current one.
RE has two main hyperparameters, the population size and the tournament size, which we set to $100$ and $10$, respectively. 

\textbf{Reinforcement Learning (RL)}: Starting from a uniform distribution over the values of each hyperparameter, we used REINFORCE to optimize the probability values directly (see also~\citet{ying-arxiv19}. 
  After performing a grid search, we set the learning rate for REINFORCE to $0.1$ and used an exponential moving average as baseline for the reward function with a momentum of $0.9$.

\textbf{Bohamiann}: We used a 3 layer fully connected neural network with 50 units and tanh activation functions in each layer. We set the step length for the adaptive SGHMC sampler~\citep{springenberg-nips16} to $0.01$ and the batch size to $8$. Starting from a chain length of 20000 steps, the number of burn-in steps was linearly increased by a factor of 10 times the number of observed function values.
  To optimize the acquisition function, we used a simple local search method, that, starting from a random configuration, evaluates the one-step neighborhood and then jumps to the neighbor with the highest acquisition value until it either reaches the maximum number of steps or converges to a local optimum.

\begin{figure}[h!]
\centering
\includegraphics[width=.48\textwidth]{plots/comparison_protein.pdf}
\includegraphics[width=.48\textwidth]{plots/robustness_protein.pdf}\\
\includegraphics[width=.48\textwidth]{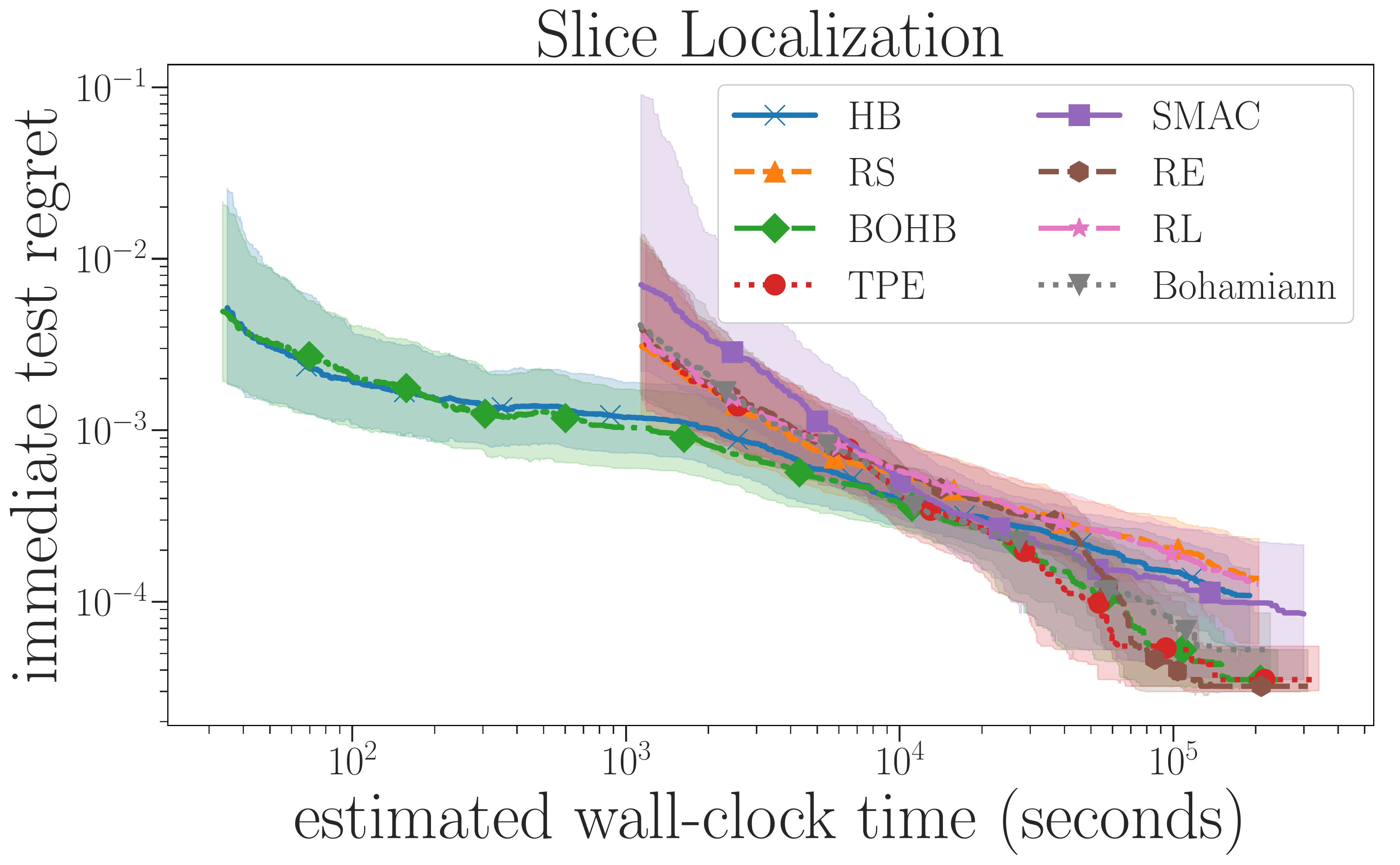}
\includegraphics[width=.48\textwidth]{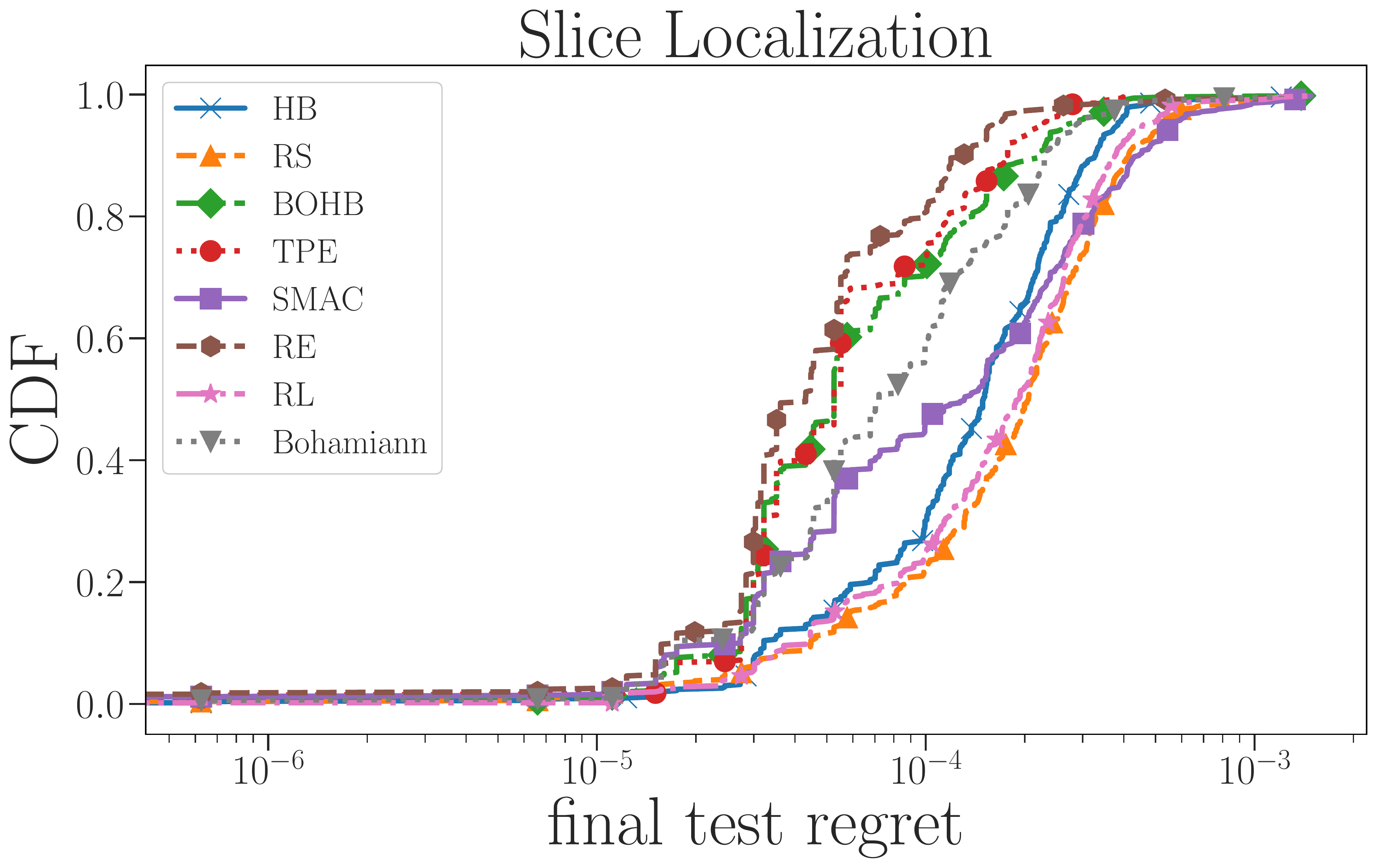}\\
\includegraphics[width=.48\textwidth]{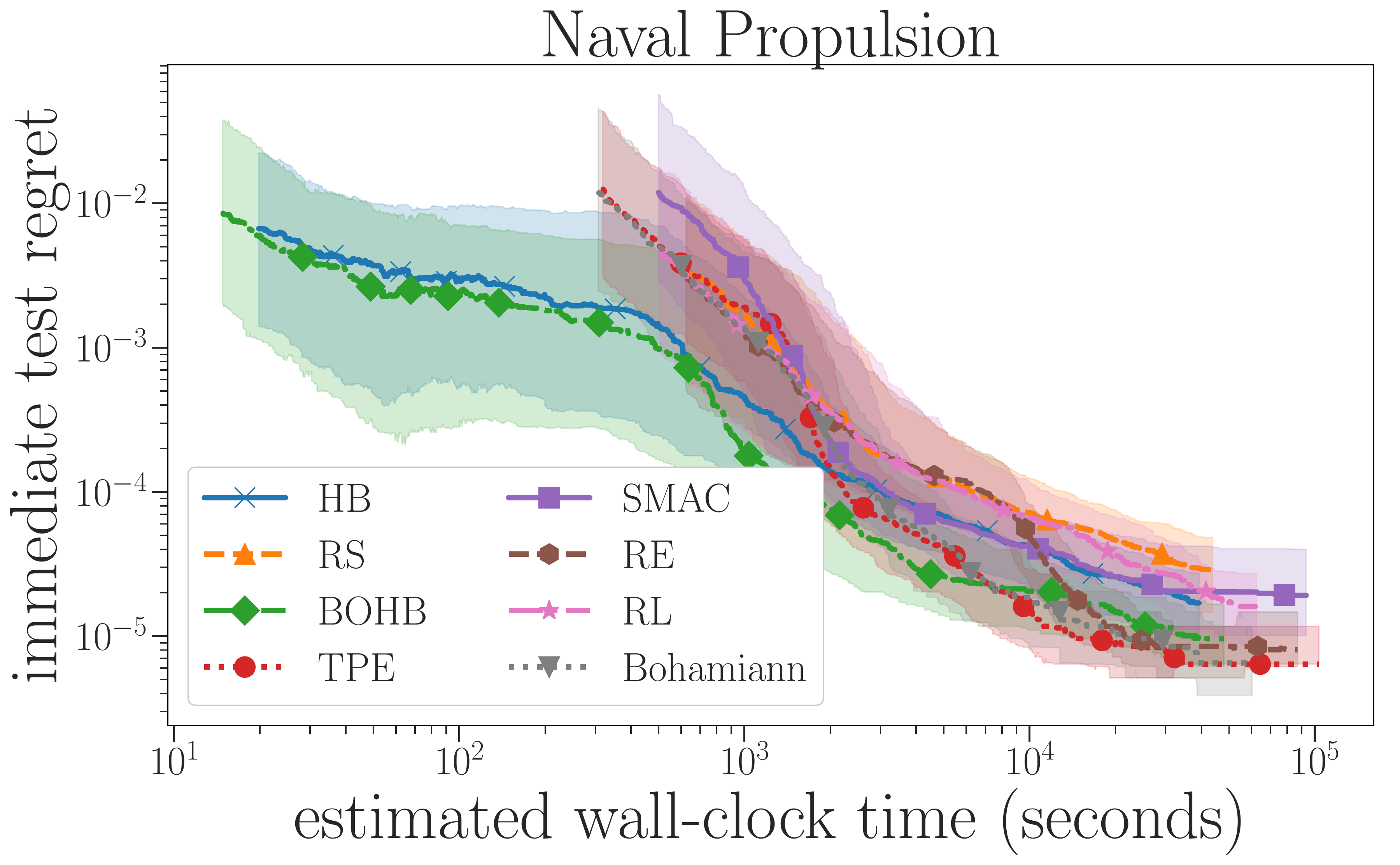}
\includegraphics[width=.48\textwidth]{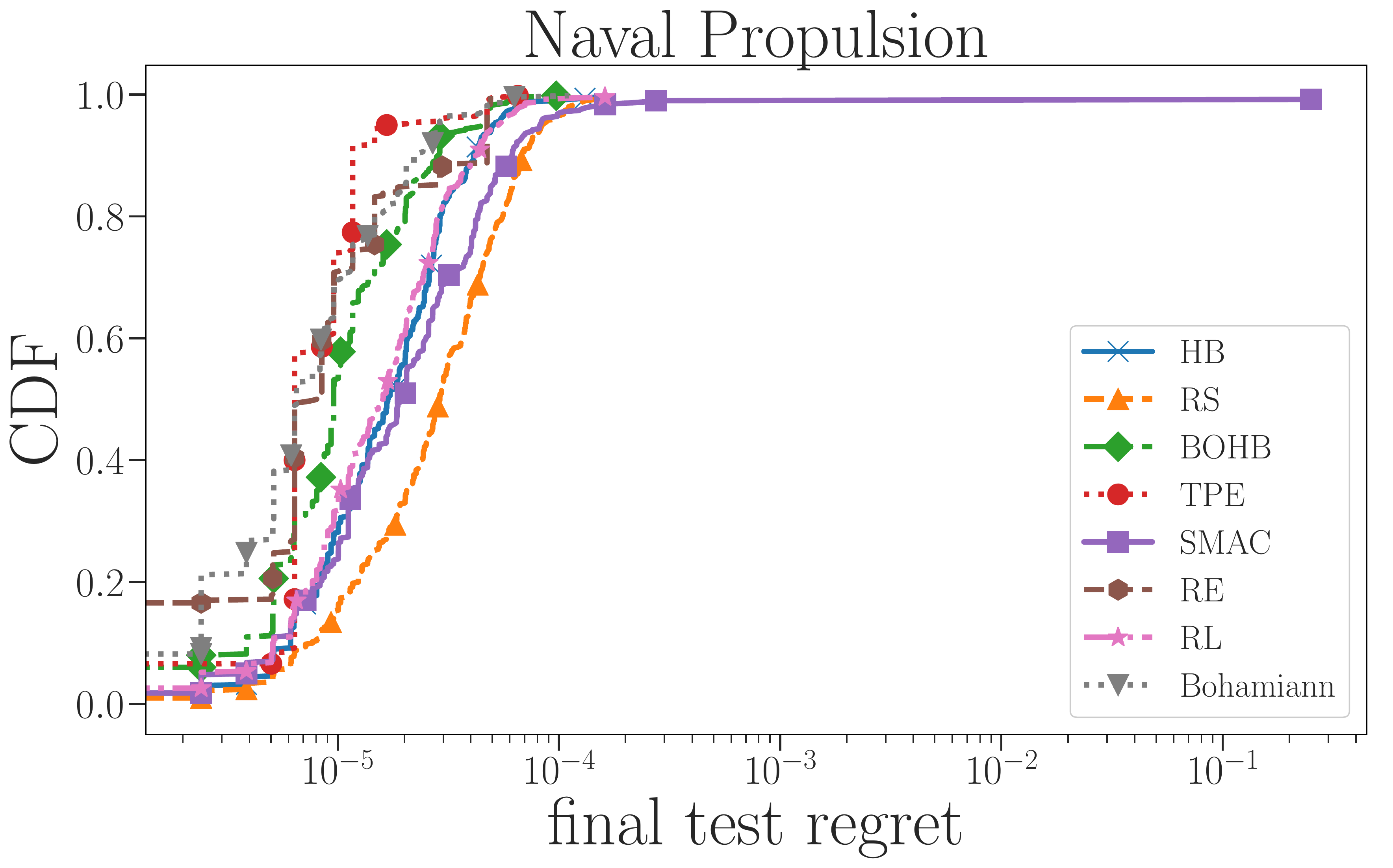}\\
\includegraphics[width=.48\textwidth]{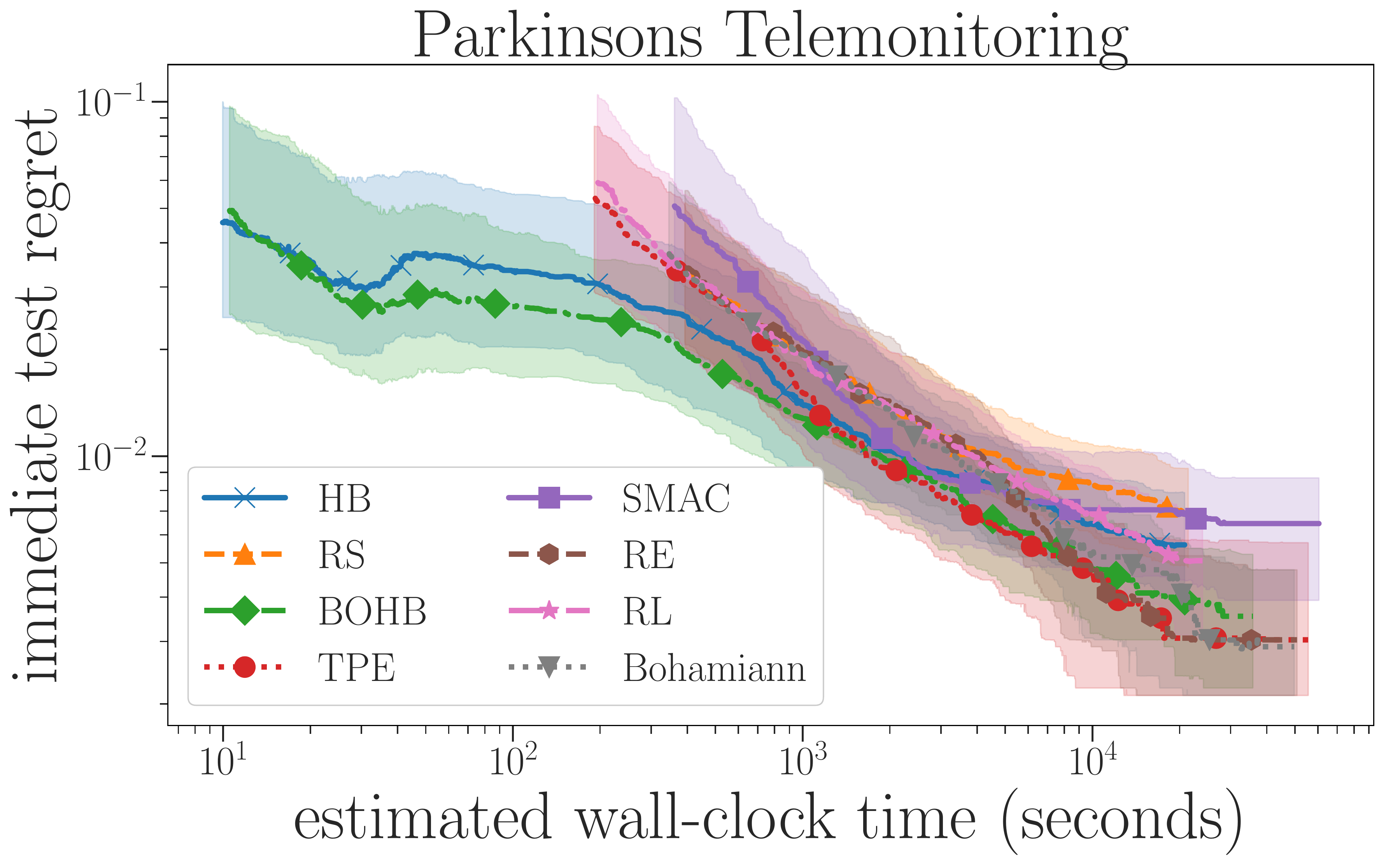}
\includegraphics[width=.48\textwidth]{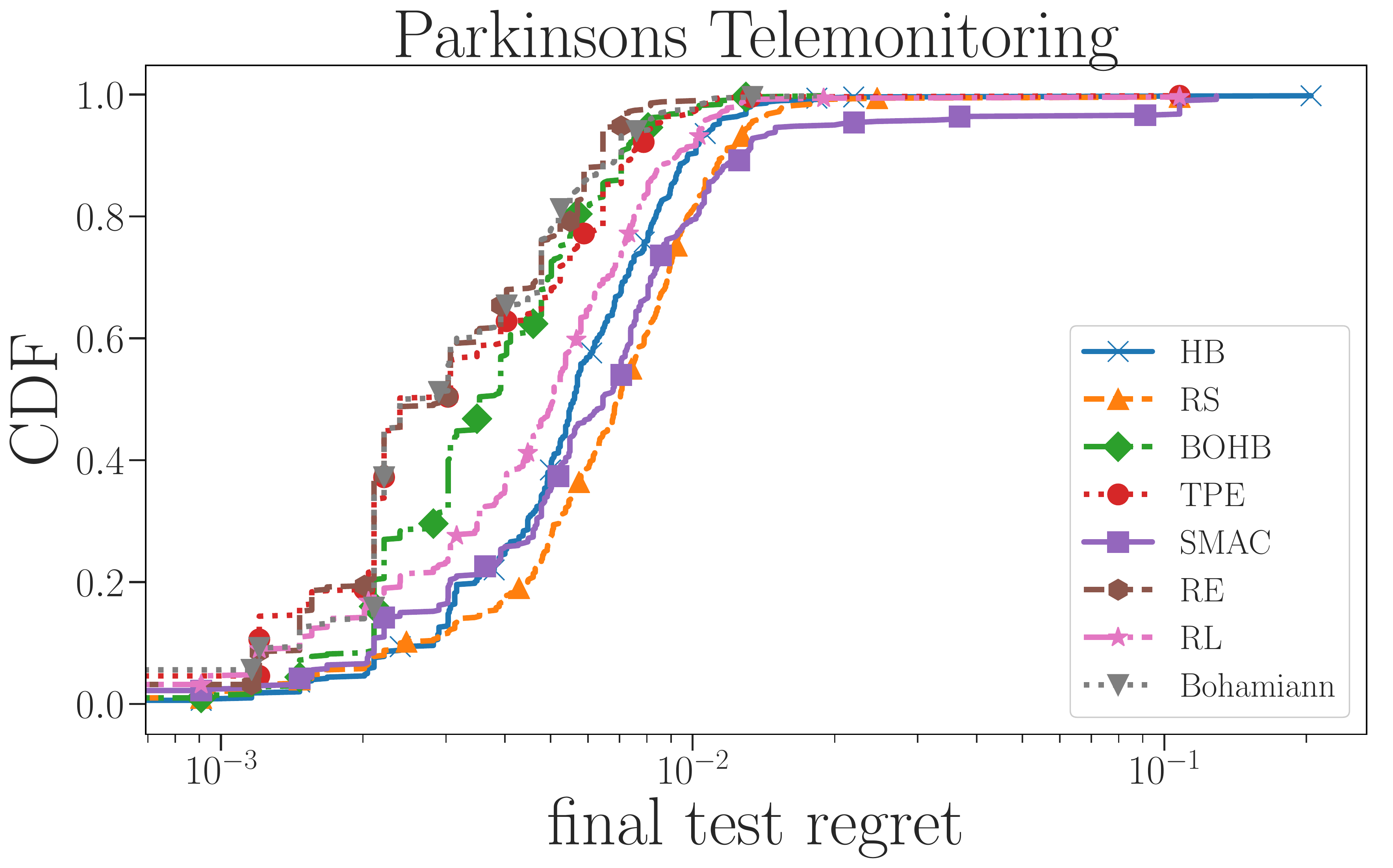}
\caption[Comparison HPOBench all dataset]{\label{fig:comparison_all} Left column: Comparison of various HPO methods on all the datasets. For each method, we plot the median and the 25th and 75th quantile of the test regret of the incumbent (determined based on the validation performance) across 500 independent runs. Right column: The empirical cumulative distribution of the final regret over all runs after $10^5$ seconds for \hpobench-Protein, $3 \times 10^5$ seconds for \hpobench-Slice, $6 \times 10^4$ for \hpobench-Naval and $3 \times 10^{4}$ seconds for \hpobench-Parkinson.}
\label{fig:comparison_hpobench_all}
\end{figure}

Figure~\ref{fig:comparison_hpobench_all} shows the comparison and the robustness of all considered hyperparameter optimization methods on the four tabular benchmarks.
We performed 500 independent runs for each method and report the mean and the standard error of the mean across all runs.
For a detailed analysis of the results see the main text.

\end{document}